\pdfoutput=1

\documentclass[11pt]{article}

\usepackage[final]{acl}

\usepackage{times}
\usepackage{latexsym}
\usepackage{booktabs}
\usepackage{multirow}
\usepackage{graphicx}
\usepackage{caption}
\usepackage{subcaption}
\usepackage{amsfonts}
\usepackage{amsmath}
\usepackage{algorithm}
\usepackage{algorithmic}
\usepackage{amssymb}
\usepackage{pifont}
\usepackage{xcolor}
\usepackage{bm}
\usepackage{tabularx}
\usepackage{pdflscape}
\usepackage{longtable}
\usepackage{array}
\usepackage{diagbox}
\usepackage{rotating}
\usepackage{adjustbox}
\usepackage{enumitem}

\usepackage[T1]{fontenc}

\usepackage[utf8]{inputenc}

\usepackage{microtype}

\usepackage{inconsolata}

\usepackage{graphicx}

\usepackage{amsmath,amsfonts,bm}

\def\eqref#1{equation~\ref{#1}}

\def\1{\bm{1}}

\DeclareMathAlphabet{\mathsfit}{\encodingdefault}{\sfdefault}{m}{sl}
\SetMathAlphabet{\mathsfit}{bold}{\encodingdefault}{\sfdefault}{bx}{n}

\usepackage{booktabs}
\usepackage{multirow}
\usepackage{graphicx}
\usepackage{caption}
\usepackage{subcaption}
\usepackage{amsfonts}
\usepackage{algorithm}
\usepackage{algorithmic}
\usepackage{amsfonts}
\usepackage{pifont}
\usepackage{xcolor}
\usepackage{xspace}
\usepackage{makecell}

\definecolor{deepblue}{rgb}{0,0,0.5}
\definecolor{officeblue}{RGB}{0,102,204}
\definecolor{deepred}{rgb}{0.6,0,0}
\definecolor{deepgreen}{rgb}{0,0.5,0}
\definecolor{mybrickred}{RGB}{182,50,28}
\definecolor{fillcolor}{RGB}{216,217,252}

\renewcommand{\algorithmiccomment}[1]{\bgroup\hfill $\triangleright$ ~#1\egroup}

\definecolor{color_m}{RGB}{72,117,170}
\definecolor{color_f}{RGB}{201,89,72}
\definecolor{color_c}{RGB}{230,230,230}
\definecolor{color_e}{RGB}{100,155,74}
\definecolor{ccon}{HTML}{fee9d4}
\definecolor{cood}{HTML}{d8f0d3}
\definecolor{cid}{HTML}{dae8f5}
\definecolor{gg}{HTML}{e2f0cb}

\newcommand{\ourbench}[0]{\textsc{EffiVLM-Bench}\xspace}

\title{\ourbench: A Comprehensive Benchmark for Evaluating Training-Free Acceleration in Large Vision-Language Models}

\author{
\textbf{
    Zekun Wang$^{1}$\thanks{Equal Contribution.},
    Minghua Ma$^{1}$$^*$,
    Zexin Wang$^{1}$$^*$, Rongchuan Mu$^{1}$$^*$}\\
\textbf{
    Liping Shan$^{2}$,
    Ming Liu$^{1,2}$,
    Bing Qin$^{1,2}$} \\
    $^{1}$Harbin Institute of Technology, Harbin, China \\
    $^{2}$Pengcheng Laboratory, Shenzhen, China. \\
    $^{3}$Du Xiaoman Science Technology Co., Ltd, Beijing, China \\
 \texttt{\{zkwang,mhma,zxwang,rcmu,mliu,qinb\}@ir.hit.edu.cn} \quad \\
  \normalsize 
  \textbf{Project Page:} \url{https://effivlm-bench.github.io/}
}

\begin{document}
\maketitle

\begin{abstract}
Large Vision-Language Models (LVLMs) have achieved remarkable success, yet their significant computational demands hinder practical deployment. 
While efforts to improve LVLM efficiency are growing, existing methods lack comprehensive evaluation across diverse backbones, benchmarks, and metrics.
In this work, we systematically evaluate mainstream acceleration techniques for LVLMs, categorized into token and parameter compression.
We introduce \ourbench, a unified framework for assessing not only absolute performance but also generalization and loyalty, while exploring Pareto-optimal trade-offs.
Our extensive experiments and in-depth analyses offer insights into optimal strategies for accelerating LVLMs.
We open-source code and recipes for \ourbench to foster future research.
\end{abstract}
\section{Introduction}
Large vision-language models (LVLMs)~\citep{gpt4o, team2024gemini, lu2024deepseekvl, wang2024qwen2, Qwen2.5-VL} have rapidly advanced, transforming multimodal AI and showing great potential for real-world applications~\citep{claude-computer-use,xu2024aguvis, openai-operator, li2025perception}.
However, their remarkable capabilities are often overshadowed by massive computational and memory costs, severely hindering practical deployment. 
To this end, some studies propose more efficient architectures~\citep{chu2023mobilevlm,zhou2024tinyllava,yao2024minicpmv,liu2024nvila} or incorporate distillation~\citep{shu2024llavamod,cai2024llavakd,li2024mini} to improve efficiency. 
But these typically demand full retraining, incurring substantial overhead~\citep{an2024flap}.

As a result, there has been a growing focus on training-free acceleration methods for LVLMs, which are more economical and can be broadly classified into two representative categories: \textbf{token compression}, eliminating redundant tokens in inputs~\citep{chen2024fastv, zhang2024sparsevlm} or KV cache~\citep{wan2024look,tu2024vlcache}, and \textbf{parameter compression}, which reduces parameter size by pruning~\citep{sung2023ecoflap} or quantization~\citep{lin2024awq,yu2025mquant}.
However, the scope of performance evaluation for these methods remains limited in several key aspects:
(1) \textit{Outdated Model Architectures}: Evaluations are ofter stuck on outdated models like LLaVA~\citep{liu2023visual} or LLaVA-v1.5~\citep{liu2024improved}, without considering state-of-the-art LVLMs with the dynamic resolution processing mechanism~\citep{li2024llava,wang2024qwen2,chen2024expanding}.
(2) \textit{Limited Benchmarks}: Assessments typically use general VQA tasks, neglecting more challenging benchmarks that require advanced capabilities such as OCR or long-context generation.
(3) \textit{Narrow Evaluation Metrics}: The focus is solely on absolute performance, overlooking other critical metrics such as generational quality and loyalty of compression methods. 
This narrow focus leaves a significant gap in understanding how these techniques generalize across broader scenarios, which severely limits their practical use. Furthermore, there is a lack of systematic exploration into the crucial trade-offs between performance and efficiency (actual inference time), which is essential for real-world LVLM deployment.

To address these limitations, we propose \ourbench, a unified evaluation framework for systematically assessing training-free acceleration methods of LVLMs. \ourbench spans a wide range of representative model architectures and tasks, employing comprehensive metrics for performance, generalization, loyalty, and efficiency.
With \ourbench, we conduct a thorough comparison of mainstream token and parameter compression methods.
We explore Pareto-optimal performance-efficiency trade-offs and offer a nuanced understanding of their strengths and limitations.
We hope our \ourbench can provide the research community with insightful perspectives on LVLM acceleration, paving the way for more effective and sustainable deployment.
\section{Related Work}
\label{sec:related_work}
Recent advancements in LVLMs~\citep{gpt4o, team2024gemini, li2024llava, wang2024qwen2, deitke2024molmo, chen2024expanding, Qwen2.5-VL} have led to significant improvements in various multimodal tasks.
Despite their remarkable capabilities, the computational overhead remains a critical bottleneck of LVLMs for real-world deployment. 
Some LVLM studies introduce more efficient model architectures~\citep{chu2023mobilevlm,zhou2024tinyllava,yao2024minicpmv,chen2024allava,luo2024gammamod} or knowledge distillation~\citep{wang2021distilled, wang2023efficientvlm, li2024mini} to enhance efficiency.
However, these methods typically require training from scratch or retraining, which incurs significant computational overhead.
Thus, increasing attention has been paid to training-free acceleration methods for LVLMs, the focus of this paper, which can be broadly categorized into token compression and parameter compression.

\paragraph{Token Compression}
The lengthy input tokens of LVLM impose a substantial computational burden, mainly due to the quadratic scaling of computational costs with the number of input tokens in the attention.
Some studies focus on directly \textbf{token pruning}, which prune the redundant visual tokens during the forward process~\citep{li2024tokenpacker,wang2024smarttrim,cha2024honeybee,huang2024dynamicllava}.
As for training-free methods, LLaVA-PruMerge~\cite{shang2024llavaprumerge} first prunes uninformative tokens in visual encoder and then merges the remaining ones using KNN techniques. 
\citet{yang2024visionzip} also dynamically prunes the tokens in the visual encoder and then merges the pruned tokens to the next layer, while \citet{chen2024fastv} and \citet{zhang2024sparsevlm} leverage textual information to guide visual token pruning in the initial layers of LLMs.
Another line of work mainly focuses on \textbf{KV cache compression}.
The inference time during LVLM generation is memory-bound, primarily due to the substantial memory consumption of KV caches, particularly when processing high-resolution images.
KV cache compression leverages attention sparsity to select fewer key-value pairs, thereby reducing memory overhead and enhancing inference speed.
Recent efforts in KV cache compression have evolved from techniques for LLMs~\citep{xiao2023streamingllm,zhang2023h2o,li2024snapkv,ge2024model,cai2024pyramidkv,xu2024thinkk} to methods tailored for LVLMs. 
VL-Cache~\citep{tu2024vlcache} introduces a modality-aware strategy that dynamically allocates cache budgets across layers and incorporates a token scoring mechanism tailored to the unique roles of visual and textual tokens. 
Similarly, LOOK-M~\citep{wan2024look} adopts a \textit{look-once} optimization that minimizes redundant cache entries through eviction and KV pair merging. 
Both methods show that targeted compression can significantly reduce memory usage and accelerate decoding in multimodal long-context scenarios while maintaining performance.

\paragraph{Parameter Compression}
Besides reducing the lengthy inputs, compressing the large model parameter size also benefits the efficiency of LVLM. 
Weight pruning removes redundant parameters to reduce parameter size and computations while preserving accuracy~\citep{sun2023simple,frantar2023sparsegpt,wang2023efficientvlm,an2024flap,wang2024cfsp}.  
Quantization\cite{frantar2022gptq,lin2024awq} converts full-precision weights and activations into lower-precision formats (e.g., int8 or int4). 
Most existing work focuses on compressing the parameters of the LLM backbone in LVLMs, while recent efforts have also begun exploring the compression of the visual encoder~\citep{yu2025mquant}.
In this work, we concentrate on reducing the parameters of LLMs, which is a key bottleneck of LVLM efficiency.
\section{\ourbench}
\label{sec:benchmark}
In this section, we introduce the details of \ourbench, including metrics (Section~\ref{sec:metrics}) and tasks with model architectures (Section~\ref{sec:models_and_tasks}).

\subsection{Metrics}
\label{sec:metrics}
Let $c$ denote the compression method, $m$ the target model, and $b$ the target benchmark. $\mathbb{E}(\cdot)$ denotes the mean operation.

\paragraph{Performance}
For each method $c$ on model $m$, we define the overall performance $\text{OP}^{m,c}$ as:
\begin{equation}
\text{OP}^{m,c} = \sqrt{\frac{1}{B} \sum_{b=1}^{B} \mathbb{E} \left(\frac{EM^{m,c}_{b}}{EM^{m}_{b}}\right)^2 },
\end{equation}
where $EM^{m,c}_{b}$ is the evaluation performance metric (e.g., accuracy) and $EM^{m}_{b}$ represents the corresponding metric for the original model.
$B$ denotes the number of benchmarks.

\paragraph{Generalization}
To evaluate the generalization of compression method $c$, we define the metric $\text{OG}^{c}$ as the coefficient of variation of performance across benchmarks and models:
\begin{equation}
\text{OG}^{c} = \frac{\sigma_b \left(\mathbb{E}_m \left[ \frac{EM^{m,c}_{b}}{EM^{m}_{b}} \right] \right)}{\mathbb{E}_{b,m} \left[ \frac{EM^{m,c}_{b}}{EM^{m}_{b}} \right]}
\end{equation}
where $ \sigma_b(\cdot) $ represents standard deviation across benchmarks. 
A lower $\text{OG}^{c}$ signifies more consistent relative performance (better generalization), while a higher value suggests greater sensitivity to variations in models or benchmarks.

\paragraph{Loyalty}
Loyalty ($\text{OL}^{c}$) measures how well a compression method $c$ preserves an original model's predictions ($P^{m}_{b}$) with its compressed version ($P_{b}^{m,c}$). Ideally, compression should maintain the original model's behavior, avoiding new biases or unexpected performance alterations~\citep{xu2021beyond}.
Specifically, we define the loyalty metric as:
\begin{equation}
\text{OL}^{c} = \mathbb{E}_{b,m} \left[ \mathbb{I}(P_{b}^{m,c}, P^{m}_{b}) \right]
\end{equation}
where $\mathbb{I}(P_1, P_2)$ is an agreement function between predictions.
Higher $\text{OL}^{c}$ indicates better loyalty.

\paragraph{Efficiency}
Actual inference time is our efficiency measure for a compression method $c$. Unlike FLOPs or parameter counts, it directly reflects real-world latency, which varies significantly with model architecture (e.g., depth vs.\ width) even for similar theoretical complexities.
Specifically, for model $ m $ using compression method $ c $ on benchmark $ b $, we measure the speedup of inference time per fixed number of samples:
\begin{equation}
\text{OE}^{c} = \mathbb{E}_{b,m} \left[ \frac{T^{m}_{b}}{T^{m,c}_{b}} \right],
\end{equation}
where $ T^{*,c} $ and $ T^{*} $ represent the latency of compressed and original models, respectively. 
More details of \ourbench are shown in Appendix~\ref{sec:app_implementation_details}.

\subsection{Tasks and Models}
\label{sec:models_and_tasks}
\ourbench evaluates across 17 widely-used benchmarks, including DocVQA\cite{mathew2020docvqa}, ChartQA\cite{masry2022chartqa}, TextVQA\cite{singh2019textvqa}, OCRBench\cite{Liu_2024ocrbench}, AI2D\cite{kembhavi2016diagramai2d}, GQA\cite{hudson2019gqa}, MMMU\cite{yue2023mmmu}, MME\cite{fu2024mmecomprehensiveevaluationbenchmark}, RealworldQA\cite{grok15v}, MMStar\cite{chen2024wemmstar}, MathVista\cite{lu2024mathvista}, LLaVA-Wilder\cite{li2024llavanext-strong}, MMBench\cite{liu2024mmbenchmultimodalmodelallaround}, MMVet\cite{yu2023mmvet}, ImageDC\cite{li2024llavanext-strong}, MP-DocVQA\cite{tito2022hierarchicalmpdocvqa} and MovieChat\cite{song2023moviechat}.
These cover diverse tasks---from document understanding and chart interpretation to real-world QA---across single-image, multi-image, and video scenarios (details in Appendix~\ref{sec:appendix_evaluation_benchmarks}).
\ourbench includes three frontier LVLMs: LLaVA-OneVision(OV)-7B~\citep{li2024llava}, Qwen2-VL-7B~\citep{wang2024qwen2}, and InternVL2.5-38B~\citep{chen2024expanding}, spanning diverse model sizes for comprehensive evaluation.
Furthermore, the modular design of \ourbench allows straightforward extension to support new tasks and emerging LVLMs.
\begin{table*}[t!]
\centering
\small
\resizebox{\linewidth}{!}{%
\begin{tabular}{c|c|c|*{15}c|c}

\toprule
\raisebox{5ex}{\textbf{Models}} & \raisebox{5ex}{\textbf{Budgets}} & 
\raisebox{5ex}{\diagbox{\textbf{Methods}}{\textbf{Benchmarks}}}
& 
\raisebox{1ex}{\adjustbox{valign=b}{\rotatebox[origin=b]{90}{DocVQA}}} & 
\raisebox{1ex}{\adjustbox{valign=b}{\rotatebox[origin=b]{90}{ChartQA}}} & 
\raisebox{1ex}{\adjustbox{valign=b}{\rotatebox[origin=b]{90}{TextVQA}}} & 
\raisebox{1ex}{\adjustbox{valign=b}{\rotatebox[origin=b]{90}{OCRBench}}} & 
\raisebox{1ex}{\adjustbox{valign=b}{\rotatebox[origin=b]{90}{AI2D}}} & 
\raisebox{1ex}{\adjustbox{valign=b}{\rotatebox[origin=b]{90}{GQA}}} & 
\raisebox{1ex}{\adjustbox{valign=b}{\rotatebox[origin=b]{90}{MMMU}}} & 
\raisebox{1ex}{\adjustbox{valign=b}{\rotatebox[origin=b]{90}{MME}}} & 
\raisebox{1ex}{\adjustbox{valign=b}{\rotatebox[origin=b]{90}{RealworldQA}}} & 
\raisebox{1ex}{\adjustbox{valign=b}{\rotatebox[origin=b]{90}{MMStar}}} & 
\raisebox{1ex}{\adjustbox{valign=b}{\rotatebox[origin=b]{90}{MathVista}}} & 
\raisebox{1ex}{\adjustbox{valign=b}{\rotatebox[origin=b]{90}{LLaVA-Wilder}}} & 
\raisebox{1ex}{\adjustbox{valign=b}{\rotatebox[origin=b]{90}{MMBench}}} & 
\raisebox{1ex}{\adjustbox{valign=b}{\rotatebox[origin=b]{90}{MMVet}}} & 
\raisebox{1ex}{\adjustbox{valign=b}{\rotatebox[origin=b]{90}{ImageDC}}} & 
\raisebox{5ex}{\textbf{OP}} \\
\midrule
\multirow{10}{*}{LLaVA-OneVision-7B} 
   & \multirow{3}{*}{1\%} 
      & FastV     & 8   & 14.00 & 9.18  & 27   & 66.35 & 35.04 & 40.89  & 697.7  & 36.86 & 28.97 & 36.40 & 49.10 & 27.69 & 12.30 & 20.95 & 0.48 \\
   &                 & VisionZip  & \textbf{35}  & \textbf{35.16} & 44.48 & \textbf{194}  & \textbf{72.11} & 53.69 & 42.56  & \textbf{1704.2} & 53.86 & 41.30 & 39.30 & \textbf{71.20} & \textbf{74.10} & 28.60 & \textbf{87.50} & \bf 0.75 \\
   &                 & PruMerge+ & 27  & 34.88 & \textbf{44.66} & 121  & 71.08 & \textbf{54.18} & \textbf{43.44}  & 1639.2 & \textbf{58.16} & \textbf{42.81} & 39.20 & 63.60 & 73.09 & \textbf{34.50} & 76.60 & 0.74 \\
\cmidrule(lr){2-19}
   & \multirow{3}{*}{10\%} 
      & FastV     & 48  & 43.16 & 52.37 & 190  & 72.57 & 49.63 & \textbf{45.33}  & 1669.4 & 53.86 & 44.63 & 40.30 & 66.90 & 70.12 & 31.30 & 77.75 & 0.76 \\
   &                 & VisionZip  & \textbf{56}  & \textbf{49.88} & 57.26 & \textbf{352}  & 77.97 & \textbf{58.57} & 44.11  & \textbf{1915.0} & 59.87 & 45.45 & \textbf{45.00} & \textbf{71.80} & \textbf{78.86} & 36.60 & \textbf{87.50} & \bf 0.84 \\
   &                 & PruMerge+ & 37  & 40.96 & 55.59 & 203  & 74.77 & 58.54 & 44.44  & 1872.7 & \textbf{61.17} & \textbf{47.56} & 43.50 & 65.90 & 78.47 & \textbf{37.00} & 85.35 & 0.81 \\
\cmidrule(lr){2-19}
   & \multirow{3}{*}{40\%} 
      & FastV     & \textbf{80}  & \textbf{69.20} & 72.48 & 488  & 86.23 & 60.56 & 46.33  & 1937.2 & 62.75 & \textbf{53.91} & 50.50 & 70.50 & \textbf{81.22} & 47.70 & 86.35 & \bf 0.94 \\
   &                 & VisionZip  & 72  & 67.04 & 68.21 & \textbf{500}  & 83.84 & 61.23 & 46.11  & 1956.8 & 63.01 & 51.17 & 52.60 & \textbf{71.60} & 80.43 & \textbf{48.30} & \textbf{87.55} & 0.93 \\
   &                 & PruMerge+ & 49  & 51.40 & 67.79 & 382  & 79.82 & \textbf{61.78} & 45.55  & 1924.0 & \textbf{64.70} & 53.38 & 48.00 & 68.10 & 80.88 & 46.50 & 85.55 & 0.88 \\
\cmidrule(lr){2-19}
   & 100\% & Original & 87 & 80.00 & 74.79 & 595 & 89.96 & 61.92 & 45.44 & 1974.1 & 65.88 & 58.75 & 58.20 & 71.40 & 83.12 & 55.00 & 87.25 & 1.00 \\
\midrule
\multirow{10}{*}{Qwen2-VL-7B} 
   & \multirow{3}{*}{1\%} 
      & FastV     & 19  & 12.24 & 22.45 & 71   & 65.22 & 39.79 & 44.67  & 1330.0 & 41.18 & 26.83 & 33.70 & 51.40 & 21.91 & 18.10 & 24.60 & 0.51 \\
   &                 & VisionZip  & \textbf{52}  & 34.76 & \textbf{61.97} & \textbf{187}  & \textbf{70.23} & 48.64 & \textbf{46.56}  & 1702.0 & \textbf{58.56} & \textbf{35.77} & 39.10 & 60.80 & 48.99 & \textbf{39.00} & \textbf{66.15} & \bf 0.70 \\
   &                 & PruMerge+ & 45  & \textbf{38.80} & 60.16 & 162  & 68.04 & \textbf{51.45} & 45.67  & \textbf{1772.2} & 56.86 & 35.70 & \textbf{39.50} & \textbf{63.00} & \textbf{50.67} & 34.60 & 63.10 & 0.69 \\
\cmidrule(lr){2-19}
   & \multirow{3}{*}{10\%} 
      & FastV     & 68  & 31.04 & 68.72 & 272  & 73.28 & 50.33 & 48.00  & 1750.0 & 55.82 & 40.09 & 40.40 & 66.10 & 63.06 & 42.90 & 78.70 & 0.76 \\
   &                 & VisionZip  & \textbf{75}  & \textbf{57.88} & \textbf{72.27} & \textbf{318}  & \textbf{76.06} & 55.08 & \textbf{48.89}  & 1860.0 & 60.65 & 42.56 & \textbf{44.50} & \textbf{67.20} & 63.06 & \textbf{45.10} & 79.85 & \bf 0.81 \\
   &                 & PruMerge+ & 61  & 48.76 & 68.16 & 283  & 70.98 & \textbf{56.77} & 47.67  & \textbf{1981.2} & \textbf{62.09} & \textbf{43.17} & 43.70 & \textbf{67.20} & \textbf{68.27} & 42.40 & \textbf{81.80} & 0.80 \\
\cmidrule(lr){2-19}
   & \multirow{3}{*}{40\%} 
      & FastV     & \textbf{92}  & 67.40 & \textbf{79.52} & 532  & \textbf{86.53} & 58.35 & 49.22  & \textbf{2185.0} & \textbf{66.93} & 50.74 & 47.10 & \textbf{71.60} & 75.50 & 59.40 & \textbf{85.00} & 0.92 \\
   &                 & VisionZip  & 91  & \textbf{74.04} & 79.25 & \textbf{571}  & 85.10 & 60.21 & \textbf{49.44}  & 2150.0 & 63.79 & 51.09 & \textbf{53.40} & 70.50 & 73.76 & \textbf{59.50} & 84.70 & \bf 0.93 \\
   &                 & PruMerge+ & 85  & 68.96 & 76.97 & 490  & 75.68 & \textbf{60.64} & 49.00  & 2169.7 & 66.27 & \textbf{51.94} & 51.70 & 70.00 & \textbf{75.56} & 54.90 & 84.95 & 0.91 \\
\cmidrule(lr){2-19}
   & 100\% & Original & 95 & 81.56 & 81.82 & 813 & 91.02 & 62.30 & 50.77 & 2327.8 & 66.53 & 57.11 & 58.30 & 72.70 & 78.25 & 67.40 & 86.35 & 1.00 \\
\midrule
\multirow{10}{*}{InternVL2.5-38B} 
   & \multirow{3}{*}{1\%} 
      & FastV     & 11  & 14.88 & 10.64 & 23  & 67.64 & 35.49 & 51.22  & 1264.1 & 44.44 & 29.64 & 34.60 & 52.70 & 22.64 & 7.10  & 20.20 & 0.47 \\
   &                 & VisionZip  & \textbf{15}  & \textbf{18.56} & \textbf{21.21} & \textbf{57}  & \textbf{69.04} & 42.34 & \textbf{53.89}  & 1612.1 & \textbf{52.67} & 35.25 & 32.90 & \textbf{58.80} & 45.79 & \textbf{23.30} & 44.35 & 0.55 \\
   &                 & PruMerge+ & 12  & 16.44 & 15.20 & 36  & 68.91 & \textbf{47.43} & 53.44  & \textbf{1655.3} & 47.97 & \textbf{35.34} & \textbf{34.90} & 55.80 & \textbf{50.73} & 11.50 & \textbf{58.60} & \bf 0.56 \\
\cmidrule(lr){2-19}
   & \multirow{3}{*}{10\%} 
      & FastV     & 13  & 16.76 & 17.41 & 40  & 69.55 & 45.88 & 54.11  & 1587.9 & 49.01 & 39.39 & 38.10 & 61.70 & 53.13 & 24.20 & 62.65 & 0.58 \\
   &                 & VisionZip  & \textbf{39}  & \textbf{34.84} & \textbf{54.34} & \textbf{303} & \textbf{78.98} & \textbf{58.77} & \textbf{56.89}  & \textbf{2008.6} & \textbf{64.18} & \textbf{52.17} & \textbf{44.60} & \textbf{68.20} & \textbf{78.30} & \textbf{46.60} & \textbf{81.45} & \bf 0.76 \\
   &                 & PruMerge+ & 19  & 26.28 & 31.09 & 62  & 71.73 & 56.22 & 53.66  & 1880.1 & 56.99 & 45.34 & 44.10 & 63.90 & 72.93 & 35.60 & 79.30 & 0.67 \\
\cmidrule(lr){2-19}
   & \multirow{3}{*}{40\%} 
      & FastV     & 55  & 40.60 & 56.04 & 268 & 82.15 & 56.54 & 54.66  & 2084.8 & 64.31 & 53.56 & 50.80 & 69.50 & 79.26 & 44.90 & 83.25 & 0.78 \\
   &                 & VisionZip  & \textbf{77}  & \textbf{77.52} & \textbf{78.07} & \textbf{609} & \textbf{91.77} & \textbf{63.49} & \textbf{58.56}  & \textbf{2397.2} & \textbf{70.72} & \textbf{64.29} & 59.50 & \textbf{72.60} & \textbf{85.31} & \textbf{60.80} & \textbf{85.65} & \bf 0.93 \\
   &                 & PruMerge+ & 53  & 58.56 & 67.85 & 454 & 84.03 & 62.86 & 56.11  & 2338.0 & 69.41 & 59.94 & 55.00 & 71.60 & 83.52 & 55.00 & 85.15 & 0.84 \\
\cmidrule(lr){2-19}
   & 100\% & Original & 94 & 88.04 & 82.87 & 802 & 95.11 & 64.48 & 61.56 & 2453.9 & 72.41 & 69.80 & 70.20 & 74.80 & 86.94 & 69.40 & 86.45 & 1.00 \\
\bottomrule
\end{tabular}
}
\caption{Main results of various visual token prune methods on different models and benchmarks. \textbf{Bold} denotes the best result under the same setting.}
\label{tab:token_prune_results}
\end{table*}

\section{Results}
\label{sec:results}

\subsection{Token Compression}
\paragraph{Setup}
We evaluate token compression effectiveness by examining two mainstream approaches: (1) token pruning, including FastV~\citep{chen2024fastv}, VisionZip~\citep{yang2024visionzip}, and PruMerge+~\citep{shang2024llavaprumerge}, which eliminates redundant visual tokens. (2) KV cache compression, including streamingLLM~\citep{xiao2023streamingllm}, H2O~\citep{zhang2023h2o}, SnapKV~\citep{li2024snapkv}, PyramidKV~\citep{cai2024pyramidkv}, LOOK-M~\citep{wan2024look}, and VL-Cache~\citep{tu2024vlcache}.
Notably, the last two methods are specifically designed for LVLMs.
More implementation details are in Appendix~\ref{sec:appendix_math_modeling} and~\ref{sec:app_implementation_details}.
We focus on retention budgets up to 40\%, as methods generally maintain original performance at higher budgets.
Main results of token pruning and KV cache compression are shown in Table~\ref{tab:token_prune_results} and Figure~\ref{fig:main_kvcache}. More results of KV cache compression are provided in Appendix~\ref{sec:appendix_main_kvcache} and Table~\ref{tab:kv_cache_results}.

\paragraph{\textbf{\emph{Observation 1}}}
\textbf{Token compression performance is task-dependent and shows significant sensitivity to benchmark and model.}
Most methods are stable at higher budgets but degrade sharply at 1\%, especially on tasks requiring fine-grained visual detail (e.g., OCRBench) or long outputs (e.g., LLaVA-Wilder, ImageDC).
For token pruning on a 1\% budget, pruning tokens within the visual encoder (e.g., VisionZip and PruMerge+) consistently outperforms those pruning in the LLM backbone (e.g., FastV). On LLaVA-OV-7B, FastV's relative performance plummets to 48\%, whereas VisionZip retains 75\% (Further analysis in Section~\ref{sec:sink_token}.).
For KV cache compression, retaining sufficient tokens is crucial for preserving fine-grained information, but sensitivity varies across architectures and budgets. Notably, LLaVA-OV-7B processes thousands of tokens, so even at a 1\% budget, it retains more tokens (40.17) than Qwen2-VL-7B (7.62) and InternVL2.5-38B (15.14), leading to a smaller performance drop. These findings underscore the need to tailor compression methods to model and task specifics, and to rigorously assess their cross-benchmark generalization.

\begin{figure*}[t!]
\centering
\includegraphics[width=1.0\textwidth]{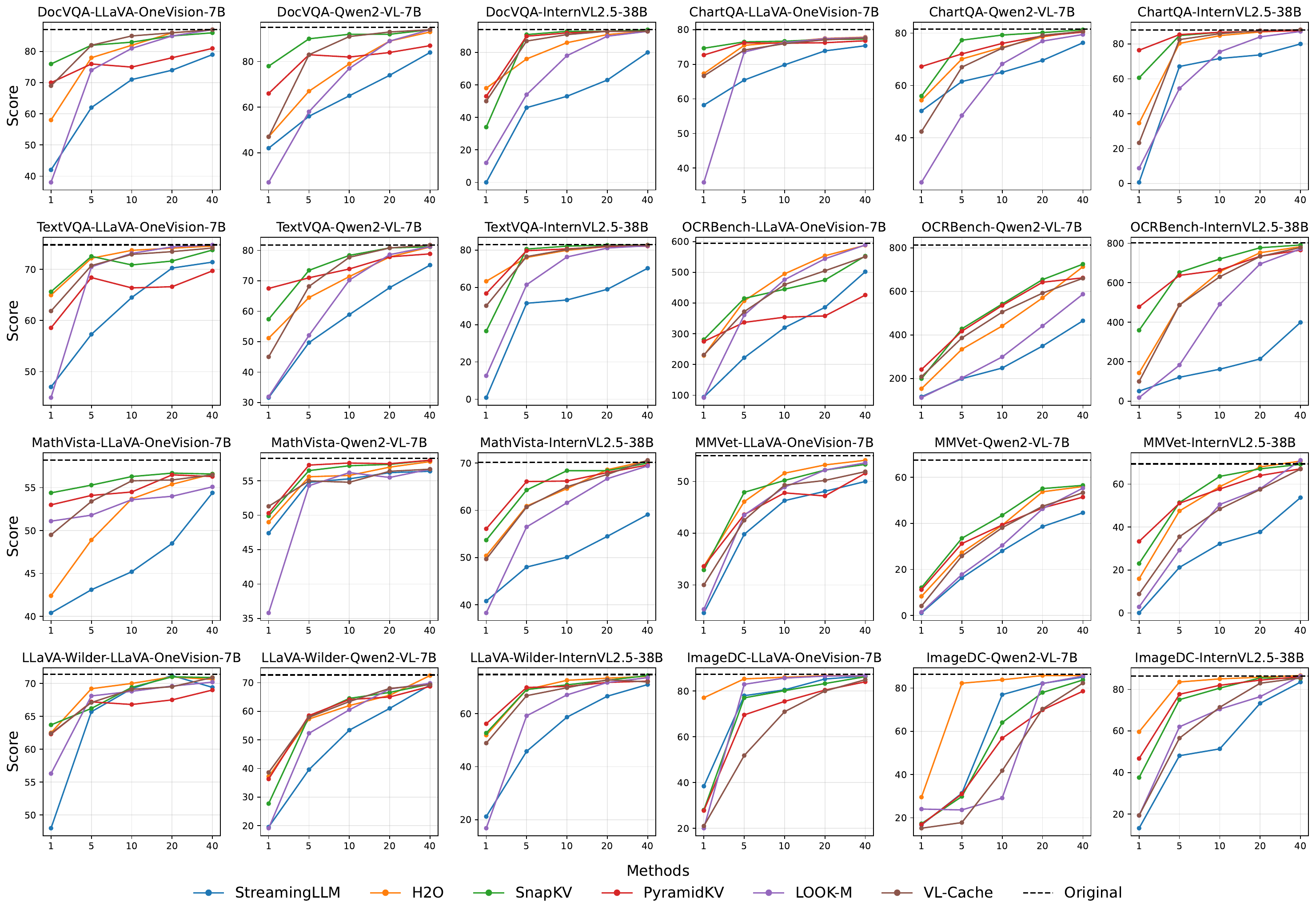}
\caption{Performance comparison of KV cache compression methods across multiple benchmarks and models.
}\label{fig:main_kvcache}
\end{figure*}

\paragraph{\textbf{\emph{Observation 2}}} 
\textbf{KV cache compression outperforms token pruning in generalization and loyalty.}
Table~\ref{tab:stab_and_loy} shows H2O and PyramidKV leading in overall performance across metrics at high (40\%) and low (10\%) budgets, respectively.
KV cache compression methods generally exhibit superior generalization and loyalty compared to token pruning. Thus, they should be prioritized when these aspects are critical.
Meanwhile, we observe that LOOK-M performs well at higher budgets, but it exhibits a significant drop at lower budgets, which aligns with results in Figure~\ref{fig:main_kvcache}, highlighting its high sensitivity under extremely aggressive constraints(detailed analysis is presented in Section~\ref{sec:merge}).

\begin{table}[t!]
    \centering
    \small 
    \resizebox{1.0\linewidth}{!}{
    \begin{tabular}	{l | cc cc}
        \toprule
        \multirow{2}{*}{\bf Methods} & \multicolumn{2}{c}{$\text{OG}^{c} \downarrow$} & \multicolumn{2}{c}{$\text{OL}^{c}\uparrow$} \\
        & 1\% & 40\% & 1\% & 40\%\\
        \midrule
        StreamingLLM & 49.76 & 12.05 & 33.26& 85.98\\
        H2O & 30.93 & \textbf{2.05} &53.02 &\bf 94.57 \\
       SnapKV & 31.98 & 2.34 &55.85 & 93.31\\
       PyramidKV & \textbf{29.76} & 5.31 &\bf 59.52 &91.19 \\
       LOOK-M & 51.72 & 3.46 &29.67 & 93.43\\
       VL-Cache & 43.25 & 3.80 & 55.00& 92.04\\
       FastV & 58.46 & 11.87 &40.89 &80.57 \\
       VisionZip & 31.32& 5.86 & 59.24& 80.52\\
       PruMerge+ & 34.62& 13.47& 56.85&83.07 \\
        \bottomrule        
    \end{tabular}
    }
    \caption{Overall performance of generalization and loyalty for different token compression methods, with generalization calculated across all benchmarks and loyalty on a subset(MathVista,LLaVA-Wilder,MMVet). \textbf{Bold} indicates the best under the same setting.}
    \label{tab:stab_and_loy}
\end{table}

\begin{figure*}[t!]
\centering
\includegraphics[width=0.9\textwidth]{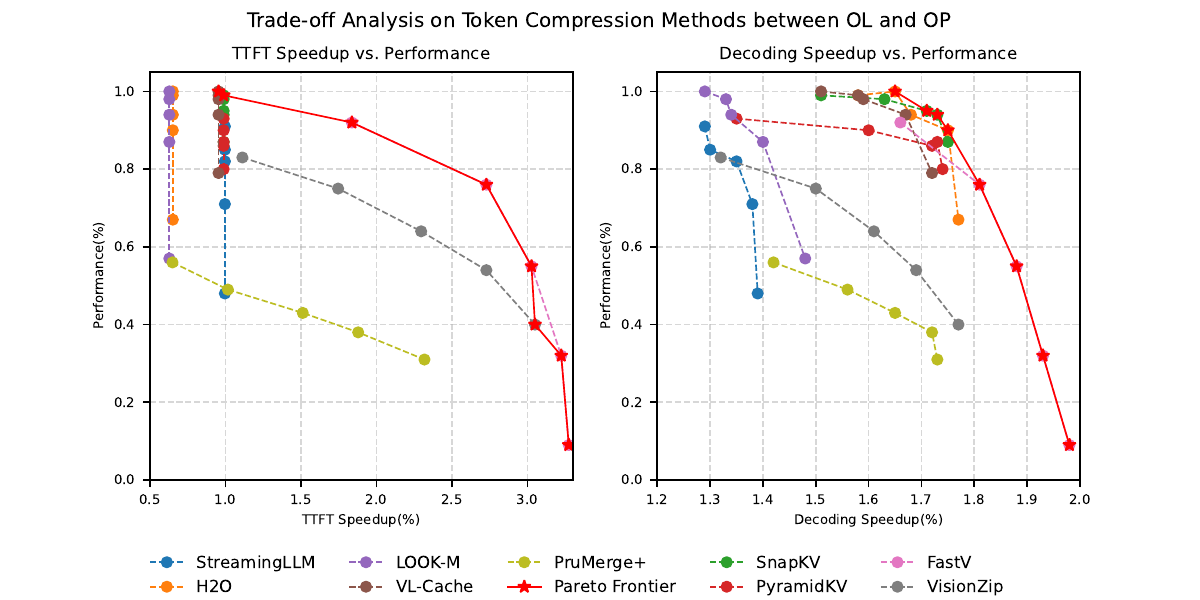}
\caption{Trade‐off Analysis on Token Compression Methods between Efficiency and performance. \textbf{Left:} Trade‐off between TTFT (time‐to‐first‐token) and Performance across different methods. \textbf{Right:} Trade‐off between Speedup and performance across different methods. All metrics are expressed as ratios relative to the original model.
}\label{fig:pareto}
\end{figure*}

\paragraph{\textbf{\emph{Observation 3}}} 
\textbf{Selecting token pruning or KV cache compression based on task statistics to achieve better performance-efficiency trade-off.}
Actual inference time can be broken down into two components: time-to-first-token (TTFT), which reflects the prefill overhead, and decoding time, which measures the latency of subsequent token generation.
Figure~\ref{fig:pareto} illustrates the trade-offs between efficiency and performance for token compression methods. 
KV cache methods yield limited TTFT speedup as they recompute attention weights for token selection during prefill\footnote{When using mechanisms like Flash-Attention2~\citep{dao2023flashattention2} that don't provide full attention matrices for selection, recalculation becomes necessary.}. 
This computation overhead is budget-independent, keeping their TTFT speedup nearly constant (see vertical lines in left part of Figure~\ref{fig:pareto}).
In contrast, token pruning removes visual tokens during prefill, drastically reducing TTFT (e.g., 3.2× speedup at 1\% budget), making it ideal for short-response tasks like VQA. For instance, at comparable overall performance, H2O achieves only 0.65× TTFT speedup, while VisionZip delivers 2.29× TTFT speedup.
For decoding latency (right), KV cache and token pruning methods show similar speedups under the same budget. However, KV cache methods generally outperform token pruning at low budgets for tasks requiring long outputs (e.g., LLaVA-Wilder, ImageDC). Conversely, for tasks involving high-resolution images, the fixed cost of computing attention matrices in KV cache compression leads to significant memory overhead, favoring token pruning in these scenarios.

\paragraph{\textbf{\emph{Observation 4}}}
\textbf{Consistent performance trends of token compression across single-image, multi-images, and video tasks.}
We further evaluate token compression methods on multi-image and video tasks.
Table~\ref{tab:multi_image} shows the results based on Qwen2-VL-7B.
Under a 1\% budget, SnapKV achieves the best performance in MP-DocVQA, while LOOK-M suffers a significant drop, consistent with the results in DocVQA. 
In MovieChat, VL-Cache outperforms other methods.
Notably, at higher budgets, some compression methods, such as SnapKV and VL-Cache, can maintain the original model's performance while achieving substantial speedups.
These findings indicate that token compression remains consistently effective across single-image, multi-image, and video scenarios. Future work will explore more challenging tasks for further validation.

\begin{table}[t]
\centering
\resizebox{1.0\linewidth}{!}{
\begin{tabular}{l|l|cccc}
\toprule

\textbf{Benchmarks} & \textbf{Methods} & 1\% & 10\% & 40\% & 100\% \\
\midrule
\multirow{6}{*}{\shortstack{MP-DocVQA\\(Split:val)\\(Metric:ANLS)}}

& StreamingLLM  & 36.74 & 51.38 & 69.04 & 85.91\\
& H2O           & 38.73 & 63.02 & 80.67 & 85.91\\
& SnapKV        & \textbf{69.39} & 80.73 & 84.11 & 85.91\\
& PyramidKV     & 63.44 & 75.93 & 79.33 & 85.91\\
& LOOK-M        & 18.38 & 57.09 & 80.69 & 85.91\\
& VL-Cache      & 57.89 & \textbf{82.44} & \textbf{85.46} & 85.91\\
\midrule

\multirow{6}{*}{\shortstack{MovieChat\\(Split:test)\\(Metric:Acc)}}

& StreamingLLM  & 13.11 & 34.63 & 37.03 & 37.73\\
& H2O           & 14.23 & \textbf{37.15} & 37.19 & 37.73\\
& SnapKV        & 28.29 & 36.36 & 37.53 & 37.73\\
& PyramidKV     & 28.26 & \textbf{37.15} & 36.62 & 37.73\\
& LOOK-M        & 11.95 & 36.49 & 37.57 & 37.73\\
& VL-Cache      & \textbf{28.51} & 35.99 & \textbf{37.65} & 37.73\\

\bottomrule
\end{tabular}
}
\caption{Results of token compression methods on Qwen2-VL-7B evaluated with multi-image and video benchmarks.}
\label{tab:multi_image}
\end{table}

\subsection{Parameter Compression}
\paragraph{Setup}
Pruning and quantization are two mainstream approaches for compressing LVLMs.
We evaluate three pruning methods: EcoFLAP~\citep{sung2023ecoflap}, Wanda~\citep{sun2023simple}, and SparseGPT~\citep{frantar2023sparsegpt}; and two quantization methods: AWQ~\citep{lin2024awq} and GPTQ~\citep{frantar2022gptq}. EcoFLAP is specifically designed for LVLMs, while the others are widely-used for general LLM compression.
Details are shown in Appendix~\ref{sec:appendix_method} and~\ref{sec:appendix_math_modeling}.

\paragraph{\textbf{\emph{Observation 5}}}
\textbf{Parameter compression generally preserve performance more effectively than token compression.}
Even at higher compression ratios (50\% or 2:4 sparsity), the overall performance remains relatively stable. We observe that quantization methods, such as AWQ, tend to preserve higher performance compared to pruning. However, since neither method shortens input token sequences, token compression remains essential for tasks involving very long inputs or high-resolution images.
Importantly, the two types of compression are orthogonal that can be effectively combined, 
Crucially, these two compression types are orthogonal and can be effectively combined and more results are shown in Appendix~\ref{sec:app_more_params_compression}.

\begin{table*}[t!]
\centering
\small
\resizebox{\linewidth}{!}{
\begin{tabular}{c|l|*{15}{c}|c}
\toprule
\raisebox{5ex}{\textbf{Settings}} & \raisebox{5ex}{\textbf{Methods}} & 
\rotatebox{90}{DocVQA} & \rotatebox{90}{ChartQA} & \rotatebox{90}{TextVQA} & 
\rotatebox{90}{OCRBench} & \rotatebox{90}{AI2D} & \rotatebox{90}{GQA} & 
\rotatebox{90}{MMMU} & \rotatebox{90}{MME} & \rotatebox{90}{RealworldQA} & 
\rotatebox{90}{MMStar} & \rotatebox{90}{MathVista} & \rotatebox{90}{LLaVA-Wilder} & 
\rotatebox{90}{MMBench} & \rotatebox{90}{MMVet} & \rotatebox{90}{ImageDC} & 
\raisebox{5ex}{\textbf{OP}}\\
\midrule
\multirow{1}{*}{100\%} 
  & Original    & 87   & 80.00  & 74.79  & 595  & 89.96  & 61.92  & 45.44  & 1974.1  & 65.88  & 58.75  & 58.20  & 71.40  & 83.12  & 55.00  & 87.25 & 1.000 \\
\midrule
\multirow{3}{*}{20\%} 
  & EcoFLAP    & \textbf{87}   & \textbf{79.32}  & 74.70  & 573  & \textbf{89.86}  & 61.26  & 46.56  & 1951.0  & 64.84  & \textbf{60.27}  & 55.80  & 70.20  & 82.17  & 57.70  & 85.75 & \textbf{0.995} \\
  & Wanda      & 86   & 78.84  & \textbf{75.70}  & 595  & 81.12  & 61.71  & \textbf{46.67}  & 1939.8  & \textbf{66.14}  & 59.11  & 54.10  & 68.60  & 82.17  & \textbf{60.10}  & \textbf{86.90} & 0.992 \\
  & SparseGPT  & \textbf{87}   & 77.48  & 73.92  & \textbf{598}  & 80.67  & \textbf{61.95}  & 45.78  & \textbf{1957.2}  & \textbf{66.14}  & 58.28  & \textbf{56.30}  & \textbf{70.30}  & \textbf{83.18}  & 55.40  & 86.10 & 0.986 \\
\midrule
\multirow{3}{*}{50\%} 
  & EcoFLAP    & \textbf{82}   & 73.24  & 71.06  & 502  & 75.78  & 59.81  & 41.56  & 1732.8  & 63.40  & 54.23  & \textbf{50.20}  & 53.80  & 78.69  & 39.80  & 57.55 & 0.877 \\
  & Wanda      & 81   & \textbf{76.96}  & \textbf{71.30}  & \textbf{582}  & 72.12  & 60.10  & 41.67  & 1726.8  & \textbf{64.18}  & \textbf{54.41}  & 48.50  & \textbf{65.20}  & 42.26  & 47.20  & \textbf{85.50} & 0.900 \\
  & SparseGPT  & 81   & 72.08  & 70.20  & 534  & \textbf{77.66}  & \textbf{61.16}  & \textbf{42.89}  & \textbf{1741.1}  & 66.01  & 53.05  & 49.30  & 64.90  & \textbf{79.82}  & \textbf{47.30}  & 85.00 & \textbf{0.921} \\
\midrule
\multirow{3}{*}{2:4} 
  & EcoFLAP    & 67   & 62.02  & 60.70  & 483  & 65.19  & 52.68  & 32.89  & \bf 1309.1  & 49.54  & 45.75  & 36.00  & 49.30  & 71.35  & 30.20  & 81.05 & 0.760 \\
  & Wanda      & 69   & \textbf{63.52}  & 59.71  & \textbf{499}  & \textbf{67.39}  & 53.09  & 34.22  & 1282.7  & 57.52  & 43.09  & \textbf{38.80}  & 51.60  & \textbf{72.81}  & 30.60  & \textbf{81.65} & \textbf{0.779} \\
  & SparseGPT  & \textbf{74}   & 62.68  & \textbf{65.60}  & 426  & 67.10  & \textbf{56.97}  & \textbf{34.44}  & 1296.2  & \textbf{62.35}  & \textbf{45.99}  & 35.60  & \textbf{52.90}  & 35.59  & \textbf{35.70}  & 81.30 & 0.772 \\
\midrule
\multirow{2}{*}{W4A16$^\dagger$} 
  & AWQ        & 84   & 75.52  & \textbf{72.78}  & \textbf{575}  & 85.23  & 61.84  & \textbf{45.44}  & \textbf{1978.3}  & \textbf{67.58}  & 53.70  & 52.60  & \textbf{70.50}  & 78.92  & \bf 54.80  & \textbf{87.05} & 0.972 \\
  & GPTQ       & \textbf{86}   & \textbf{75.83}  & 72.36  & 571  & \textbf{85.40}  & \textbf{61.90}  & 44.82  & 1970.2  & 66.24  & \textbf{56.25}  & \textbf{55.80}  & 70.30  & \textbf{78.69}  & 54.30  & 86.55 & \textbf{0.975} \\
\bottomrule
\end{tabular}
}
\caption{Main results of various parameter compression methods on LLaVA-OneVision-7B evaluated on different tasks, grouped by \emph{Setting}. $^\dagger$ indicates that only the LLM backbone is quantized. 
}
\label{tab:model_compression}
\end{table*}

\section{Discussion}
\label{sec:analysis}

\subsection{Revisiting Layer Adaptive Mechanism}
\label{sec:analysis_layer_adaptive}

\begin{figure}[t!]
\centering
\includegraphics[width=\columnwidth, keepaspectratio]{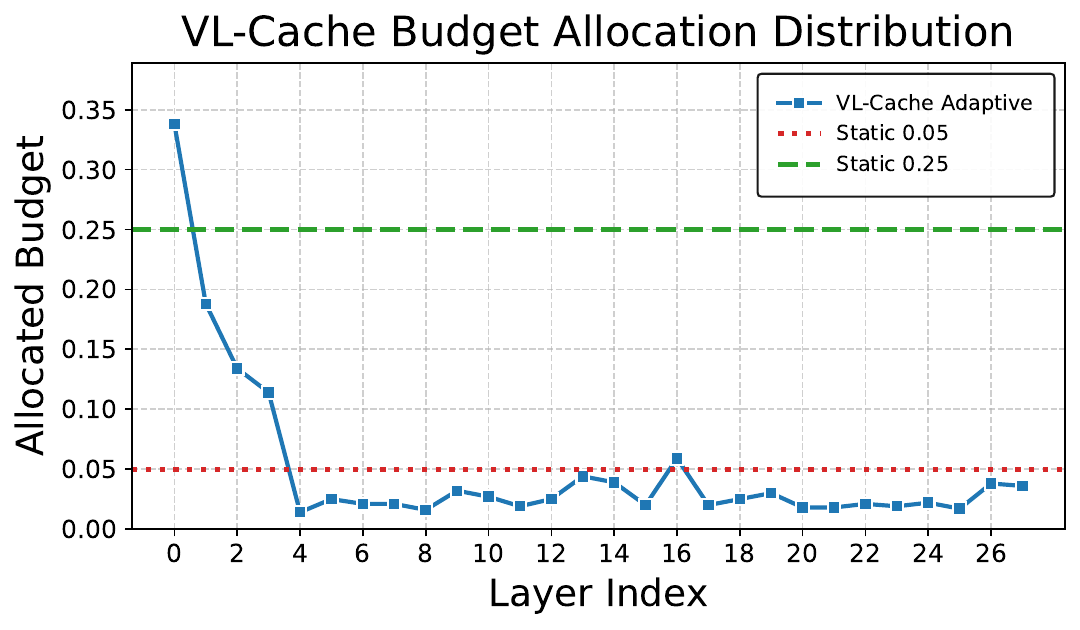}
\caption{Layer-wise Budget Allocation of VL-Cache on LLaVA-OneVision-7B (5\% Total Budget Constraint, OCRBench Example).
}\label{fig:main_layer_budget}
\end{figure}

While \citet{cai2024pyramidkv} shows that layer-adaptive sparsity benefits KV cache compression in LLM, \ourbench results reveal that PyramidKV underperforms SnapKV at low sparsity budgets despite using the same token selection metric, due to its layer-adaptive strategy.
This suggests that \textbf{layer-adaptive sparsity is not always advantageous in LVLM compression, especially under low sparsity}.
To further probe this, we analyze VL-Cache, a layer-adaptive KV cache compression method proposed for LVLMs. 
As shown in Figure~\ref{fig:main_layer_budget} and Appendix~\ref{sec:additional-figures-vlcache-budget}:
At 5\% budget, the $0$-$th$ layer has nearly 7$\times$ of the average budget. 
This aggressive front-loading reduces budgets for subsequent layers, effectively starving them (less than the average).
In addition, visualizations of the tokens selected in the $0$-$th$ layer (shown in Figure~\ref{fig:vlcache_case_row1} and Appendix~\ref{sec:visualization-vlcache-visual-tokens}) reveal that most of the tokens chosen are irrelevant.
These findings suggest that reducing the budget of early layers while increasing the average allocation for later layers may lead to better performance. Based on the VL-Cache layer adaptive strategy, we propose a hybrid allocation strategy: a portion (40\% or 80\%) of the total budget was evenly distributed across all layers, and the remaining (60\% or 20\%) is adaptively allocated according to the original allocation strategy. As shown in Table~\ref{tab:add_base_vlcache}, evenly allocating 80\% of the total budget produced the best results.

\begin{table}[t]
\centering
\resizebox{1.0\linewidth}{!}{
\small
\begin{tabular}{@{}l|l|c c c@{}}
\toprule
\textbf{Models} & \textbf{Benchamrks} & A-Only & U-40\% & U-80\% \\
\midrule
\multirow{4}{*}{\shortstack{LLaVA-OneVision-7B}} 
& DocVQA    & 82   & 83   & \textbf{84} \\
& OCRBench  & 372  & 398  & \textbf{410} \\
& ChartQA   & 74.12 & 75.84 & \textbf{76.24} \\
& TextVQA   & 70.73 & 71.83 & \textbf{72.36} \\

\midrule
\multirow{4}{*}{\shortstack{Qwen2-VL-7B}} 
& DocVQA    & 83   & 85   & \textbf{86} \\
& OCRBench  & 386  & 410  & \textbf{424} \\
& ChartQA   & 67.00 & 67.84 & \textbf{68.24} \\
& TextVQA   & 68.17 & 69.93 & \textbf{70.98} \\

\bottomrule
\end{tabular}
}
\caption{VL-Cache Budget Allocation (5\% Total Budget) with Hybrid Strategies: A-Only (adaptive-Only Allocation) vs. U-40\% (40\% uniform + 60\% adaptive) vs. U-80\% (80\% uniform + 20\% adaptive).}
\label{tab:add_base_vlcache}
\end{table}


\subsection{Revisiting Head Adaptive Mechanism}
Previous KV cache compression methods diverge on whether heads within a layer should select identical cached tokens.
We explore head-adaptive token selection (i.e., allowing heads in the same layer to select different tokens) at a 1\% budget.
Table~\ref{tab:ablation_head_adaptive} shows that head-adaptivity significantly improves performance. This improvement likely stems from different KV or query heads capturing distinct information patterns, enabling head-adaptive selection to preserve critical information under tight budget constraints better. Detailed head attention distributions are in Appendix~\ref{sec:appendix_head_attention_distribution}.

\begin{table}[htbp]
\centering
\scriptsize
\renewcommand{\arraystretch}{0.9}
\setlength{\tabcolsep}{2pt}

\begin{adjustbox}{max width=\linewidth}  
\begin{tabular}{@{}l l *{3}{cc}@{}}
\toprule
\multirow{2}{*}{\textbf{Models}}
  & \multirow{2}{*}{\textbf{Benchmarks}} 
  & \multicolumn{2}{c}{\textbf{H2O}} 
  & \multicolumn{2}{c}{\textbf{SnapKV}} 
  & \multicolumn{2}{c}{\textbf{VL-Cache}} \\
\cmidrule(lr){3-4}\cmidrule(lr){5-6}\cmidrule(lr){7-8}
& & w/ & w/o
  & w/ & w/o
  & w/ & w/o \\
\midrule
\multirow{4}{*}{LLaVA-OneVision-7B}
 & ChartQA      
   & \textbf{67.32} & 65.12 
   & \textbf{74.64} & 73.84 
   & \textbf{68.64} & 66.64 \\
 & TextVQA      
   & \textbf{64.96} & 62.33 
   & \textbf{65.62} & 65.60 
   & 61.76 & \textbf{61.85} \\
 & MathVista    
   & \textbf{42.40}  & 41.60  
   & \textbf{54.40}  & 51.20  
   & \textbf{52.20}  & 49.50  \\
 & LLaVA-Wilder 
   & \textbf{62.50}  & 60.60 
   & \textbf{63.70}  & 61.40  
   & \textbf{62.80}  & 62.40  \\
\midrule

\multirow{4}{*}{Qwen2-VL-7B}
 & ChartQA      
   & \textbf{54.36} & 45.92 
   & 55.96 & \textbf{56.16} 
   & \textbf{57.76} & 53.32 \\
 & TextVQA      
   & \textbf{51.16} & 42.85 
   & \textbf{57.38} & 52.96 
   & \textbf{50.69} & 47.03 \\
 & MathVista    
   & \textbf{49.00}  & 48.50  
   & \textbf{49.90}  & 49.20  
   & \textbf{52.90}  & 52.30  \\
 & LLaVA-Wilder 
   & \textbf{37.20}  & 31.60  
   & \textbf{27.70}  & 26.60  
   & \textbf{42.30}  & 39.80  \\

\bottomrule
\end{tabular}
\end{adjustbox}
\caption{Effect of head-adapive strategy for KV cache compression methods.}
\label{tab:ablation_head_adaptive}
\end{table}

\subsection{Attention Sink Tokens in LVLMs}
\label{sec:sink_token}

\citet{xiao2023streamingllm} identified \textit{attention sinks} in Transformers: a few initial tokens attract disproportionate attention regardless of semantic relevance, and their removal significantly degrades performance.
We explore sink tokens in both textual and visual modalities within LVLMs.
For textual sinks, comparing a random baseline against StreamingLLM (which preserves these sink tokens) shows StreamingLLM significantly improves performance at a 1\% budget (Table~\ref{tab:kv_cache_results}), underscoring their importance.
Similarly, visual information tends to concentrate on a set of image tokens after passing through the visual encoder~\citet{darcet2024vision}.
We find that FastV (using text-guided metrics for visual token pruning) underperforms VisionZip (relying solely on the visual encoder's attention map). We hypothesize this is because visual sink tokens critically impact performance, yet text-guided metrics often fail to capture them.
Two ablations on ChartQA using Qwen2-VL-7B (Table~\ref{tab:visionzip_exp}) further support this hypothesis: \textbf{A1}: Prohibiting FastV from selecting from the top 10\% most critical visual tokens slightly degraded performance. \textbf{A2}: Conversely, forcing FastV to prioritize these top 10\% tokens significantly improved performance.
These results indicate that text-based selection can overlook critical visual sink tokens, and that retaining these sinks is vital, particularly under low budgets.
The recent attention-sink-free gated attention model of \citet{qiu2025gated} also warrants future exploration.
More visualizations are available in Appendix~\ref{sec:appendix_visual_sink}.

\begin{table}[t!]
\centering
\small
\setlength{\tabcolsep}{3pt}
\begin{tabularx}{\linewidth}{@{} l | *{3}{>{\centering\arraybackslash}X} @{}}
\toprule
\multirow{2}{*}{\textbf{Methods}} & \multicolumn{3}{c}{\textbf{Budgets}} \\
\cmidrule{2-4}
 & 10\% & 20\% & 40\% \\
\midrule
FastV\textsubscript{origin} & 31.04 & 54.20 & 67.40 \\
FastV\textsubscript{A1} & 30.36 & 52.40 & 65.40 \\
FastV\textsubscript{A2} & \bfseries 45.56 & \bfseries 58.60 & \bfseries 70.04 \\
\bottomrule
\end{tabularx}
\caption{Comparison of results for FastV\textsubscript{origin}, FastV\textsubscript{A1}, and FastV\textsubscript{A2} on ChartQA with Qwen2-VL-7B. 
}
\label{tab:visionzip_exp}
\end{table}

\subsection{How to Merge Evicted Tokens}
\label{sec:merge}
The merge operation shows promise for \textit{recovering} evicted information~\citep{bolyatoken}.
Current methods differ: LOOK-M and PruMerge+ merge evicted tokens \textit{into} retained ones, whereas VisionZip \textit{concatenates} them.
This poses the question: \textbf{What is an effective merge strategy for LVLMs?}
Our experiments reveal LOOK-M's performance drops at low budgets. We hypothesize that at a 1\% budget, its text-prior mechanism discards visual tokens, which are then merged into remaining text tokens. This cross-modal fusion, we argue, disrupts critical textual features (e.g., sink tokens), degrading performance..
Thus, we modify LOOK-M to merge evicted tokens only within the same modality. Table \ref{tab:lookm_merge_exp} shows our modification consistently improves performance over the original LOOK-M, underscoring that modality-specific merging is crucial for LVLM token compression. Specific examples are in Appendix~\ref{sec:appedix_merge_dialog}.
\begin{table}[t!]
\centering
\small
\resizebox{\linewidth}{!}{%
\begin{tabular}{c|l|c*{6}{>{\centering\arraybackslash}c}}
\toprule
\raisebox{5ex}{\textbf{Budgets}} & 
\raisebox{5ex}{\diagbox{\textbf{Methods}}{\textbf{Benchmarks}}}
& 
\raisebox{1ex}{\adjustbox{valign=b}{\rotatebox[origin=t]{90}{DocVQA}}} & 
\raisebox{1ex}{\adjustbox{valign=b}{\rotatebox[origin=t]{90}{ChartQA}}} & 
\raisebox{1ex}{\adjustbox{valign=b}{\rotatebox[origin=t]{90}{TextVQA}}} & 
\raisebox{1ex}{\adjustbox{valign=b}{\rotatebox[origin=t]{90}{OCRBench}}} & 
\raisebox{1ex}{\adjustbox{valign=b}{\rotatebox[origin=t]{90}{MathVista}}} & 
\raisebox{1ex}{\adjustbox{valign=b}{\rotatebox[origin=t]{90}{LLaVA-Wilder}}} \\
\midrule

\multirow{2}{*}{1\%} 
      & LOOK-M\textsubscript{origin}   & 38 & 34.44 & 44.74 & 81 & 44.30 & 43.40 \\
   & LOOK-M\textsubscript{change}  & \bf44 & \bf56.16 & \bf48.12 & \bf117 & \bf51.80 & \bf59.30 \\
\cmidrule(lr){1-8}
\multirow{2}{*}{5\%} 
      & LOOK-M\textsubscript{origin}   & \bf74 & 73.72 & 70.29 & 375 & 53.10 & 65.50 \\
   & LOOK-M\textsubscript{change}  & \bf74 & \bf75.04 & \bf70.89 & \bf406 & \bf53.50 & \bf69.00 \\

\bottomrule
\end{tabular}
}
\caption{The performance comparison between LOOK-M with origin merge(LOOK-M\textsubscript{origin}) and modality-specific merge(LOOK-M\textsubscript{change}) on LLaVA-OV-7B.}
\label{tab:lookm_merge_exp}
\end{table}

\section{Conclusion}
This paper introduces \ourbench, a comprehensive benchmark for systematically evaluating token and parameter compression methods for LVLMs across diverse tasks, models, and metrics. Our empirical results reveal inherent trade-offs—performance, generalization, loyalty, and efficiency—tied to different compression strategies. Analysis of factors such as attention sinks and layer- or head-adaptive sparsity provides practical insights for optimizing these techniques. We hope that \ourbench offers a robust foundation for advancing future research in LVLM compression.

\section*{Limitations}
Our work has several limitations: (1) Although \ourbench systematically evaluates both token and parameter compression approaches, it focuses primarily on a subset of representative LVLM models and tasks, leaving the performance on other architectures and more specialized domains underexplored.
(2) We only consider training-free methods in this study, and the incorporation of training-based compression algorithms could provide deeper insights into the balance between performance gains and resource overhead. (3) Although our analysis delves into several mechanisms under extremely low budgets, there may be additional factors in token and parameter compression methods. In future work, we plan to extend \ourbench to include more tasks, models, and methods, offering a more comprehensive assessment of LVLM compression.

\bibliography{custom}

\begin{thebibliography}{75}
\providecommand{\natexlab}[1]{#1}

\bibitem[{An et~al.(2024)An, Zhao, Yu, Tang, and Wang}]{an2024flap}
Yongqi An, Xu~Zhao, Tao Yu, Ming Tang, and Jinqiao Wang. 2024.
\newblock \href {https://doi.org/10.1609/aaai.v38i10.28960} {Fluctuation-based adaptive structured pruning for large language models}.
\newblock In \emph{Thirty-Eighth {AAAI} Conference on Artificial Intelligence, {AAAI} 2024, Thirty-Sixth Conference on Innovative Applications of Artificial Intelligence, {IAAI} 2024, Fourteenth Symposium on Educational Advances in Artificial Intelligence, {EAAI} 2014, February 20-27, 2024, Vancouver, Canada}, pages 10865--10873. {AAAI} Press.

\bibitem[{Anthropic(2024)}]{claude-computer-use}
Anthropic. 2024.
\newblock \href {https://www.anthropic.com/news/developing-computer-use} {Developing a computer use model}.

\bibitem[{Bolya et~al.()Bolya, Fu, Dai, Zhang, Feichtenhofer, and Hoffman}]{bolyatoken}
Daniel Bolya, Cheng-Yang Fu, Xiaoliang Dai, Peizhao Zhang, Christoph Feichtenhofer, and Judy Hoffman.
\newblock Token merging: Your vit but faster.
\newblock In \emph{The Eleventh International Conference on Learning Representations}.

\bibitem[{Cai et~al.(2024{\natexlab{a}})Cai, Zhang, He, He, Tong, Gan, Wang, and Bai}]{cai2024llavakd}
Yuxuan Cai, Jiangning Zhang, Haoyang He, Xinwei He, Ao~Tong, Zhenye Gan, Chengjie Wang, and Xiang Bai. 2024{\natexlab{a}}.
\newblock \href {https://doi.org/10.48550/ARXIV.2410.16236} {Llava-kd: {A} framework of distilling multimodal large language models}.
\newblock \emph{CoRR}, abs/2410.16236.

\bibitem[{Cai et~al.(2024{\natexlab{b}})Cai, Zhang, Gao, Liu, Liu, Lu, Xiong, Dong, Chang, Hu et~al.}]{cai2024pyramidkv}
Zefan Cai, Yichi Zhang, Bofei Gao, Yuliang Liu, Tianyu Liu, Keming Lu, Wayne Xiong, Yue Dong, Baobao Chang, Junjie Hu, and 1 others. 2024{\natexlab{b}}.
\newblock Pyramidkv: Dynamic kv cache compression based on pyramidal information funneling.
\newblock \emph{arXiv preprint arXiv:2406.02069}.

\bibitem[{Cha et~al.(2024)Cha, Kang, Mun, and Roh}]{cha2024honeybee}
Junbum Cha, Wooyoung Kang, Jonghwan Mun, and Byungseok Roh. 2024.
\newblock Honeybee: Locality-enhanced projector for multimodal {LLM}.
\newblock In \emph{{IEEE/CVF} Conference on Computer Vision and Pattern Recognition, {CVPR} 2024, Seattle, WA, USA, June 16-22, 2024}, pages 13817--13827. {IEEE}.

\bibitem[{Chen et~al.(2024{\natexlab{a}})Chen, Chen, Zhang, Chen, Wu, Zhang, Chen, Li, Wan, and Wang}]{chen2024allava}
Guiming~Hardy Chen, Shunian Chen, Ruifei Zhang, Junying Chen, Xiangbo Wu, Zhiyi Zhang, Zhihong Chen, Jianquan Li, Xiang Wan, and Benyou Wang. 2024{\natexlab{a}}.
\newblock Allava: Harnessing gpt4v-synthesized data for {A} lite vision-language model.
\newblock \emph{CoRR}, abs/2402.11684.

\bibitem[{Chen et~al.(2024{\natexlab{b}})Chen, Zhao, Liu, Bai, Lin, Zhou, and Chang}]{chen2024fastv}
Liang Chen, Haozhe Zhao, Tianyu Liu, Shuai Bai, Junyang Lin, Chang Zhou, and Baobao Chang. 2024{\natexlab{b}}.
\newblock An image is worth 1/2 tokens after layer 2: Plug-and-play inference acceleration for large vision-language models.
\newblock In \emph{European Conference on Computer Vision}, pages 19--35. Springer.

\bibitem[{Chen et~al.(2024{\natexlab{c}})Chen, Li, Dong, Zhang, Zang, Chen, Duan, Wang, Qiao, Lin et~al.}]{chen2024wemmstar}
Lin Chen, Jinsong Li, Xiaoyi Dong, Pan Zhang, Yuhang Zang, Zehui Chen, Haodong Duan, Jiaqi Wang, Yu~Qiao, Dahua Lin, and 1 others. 2024{\natexlab{c}}.
\newblock Are we on the right way for evaluating large vision-language models?
\newblock \emph{arXiv preprint arXiv:2403.20330}.

\bibitem[{Chen et~al.(2015)Chen, Fang, Lin, Vedantam, Gupta, Doll{\'a}r, and Zitnick}]{Chen2015MicrosoftCC}
Xinlei Chen, Hao Fang, Tsung-Yi Lin, Ramakrishna Vedantam, Saurabh Gupta, Piotr Doll{\'a}r, and C.~Lawrence Zitnick. 2015.
\newblock \href {https://api.semanticscholar.org/CorpusID:2210455} {Microsoft coco captions: Data collection and evaluation server}.
\newblock \emph{ArXiv}, abs/1504.00325.

\bibitem[{Chen et~al.(2024{\natexlab{d}})Chen, Wang, Cao, Liu, Gao, Cui, Zhu, Ye, Tian, Liu et~al.}]{chen2024expanding}
Zhe Chen, Weiyun Wang, Yue Cao, Yangzhou Liu, Zhangwei Gao, Erfei Cui, Jinguo Zhu, Shenglong Ye, Hao Tian, Zhaoyang Liu, and 1 others. 2024{\natexlab{d}}.
\newblock Expanding performance boundaries of open-source multimodal models with model, data, and test-time scaling.
\newblock \emph{arXiv preprint arXiv:2412.05271}.

\bibitem[{Chu et~al.(2023)Chu, Qiao, Lin, Xu, Yang, Hu, Wei, Zhang, Zhang, Wei et~al.}]{chu2023mobilevlm}
Xiangxiang Chu, Limeng Qiao, Xinyang Lin, Shuang Xu, Yang Yang, Yiming Hu, Fei Wei, Xinyu Zhang, Bo~Zhang, Xiaolin Wei, and 1 others. 2023.
\newblock Mobilevlm: A fast, reproducible and strong vision language assistant for mobile devices.
\newblock \emph{arXiv preprint arXiv:2312.16886}.

\bibitem[{Dao(2024)}]{dao2023flashattention2}
Tri Dao. 2024.
\newblock Flash{A}ttention-2: Faster attention with better parallelism and work partitioning.
\newblock In \emph{International Conference on Learning Representations (ICLR)}.

\bibitem[{Darcet et~al.(2024)Darcet, Oquab, Mairal, and Bojanowski}]{darcet2024vision}
Timoth{\'{e}}e Darcet, Maxime Oquab, Julien Mairal, and Piotr Bojanowski. 2024.
\newblock Vision transformers need registers.
\newblock In \emph{The Twelfth International Conference on Learning Representations, {ICLR} 2024, Vienna, Austria, May 7-11, 2024}. OpenReview.net.

\bibitem[{Dehghani et~al.(2023)Dehghani, Mustafa, Djolonga, Heek, Minderer, Caron, Steiner, Puigcerver, Geirhos, Alabdulmohsin et~al.}]{dehghani2023patch}
Mostafa Dehghani, Basil Mustafa, Josip Djolonga, Jonathan Heek, Matthias Minderer, Mathilde Caron, Andreas Steiner, Joan Puigcerver, Robert Geirhos, Ibrahim~M Alabdulmohsin, and 1 others. 2023.
\newblock Patch n’pack: Navit, a vision transformer for any aspect ratio and resolution.
\newblock \emph{Advances in Neural Information Processing Systems}, 36:2252--2274.

\bibitem[{Deitke et~al.(2024)Deitke, Clark, Lee, Tripathi, Yang, Park, Salehi, Muennighoff, Lo, Soldaini, Lu, Anderson, Bransom, Ehsani, Ngo, Chen, Patel, Yatskar, Callison{-}Burch, Head, Hendrix, Bastani, VanderBilt, Lambert, Chou, Chheda, Sparks, Skjonsberg, Schmitz, Sarnat, Bischoff, Walsh, Newell, Wolters, Gupta, Zeng, Borchardt, Groeneveld, Dumas, Nam, Lebrecht, Wittlif, Schoenick, Michel, Krishna, Weihs, Smith, Hajishirzi, Girshick, Farhadi, and Kembhavi}]{deitke2024molmo}
Matt Deitke, Christopher Clark, Sangho Lee, Rohun Tripathi, Yue Yang, Jae~Sung Park, Mohammadreza Salehi, Niklas Muennighoff, Kyle Lo, Luca Soldaini, Jiasen Lu, Taira Anderson, Erin Bransom, Kiana Ehsani, Huong Ngo, Yen{-}Sung Chen, Ajay Patel, Mark Yatskar, Chris Callison{-}Burch, and 32 others. 2024.
\newblock \href {https://doi.org/10.48550/ARXIV.2409.17146} {Molmo and pixmo: Open weights and open data for state-of-the-art multimodal models}.
\newblock \emph{CoRR}, abs/2409.17146.

\bibitem[{Frantar and Alistarh(2023)}]{frantar2023sparsegpt}
Elias Frantar and Dan Alistarh. 2023.
\newblock Sparsegpt: Massive language models can be accurately pruned in one-shot.
\newblock In \emph{International Conference on Machine Learning}, pages 10323--10337. PMLR.

\bibitem[{Frantar et~al.(2022)Frantar, Ashkboos, Hoefler, and Alistarh}]{frantar2022gptq}
Elias Frantar, Saleh Ashkboos, Torsten Hoefler, and Dan Alistarh. 2022.
\newblock Gptq: Accurate post-training quantization for generative pre-trained transformers.
\newblock \emph{arXiv preprint arXiv:2210.17323}.

\bibitem[{Fu et~al.(2024)Fu, Chen, Shen, Qin, Zhang, Lin, Yang, Zheng, Li, Sun, Wu, and Ji}]{fu2024mmecomprehensiveevaluationbenchmark}
Chaoyou Fu, Peixian Chen, Yunhang Shen, Yulei Qin, Mengdan Zhang, Xu~Lin, Jinrui Yang, Xiawu Zheng, Ke~Li, Xing Sun, Yunsheng Wu, and Rongrong Ji. 2024.
\newblock \href {https://arxiv.org/abs/2306.13394} {Mme: A comprehensive evaluation benchmark for multimodal large language models}.
\newblock \emph{Preprint}, arXiv:2306.13394.

\bibitem[{Ge et~al.(2024)Ge, Zhang, Liu, Zhang, Han, and Gao}]{ge2024model}
Suyu Ge, Yunan Zhang, Liyuan Liu, Minjia Zhang, Jiawei Han, and Jianfeng Gao. 2024.
\newblock Model tells you what to discard: Adaptive {KV} cache compression for llms.
\newblock In \emph{The Twelfth International Conference on Learning Representations, {ICLR} 2024, Vienna, Austria, May 7-11, 2024}. OpenReview.net.

\bibitem[{Hu et~al.(2024)Hu, Shang, Wan, and Feng}]{hu2024iLLaVA}
Lianyu Hu, Fanhua Shang, Liang Wan, and Wei Feng. 2024.
\newblock illava: An image is worth fewer than 1/3 input tokens in large multimodal models.
\newblock \emph{CoRR}, abs/2412.06263.

\bibitem[{Huang et~al.(2024)Huang, Zhai, Shen, Cao, Zhao, Xu, Ye, and Lin}]{huang2024dynamicllava}
Wenxuan Huang, Zijie Zhai, Yunhang Shen, Shaoshen Cao, Fei Zhao, Xiangfeng Xu, Zheyu Ye, and Shaohui Lin. 2024.
\newblock Dynamic-llava: Efficient multimodal large language models via dynamic vision-language context sparsification.
\newblock \emph{arXiv preprint arXiv:2412.00876}.

\bibitem[{Hudson and Manning(2019)}]{hudson2019gqa}
Drew~A Hudson and Christopher~D Manning. 2019.
\newblock Gqa: A new dataset for real-world visual reasoning and compositional question answering.
\newblock In \emph{Proceedings of the IEEE/CVF conference on computer vision and pattern recognition}, pages 6700--6709.

\bibitem[{Kembhavi et~al.(2016)Kembhavi, Salvato, Kolve, Seo, Hajishirzi, and Farhadi}]{kembhavi2016diagramai2d}
Aniruddha Kembhavi, Mike Salvato, Eric Kolve, Minjoon Seo, Hannaneh Hajishirzi, and Ali Farhadi. 2016.
\newblock \href {https://arxiv.org/abs/1603.07396} {A diagram is worth a dozen images}.
\newblock \emph{Preprint}, arXiv:1603.07396.

\bibitem[{Li et~al.(2024{\natexlab{a}})Li, Zhang, Zhang, Guo, Zhang, Li, Zhang, Liu, and Li}]{li2024llavanext-strong}
Bo~Li, Kaichen Zhang, Hao Zhang, Dong Guo, Renrui Zhang, Feng Li, Yuanhan Zhang, Ziwei Liu, and Chunyuan Li. 2024{\natexlab{a}}.
\newblock \href {https://llava-vl.github.io/blog/2024-05-10-llava-next-stronger-llms/} {Llava-next: Stronger llms supercharge multimodal capabilities in the wild}.

\bibitem[{Li et~al.(2024{\natexlab{b}})Li, Zhang, Guo, Zhang, Li, Zhang, Zhang, Zhang, Li, Liu et~al.}]{li2024llava}
Bo~Li, Yuanhan Zhang, Dong Guo, Renrui Zhang, Feng Li, Hao Zhang, Kaichen Zhang, Peiyuan Zhang, Yanwei Li, Ziwei Liu, and 1 others. 2024{\natexlab{b}}.
\newblock Llava-onevision: Easy visual task transfer.
\newblock \emph{arXiv preprint arXiv:2408.03326}.

\bibitem[{Li et~al.(2024{\natexlab{c}})Li, Yuan, Liu, Tang, Wang, Qin, Zhu, and Zhang}]{li2024tokenpacker}
Wentong Li, Yuqian Yuan, Jian Liu, Dongqi Tang, Song Wang, Jie Qin, Jianke Zhu, and Lei Zhang. 2024{\natexlab{c}}.
\newblock Tokenpacker: Efficient visual projector for multimodal llm.
\newblock \emph{arXiv preprint arXiv:2407.02392}.

\bibitem[{Li et~al.(2024{\natexlab{d}})Li, Zhang, Wang, Zhong, Chen, Chu, Liu, and Jia}]{li2024mini}
Yanwei Li, Yuechen Zhang, Chengyao Wang, Zhisheng Zhong, Yixin Chen, Ruihang Chu, Shaoteng Liu, and Jiaya Jia. 2024{\natexlab{d}}.
\newblock Mini-gemini: Mining the potential of multi-modality vision language models.
\newblock \emph{arXiv preprint arXiv:2403.18814}.

\bibitem[{Li et~al.(2024{\natexlab{e}})Li, Huang, Yang, Venkitesh, Locatelli, Ye, Cai, Lewis, and Chen}]{li2024snapkv}
Yuhong Li, Yingbing Huang, Bowen Yang, Bharat Venkitesh, Acyr Locatelli, Hanchen Ye, Tianle Cai, Patrick Lewis, and Deming Chen. 2024{\natexlab{e}}.
\newblock Snapkv: Llm knows what you are looking for before generation.
\newblock \emph{arXiv preprint arXiv:2404.14469}.

\bibitem[{Li et~al.(2025)Li, Liu, Li, Zhang, Xu, Chen, Shi, Jiang, Wang, Wang, Huang, Zhao, Jiang, Hong, Wang, Tian, Huai, Luo, Luo, Zhang, Hu, and Zhang}]{li2025perception}
Yunxin Li, Zhenyu Liu, Zitao Li, Xuanyu Zhang, Zhenran Xu, Xinyu Chen, Haoyuan Shi, Shenyuan Jiang, Xintong Wang, Jifang Wang, Shouzheng Huang, Xinping Zhao, Borui Jiang, Lanqing Hong, Longyue Wang, Zhuotao Tian, Baoxing Huai, Wenhan Luo, Weihua Luo, and 3 others. 2025.
\newblock Perception, reason, think, and plan: A survey on large multimodal reasoning models.
\newblock \emph{arXiv preprint arXiv:2505.04921}.

\bibitem[{Lin et~al.(2024)Lin, Tang, Tang, Yang, Chen, Wang, Xiao, Dang, Gan, and Han}]{lin2024awq}
Ji~Lin, Jiaming Tang, Haotian Tang, Shang Yang, Wei-Ming Chen, Wei-Chen Wang, Guangxuan Xiao, Xingyu Dang, Chuang Gan, and Song Han. 2024.
\newblock Awq: Activation-aware weight quantization for on-device llm compression and acceleration.
\newblock \emph{Proceedings of Machine Learning and Systems}, 6:87--100.

\bibitem[{Liu et~al.(2024{\natexlab{a}})Liu, Li, Li, and Lee}]{liu2024improved}
Haotian Liu, Chunyuan Li, Yuheng Li, and Yong~Jae Lee. 2024{\natexlab{a}}.
\newblock Improved baselines with visual instruction tuning.
\newblock In \emph{{IEEE/CVF} Conference on Computer Vision and Pattern Recognition, {CVPR} 2024, Seattle, WA, USA, June 16-22, 2024}, pages 26286--26296.

\bibitem[{Liu et~al.(2023)Liu, Li, Wu, and Lee}]{liu2023visual}
Haotian Liu, Chunyuan Li, Qingyang Wu, and Yong~Jae Lee. 2023.
\newblock Visual instruction tuning.
\newblock In \emph{Advances in Neural Information Processing Systems 36: Annual Conference on Neural Information Processing Systems 2023, NeurIPS 2023, New Orleans, LA, USA, December 10 - 16, 2023}.

\bibitem[{Liu et~al.(2024{\natexlab{b}})Liu, Duan, Zhang, Li, Zhang, Zhao, Yuan, Wang, He, Liu, Chen, and Lin}]{liu2024mmbenchmultimodalmodelallaround}
Yuan Liu, Haodong Duan, Yuanhan Zhang, Bo~Li, Songyang Zhang, Wangbo Zhao, Yike Yuan, Jiaqi Wang, Conghui He, Ziwei Liu, Kai Chen, and Dahua Lin. 2024{\natexlab{b}}.
\newblock \href {https://arxiv.org/abs/2307.06281} {Mmbench: Is your multi-modal model an all-around player?}
\newblock \emph{Preprint}, arXiv:2307.06281.

\bibitem[{Liu et~al.(2024{\natexlab{c}})Liu, Li, Huang, Yang, Yu, Li, Yin, Liu, Jin, and Bai}]{Liu_2024ocrbench}
Yuliang Liu, Zhang Li, Mingxin Huang, Biao Yang, Wenwen Yu, Chunyuan Li, Xu-Cheng Yin, Cheng-Lin Liu, Lianwen Jin, and Xiang Bai. 2024{\natexlab{c}}.
\newblock \href {https://doi.org/10.1007/s11432-024-4235-6} {Ocrbench: on the hidden mystery of ocr in large multimodal models}.
\newblock \emph{Science China Information Sciences}, 67(12).

\bibitem[{Liu et~al.(2024{\natexlab{d}})Liu, Zhu, Shi, Zhang, Lou, Yang, Xi, Cao, Gu, Li, Li, Fang, Chen, Hsieh, Huang, Cheng, Nath, Hu, Liu, Krishna, Xu, Wang, Molchanov, Kautz, Yin, Han, and Lu}]{liu2024nvila}
Zhijian Liu, Ligeng Zhu, Baifeng Shi, Zhuoyang Zhang, Yuming Lou, Shang Yang, Haocheng Xi, Shiyi Cao, Yuxian Gu, Dacheng Li, Xiuyu Li, Yunhao Fang, Yukang Chen, Cheng{-}Yu Hsieh, De{-}An Huang, An{-}Chieh Cheng, Vishwesh Nath, Jinyi Hu, Sifei Liu, and 8 others. 2024{\natexlab{d}}.
\newblock \href {https://doi.org/10.48550/ARXIV.2412.04468} {{NVILA:} efficient frontier visual language models}.
\newblock \emph{CoRR}, abs/2412.04468.

\bibitem[{Lu et~al.(2024{\natexlab{a}})Lu, Liu, Zhang, Wang, Dong, Liu, Sun, Ren, Li, Yang, Sun, Deng, Xu, Xie, and Ruan}]{lu2024deepseekvl}
Haoyu Lu, Wen Liu, Bo~Zhang, Bingxuan Wang, Kai Dong, Bo~Liu, Jingxiang Sun, Tongzheng Ren, Zhuoshu Li, Hao Yang, Yaofeng Sun, Chengqi Deng, Hanwei Xu, Zhenda Xie, and Chong Ruan. 2024{\natexlab{a}}.
\newblock \href {https://doi.org/10.48550/ARXIV.2403.05525} {Deepseek-vl: Towards real-world vision-language understanding}.
\newblock \emph{CoRR}, abs/2403.05525.

\bibitem[{Lu et~al.(2024{\natexlab{b}})Lu, Bansal, Xia, Liu, Li, Hajishirzi, Cheng, Chang, Galley, and Gao}]{lu2024mathvista}
Pan Lu, Hritik Bansal, Tony Xia, Jiacheng Liu, Chunyuan Li, Hannaneh Hajishirzi, Hao Cheng, Kai-Wei Chang, Michel Galley, and Jianfeng Gao. 2024{\natexlab{b}}.
\newblock Mathvista: Evaluating mathematical reasoning of foundation models in visual contexts.
\newblock In \emph{International Conference on Learning Representations (ICLR)}.

\bibitem[{Luo et~al.(2024)Luo, Luo, Ji, Zhou, Sun, Shen, and Ji}]{luo2024gammamod}
Yaxin Luo, Gen Luo, Jiayi Ji, Yiyi Zhou, Xiaoshuai Sun, Zhiqiang Shen, and Rongrong Ji. 2024.
\newblock \href {https://doi.org/10.48550/arXiv.2410.13859} {{\(\gamma\)}-mod: Exploring mixture-of-depth adaptation for multimodal large language models}.
\newblock \emph{CoRR}, abs/2410.13859.

\bibitem[{Masry et~al.(2022)Masry, Long, Tan, Joty, and Hoque}]{masry2022chartqa}
Ahmed Masry, Do~Xuan Long, Jia~Qing Tan, Shafiq Joty, and Enamul Hoque. 2022.
\newblock Chartqa: A benchmark for question answering about charts with visual and logical reasoning.
\newblock \emph{arXiv preprint arXiv:2203.10244}.

\bibitem[{Mathew et~al.(2020)Mathew, Karatzas, Manmatha, and Jawahar}]{mathew2020docvqa}
Minesh Mathew, Dimosthenis Karatzas, R~Manmatha, and CV~Jawahar. 2020.
\newblock Docvqa: A dataset for vqa on document images. corr abs/2007.00398 (2020).
\newblock \emph{arXiv preprint arXiv:2007.00398}.

\bibitem[{OpenAI(2024)}]{gpt4o}
OpenAI. 2024.
\newblock \href {https://openai.com/index/hello-gpt-4o} {Hello gpt-4o}.

\bibitem[{OpenAI(2025)}]{openai-operator}
OpenAI. 2025.
\newblock \href {https://openai.com/index/introducing-operator} {Introducing operator}.

\bibitem[{Qiu et~al.(2025)Qiu, Wang, Zheng, Huang, Wen, Yang, Men, Yu, Huang, Huang et~al.}]{qiu2025gated}
Zihan Qiu, Zekun Wang, Bo~Zheng, Zeyu Huang, Kaiyue Wen, Songlin Yang, Rui Men, Le~Yu, Fei Huang, Suozhi Huang, and 1 others. 2025.
\newblock Gated attention for large language models: Non-linearity, sparsity, and attention-sink-free.
\newblock \emph{arXiv preprint arXiv:2505.06708}.

\bibitem[{Shang et~al.(2024)Shang, Cai, Xu, Lee, and Yan}]{shang2024llavaprumerge}
Yuzhang Shang, Mu~Cai, Bingxin Xu, Yong~Jae Lee, and Yan Yan. 2024.
\newblock Llava-prumerge: Adaptive token reduction for efficient large multimodal models.
\newblock \emph{arXiv preprint arXiv:2403.15388}.

\bibitem[{Shu et~al.(2024)Shu, Liao, Zhuo, Xu, Zhang, Shi, Chen, Zhong, He, Fu, Li, Li, Yu, Liu, Li, and Jiang}]{shu2024llavamod}
Fangxun Shu, Yue Liao, Le~Zhuo, Chenning Xu, Guanghao Zhang, Haonan Shi, Long Chen, Tao Zhong, Wanggui He, Siming Fu, Haoyuan Li, Bolin Li, Zhelun Yu, Si~Liu, Hongsheng Li, and Hao Jiang. 2024.
\newblock \href {https://doi.org/10.48550/ARXIV.2408.15881} {Llava-mod: Making llava tiny via moe knowledge distillation}.
\newblock \emph{CoRR}, abs/2408.15881.

\bibitem[{Singh et~al.(2019)Singh, Natarjan, Shah, Jiang, Chen, Batra, Parikh, and Rohrbach}]{singh2019textvqa}
Amanpreet Singh, Vivek Natarjan, Meet Shah, Yu~Jiang, Xinlei Chen, Dhruv Batra, Devi Parikh, and Marcus Rohrbach. 2019.
\newblock Towards vqa models that can read.
\newblock In \emph{Proceedings of the IEEE Conference on Computer Vision and Pattern Recognition}, pages 8317--8326.

\bibitem[{Song et~al.(2023)Song, Chai, Wang, Zhang, Zhou, Wu, Guo, Ye, Lu, Hwang et~al.}]{song2023moviechat}
Enxin Song, Wenhao Chai, Guanhong Wang, Yucheng Zhang, Haoyang Zhou, Feiyang Wu, Xun Guo, Tian Ye, Yan Lu, Jenq-Neng Hwang, and 1 others. 2023.
\newblock Moviechat: From dense token to sparse memory for long video understanding.
\newblock \emph{arXiv preprint arXiv:2307.16449}.

\bibitem[{Sun et~al.(2023)Sun, Liu, Bair, and Kolter}]{sun2023simple}
Mingjie Sun, Zhuang Liu, Anna Bair, and J~Zico Kolter. 2023.
\newblock A simple and effective pruning approach for large language models.
\newblock \emph{arXiv preprint arXiv:2306.11695}.

\bibitem[{Sung et~al.(2023)Sung, Yoon, and Bansal}]{sung2023ecoflap}
Yi-Lin Sung, Jaehong Yoon, and Mohit Bansal. 2023.
\newblock Ecoflap: Efficient coarse-to-fine layer-wise pruning for vision-language models.
\newblock \emph{arXiv preprint arXiv:2310.02998}.

\bibitem[{Team et~al.(2024)Team, Georgiev, Lei, Burnell, Bai, Gulati, Tanzer, Vincent, Pan, Wang et~al.}]{team2024gemini}
Gemini Team, Petko Georgiev, Ving~Ian Lei, Ryan Burnell, Libin Bai, Anmol Gulati, Garrett Tanzer, Damien Vincent, Zhufeng Pan, Shibo Wang, and 1 others. 2024.
\newblock \href {https://arxiv.org/abs/2403.05530} {Gemini 1.5: Unlocking multimodal understanding across millions of tokens of context}.
\newblock \emph{ArXiv preprint}, abs/2403.05530.

\bibitem[{Team(2025)}]{Qwen2.5-VL}
Qwen Team. 2025.
\newblock \href {https://qwenlm.github.io/blog/qwen2.5-vl/} {Qwen2.5-vl}.

\bibitem[{Tito et~al.(2022)Tito, Karatzas, and Valveny}]{tito2022hierarchicalmpdocvqa}
Rub{\`e}n Tito, Dimosthenis Karatzas, and Ernest Valveny. 2022.
\newblock Hierarchical multimodal transformers for multi-page docvqa.
\newblock \emph{arXiv preprint arXiv:2212.05935}.

\bibitem[{Tu et~al.(2024)Tu, Vashchilenko, Lu, and Xu}]{tu2024vlcache}
Dezhan Tu, Danylo Vashchilenko, Yuzhe Lu, and Panpan Xu. 2024.
\newblock Vl-cache: Sparsity and modality-aware kv cache compression for vision-language model inference acceleration.
\newblock \emph{arXiv preprint arXiv:2410.23317}.

\bibitem[{Wan et~al.(2024)Wan, Wu, Liu, Huang, Zhu, Jin, Wang, and Yuan}]{wan2024look}
Zhongwei Wan, Ziang Wu, Che Liu, Jinfa Huang, Zhihong Zhu, Peng Jin, Longyue Wang, and Li~Yuan. 2024.
\newblock Look-m: Look-once optimization in kv cache for efficient multimodal long-context inference.
\newblock In \emph{Findings of the Association for Computational Linguistics: EMNLP 2024}, pages 4065--4078.

\bibitem[{Wang et~al.(2024{\natexlab{a}})Wang, Bai, Tan, Wang, Fan, Bai, Chen, Liu, Wang, Ge et~al.}]{wang2024qwen2}
Peng Wang, Shuai Bai, Sinan Tan, Shijie Wang, Zhihao Fan, Jinze Bai, Keqin Chen, Xuejing Liu, Jialin Wang, Wenbin Ge, and 1 others. 2024{\natexlab{a}}.
\newblock Qwen2-vl: Enhancing vision-language model's perception of the world at any resolution.
\newblock \emph{arXiv preprint arXiv:2409.12191}.

\bibitem[{Wang et~al.(2023{\natexlab{a}})Wang, Zhou, Zeng, and Zhang}]{wang2023efficientvlm}
Tiannan Wang, Wangchunshu Zhou, Yan Zeng, and Xinsong Zhang. 2023{\natexlab{a}}.
\newblock Efficientvlm: Fast and accurate vision-language models via knowledge distillation and modal-adaptive pruning.
\newblock In \emph{Findings of the Association for Computational Linguistics: {ACL} 2023, Toronto, Canada, July 9-14, 2023}, pages 13899--13913. Association for Computational Linguistics.

\bibitem[{Wang et~al.(2024{\natexlab{b}})Wang, Ma, Wang, Chen, Fan, Shan, Yang, Xu, Liu, and Qin}]{wang2024cfsp}
Yuxin Wang, Minghua Ma, Zekun Wang, Jingchang Chen, Huiming Fan, Liping Shan, Qing Yang, Dongliang Xu, Ming Liu, and Bing Qin. 2024{\natexlab{b}}.
\newblock \href {https://doi.org/10.48550/ARXIV.2409.13199} {{CFSP:} an efficient structured pruning framework for llms with coarse-to-fine activation information}.
\newblock \emph{CoRR}, abs/2409.13199.

\bibitem[{Wang et~al.(2023{\natexlab{b}})Wang, Chen, Zhou, Liu, and Qin}]{wang2024smarttrim}
Zekun Wang, Jingchang Chen, Wangchunshu Zhou, Ming Liu, and Bing Qin. 2023{\natexlab{b}}.
\newblock \href {https://doi.org/10.48550/ARXIV.2305.15033} {Smarttrim: Adaptive tokens and parameters pruning for efficient vision-language models}.
\newblock \emph{CoRR}, abs/2305.15033.

\bibitem[{Wang et~al.(2021)Wang, Wang, Zhu, Liu, Qin, and Wei}]{wang2021distilled}
Zekun Wang, Wenhui Wang, Haichao Zhu, Ming Liu, Bing Qin, and Furu Wei. 2021.
\newblock \href {https://arxiv.org/abs/2112.08723} {Distilled dual-encoder model for vision-language understanding}.
\newblock \emph{CoRR}, abs/2112.08723.

\bibitem[{x.ai()}]{grok15v}
x.ai.
\newblock \href {https://x.ai/blog/grok-1.5v} {Grok-1.5 vision preview}.

\bibitem[{Xiao et~al.(2023)Xiao, Tian, Chen, Han, and Lewis}]{xiao2023streamingllm}
Guangxuan Xiao, Yuandong Tian, Beidi Chen, Song Han, and Mike Lewis. 2023.
\newblock Efficient streaming language models with attention sinks.
\newblock \emph{arXiv}.

\bibitem[{Xu et~al.(2021)Xu, Zhou, Ge, Xu, McAuley, and Wei}]{xu2021beyond}
Canwen Xu, Wangchunshu Zhou, Tao Ge, Ke~Xu, Julian~J. McAuley, and Furu Wei. 2021.
\newblock Beyond preserved accuracy: Evaluating loyalty and robustness of {BERT} compression.
\newblock In \emph{Proceedings of the 2021 Conference on Empirical Methods in Natural Language Processing, {EMNLP} 2021, Virtual Event / Punta Cana, Dominican Republic, 7-11 November, 2021}, pages 10653--10659. Association for Computational Linguistics.

\bibitem[{Xu et~al.(2024{\natexlab{a}})Xu, Wang, Wang, Lu, Xie, Saha, Sahoo, Yu, and Xiong}]{xu2024aguvis}
Yiheng Xu, Zekun Wang, Junli Wang, Dunjie Lu, Tianbao Xie, Amrita Saha, Doyen Sahoo, Tao Yu, and Caiming Xiong. 2024{\natexlab{a}}.
\newblock \href {https://doi.org/10.48550/ARXIV.2412.04454} {Aguvis: Unified pure vision agents for autonomous {GUI} interaction}.
\newblock \emph{CoRR}, abs/2412.04454.

\bibitem[{Xu et~al.(2024{\natexlab{b}})Xu, Jie, Dong, Wang, Lu, Zhou, Saha, Xiong, and Sahoo}]{xu2024thinkk}
Yuhui Xu, Zhanming Jie, Hanze Dong, Lei Wang, Xudong Lu, Aojun Zhou, Amrita Saha, Caiming Xiong, and Doyen Sahoo. 2024{\natexlab{b}}.
\newblock \href {https://doi.org/10.48550/ARXIV.2407.21018} {Think: Thinner key cache by query-driven pruning}.
\newblock \emph{CoRR}, abs/2407.21018.

\bibitem[{Yang et~al.(2024)Yang, Chen, Tian, Wang, Li, Yu, and Jia}]{yang2024visionzip}
Senqiao Yang, Yukang Chen, Zhuotao Tian, Chengyao Wang, Jingyao Li, Bei Yu, and Jiaya Jia. 2024.
\newblock Visionzip: Longer is better but not necessary in vision language models.
\newblock \emph{arXiv preprint arXiv:2412.04467}.

\bibitem[{Yao et~al.(2024)Yao, Yu, Zhang, Wang, Cui, Zhu, Cai, Li, Zhao, He et~al.}]{yao2024minicpmv}
Yuan Yao, Tianyu Yu, Ao~Zhang, Chongyi Wang, Junbo Cui, Hongji Zhu, Tianchi Cai, Haoyu Li, Weilin Zhao, Zhihui He, and 1 others. 2024.
\newblock Minicpm-v: A gpt-4v level mllm on your phone.
\newblock \emph{arXiv preprint arXiv:2408.01800}.

\bibitem[{Yu et~al.(2025)Yu, Zhou, Yang, Wang, Li, Hu, Xu, Xu, Shu, and Yuan}]{yu2025mquant}
JiangYong Yu, Sifan Zhou, Dawei Yang, Shuo Wang, Shuoyu Li, Xing Hu, Chen Xu, Zukang Xu, Changyong Shu, and Zhihang Yuan. 2025.
\newblock Mquant: Unleashing the inference potential of multimodal large language models via full static quantization.
\newblock \emph{arXiv preprint arXiv:2502.00425}.

\bibitem[{Yu et~al.(2023)Yu, Yang, Li, Wang, Lin, Liu, Wang, and Wang}]{yu2023mmvet}
Weihao Yu, Zhengyuan Yang, Linjie Li, Jianfeng Wang, Kevin Lin, Zicheng Liu, Xinchao Wang, and Lijuan Wang. 2023.
\newblock \href {https://arxiv.org/abs/2308.02490} {Mm-vet: Evaluating large multimodal models for integrated capabilities}.
\newblock \emph{Preprint}, arXiv:2308.02490.

\bibitem[{Yue et~al.(2024)Yue, Ni, Zhang, Zheng, Liu, Zhang, Stevens, Jiang, Ren, Sun, Wei, Yu, Yuan, Sun, Yin, Zheng, Yang, Liu, Huang, Sun, Su, and Chen}]{yue2023mmmu}
Xiang Yue, Yuansheng Ni, Kai Zhang, Tianyu Zheng, Ruoqi Liu, Ge~Zhang, Samuel Stevens, Dongfu Jiang, Weiming Ren, Yuxuan Sun, Cong Wei, Botao Yu, Ruibin Yuan, Renliang Sun, Ming Yin, Boyuan Zheng, Zhenzhu Yang, Yibo Liu, Wenhao Huang, and 3 others. 2024.
\newblock Mmmu: A massive multi-discipline multimodal understanding and reasoning benchmark for expert agi.
\newblock In \emph{Proceedings of CVPR}.

\bibitem[{Zhang et~al.(2024{\natexlab{a}})Zhang, Li, Zhang, Pu, Cahyono, Hu, Liu, Zhang, Yang, Li, and Liu}]{zhang2024lmmsevalrealitycheckevaluation}
Kaichen Zhang, Bo~Li, Peiyuan Zhang, Fanyi Pu, Joshua~Adrian Cahyono, Kairui Hu, Shuai Liu, Yuanhan Zhang, Jingkang Yang, Chunyuan Li, and Ziwei Liu. 2024{\natexlab{a}}.
\newblock \href {https://arxiv.org/abs/2407.12772} {Lmms-eval: Reality check on the evaluation of large multimodal models}.
\newblock \emph{Preprint}, arXiv:2407.12772.

\bibitem[{Zhang et~al.(2024{\natexlab{b}})Zhang, Fan, Ma, Zheng, Huang, Cheng, Gudovskiy, Okuno, Nakata, Keutzer et~al.}]{zhang2024sparsevlm}
Yuan Zhang, Chun-Kai Fan, Junpeng Ma, Wenzhao Zheng, Tao Huang, Kuan Cheng, Denis Gudovskiy, Tomoyuki Okuno, Yohei Nakata, Kurt Keutzer, and 1 others. 2024{\natexlab{b}}.
\newblock Sparsevlm: Visual token sparsification for efficient vision-language model inference.
\newblock \emph{arXiv preprint arXiv:2410.04417}.

\bibitem[{Zhang et~al.(2023{\natexlab{a}})Zhang, Sheng, Zhou, Chen, Zheng, Cai, Song, Tian, R\'{e}, Barrett, Wang, and Chen}]{NEURIPS2023_6ceefa7b}
Zhenyu Zhang, Ying Sheng, Tianyi Zhou, Tianlong Chen, Lianmin Zheng, Ruisi Cai, Zhao Song, Yuandong Tian, Christopher R\'{e}, Clark Barrett, Zhangyang~"Atlas" Wang, and Beidi Chen. 2023{\natexlab{a}}.
\newblock \href {https://proceedings.neurips.cc/paper_files/paper/2023/file/6ceefa7b15572587b78ecfcebb2827f8-Paper-Conference.pdf} {H2o: Heavy-hitter oracle for efficient generative inference of large language models}.
\newblock In \emph{Advances in Neural Information Processing Systems}, volume~36, pages 34661--34710. Curran Associates, Inc.

\bibitem[{Zhang et~al.(2023{\natexlab{b}})Zhang, Sheng, Zhou, Chen, Zheng, Cai, Song, Tian, R{\'e}, Barrett et~al.}]{zhang2023h2o}
Zhenyu Zhang, Ying Sheng, Tianyi Zhou, Tianlong Chen, Lianmin Zheng, Ruisi Cai, Zhao Song, Yuandong Tian, Christopher R{\'e}, Clark Barrett, and 1 others. 2023{\natexlab{b}}.
\newblock H2o: Heavy-hitter oracle for efficient generative inference of large language models.
\newblock \emph{Advances in Neural Information Processing Systems}, 36:34661--34710.

\bibitem[{Zhou et~al.(2024)Zhou, Hu, Weng, Jia, Luo, Liu, Wu, and Huang}]{zhou2024tinyllava}
Baichuan Zhou, Ying Hu, Xi~Weng, Junlong Jia, Jie Luo, Xien Liu, Ji~Wu, and Lei Huang. 2024.
\newblock Tinyllava: A framework of small-scale large multimodal models.
\newblock \emph{arXiv preprint arXiv:2402.14289}.

\end{thebibliography}
\appendix
\section{Appendix}
\label{sec:appendix}
\subsection{Evaluation Benchmarks}
\label{sec:appendix_evaluation_benchmarks}
We conduct evaluations in realistic scenarios, using diverse benchmarks that span a wide range of vision-language tasks. These are grouped into three categories:  
\paragraph{\emph{(1)~Chart, Diagram, and Document Understanding}}
This category focuses on textual and diagrammatic comprehension. Benchmarks include DocVQA~\cite{mathew2020docvqa}, ChartQA~\cite{masry2022chartqa}, TextVQA~\cite{singh2019textvqa}, OCRBench~\cite{Liu_2024ocrbench}, and AI2D~\cite{kembhavi2016diagramai2d}. These test the ability to accurately detect and recognize text in various forms (printed, handwritten, or embedded in charts/diagrams) and to understand document layouts or diagrammatic structures.

\paragraph{\emph{(2)~Perception and Multi-discipline Reasoning}}
This category targets visual perception combined with logical or knowledge-intensive reasoning. Benchmarks include GQA~\cite{hudson2019gqa}, MMMU~\cite{yue2023mmmu}, MME~\cite{fu2024mmecomprehensiveevaluationbenchmark}, MMBench~\cite{liu2024mmbenchmultimodalmodelallaround}, MMVet~\cite{yu2023mmvet}, MMStar~\cite{chen2024wemmstar}, and MathVista~\cite{lu2024mathvista}.

\paragraph{\emph{(3)~Real-world Compositional \& Interactive QA}}
This category assesses interactive and compositional reasoning in realistic conversational settings. Benchmarks like LLaVA-Wilder~\cite{li2024llavanext-strong}, RealWorldQA~\cite{grok15v}, and ImageDC~\cite{li2024llavanext-strong} test a model’s ability to handle open-ended questions, multi-turn dialogues, and complex visual-linguistic grounding, as well as its adaptation to diverse topics and coherence over extended interactions.

Across these benchmarks, we evaluate various compression methods on multimodal understanding and reasoning. Input/output lengths differ significantly by task (Table~\ref{tab:token_length}). To assess KV cache methods on longer sequences, we select MP-DocVQA~\cite{tito2022hierarchicalmpdocvqa} for multi-images and MovieChat~\cite{song2023moviechat} for video.

\begin{table*}[htbp]
\centering
\resizebox{\linewidth}{!}{
\small
\begin{tabular}{l|l|ccccccccccccccc}
\toprule
 & \raisebox{5ex}{\diagbox{\textbf{Models}}{\textbf{Benchmarks}}} & \rotatebox{90}{DocVQA} & \rotatebox{90}{ChartQA} & \rotatebox{90}{TextVQA} & \rotatebox{90}{OCRBench} & \rotatebox{90}{AI2D} & \rotatebox{90}{GQA} & \rotatebox{90}{MMMU} & \rotatebox{90}{MME} & \rotatebox{90}{RealworldQA} & \rotatebox{90}{MMStar} & \rotatebox{90}{MathVista} & \rotatebox{90}{LLaVA-Wilder} & \rotatebox{90}{MMBench} & \rotatebox{90}{MMVet} & \rotatebox{90}{ImageDC} \\
\midrule
\multirow{3}{*}{Input} 
 & LLaVA-OneVision-7B 
   & 7313  & 3882  & 5147  & 1233  & 3497  & 2693  & 3613  & 5768  & 7778  & 2226  & 2935  & 4559  & 1903  & 5006  & 2704 \\
 & Qwen2-VL-7B 
   & 2027 & 406 & 1010 & 75 & 598 & 383 & 730 & 1141 & 1776 & 288 & 465 & 904 & 251 & 1018 & 363 \\
 & InternVL2.5-38B 
   &  1835 & 1367 & 1744 & 1212 & 1389 & 1800 & 1588 & 1506 & 1834 & 1478 & 1346 & 1337 & 1387 & 1400 & 1492 \\
\midrule
\multirow{3}{*}{Output} 
 & LLaVA-OneVision-7B 
   &   2.72    &   2.92   &    2.69  &   26.04    &  1.00     &    1.27   &     1.05  &   1.03    &   1.02    &   1.00    &   2.85    &  252.12     &    1.20   &   59.07    &   414.61   \\
 & Qwen2-VL-7B 
   &    1.82   &   2.96    &   2.96    &  3.71     &  1.00     &  1.23     &    1.42   &  1.01     &    1.01   &     1.00  &   2.65    &   55.54    &      1.72 &   24.70    &  314.22    \\
 & InternVL2.5-38B 
   &   1.72    &   3.27    &   2.86    &  11.27     &  1.00     &   1.23    &    1.07   &  1.00     &  1.01     &   1.00    &    78.80   &  257.73     &    1.00   &    145.62   &   158.66   \\
\bottomrule
\end{tabular}
}
\caption{Input and Output token length of different models on various benchmarks. We select 100 samples from each benchmark for calculation. }
\label{tab:token_length}
\end{table*}

\subsection{Compression Method}
\label{sec:appendix_method}
\paragraph{Token Pruning}
Token pruning compression reduces computational overhead by eliminating redundant tokens during inference. We implement methods including FastV~\cite{chen2024fastv}, VisionZip~\cite{yang2024visionzip}, and PruMerge+~\cite{shang2024llavaprumerge}, which selectively remove redundant tokens based on relevance scores or token-level importance metrics.

\paragraph{KV Cache Compression}
We explore several KV cache compression methods that reduce memory usage during inference by selectively retaining key-value states, including StreamingLLM~\cite{xiao2023streamingllm}, H2O~\cite{NEURIPS2023_6ceefa7b}, SnapKV~\cite{li2024snapkv}, PyramidKV~\cite{cai2024pyramidkv}, LOOK-M~\cite{wan2024look}, and VL-Cache~\cite{tu2024vlcache}. 

\paragraph{Model Pruning}
Model pruning aims to reduce model size and inference cost while preserving performance. We evaluate EcoFLAP~\cite{sung2023ecoflap}, Wanda~\cite{sun2023simple}, and SparseGPT~\cite{frantar2023sparsegpt}, applying these methods only to the LLM backbone of the LVLM. 

\paragraph{Model Quantization}
Model quantization compresses models by reducing the precision of their weights and activations, thereby lowering memory requirements and accelerating computation. We consider techniques like AWQ~\cite{lin2024awq} and GPTQ~\cite{frantar2022gptq} (applied only to the LVLM's LLM backbone).

\subsection{Details of Methods}
\label{sec:appendix_math_modeling}

\paragraph{Token Pruning}

Let $T_v = \{t_{v,j}\}_{j=1}^{l_v}$ be the set of $l_v$ visual tokens, each $t_{v,j} \in \mathbb{R}^d$. Given a retention budget $b \in (0,1]$, pruning aims to select $l'_v = \lfloor b \cdot l_v \rfloor$ tokens. This typically involves a scoring function $s: \mathbb{R}^d \to \mathbb{R}$ to assess token importance.

\medskip
\noindent
\textit{FastV} uses a lightweight scoring strategy. For each visual token $t_{v,j}$, its importance $s(t_{v,j})$ is its accumulated attention score among textual tokens.
A binary mask $m \in \{0,1\}^{l_v}$ is defined by
\[
m_j =
\begin{cases}
1, & \text{if } s(t_{v,j}) \ge \tau,\\[1mm]
0, & \text{otherwise,}
\end{cases}
\]
where the threshold $\tau$ is set to ensure the binary mask $m$ satisfies the following constraint:
\[
\sum_{j=1}^{l_v} m_j = \lfloor b \cdot l_v \rfloor.
\]
The pruned visual token set is given by 
\[
T'_v = \{\,t_{v,j} \mid m_j = 0\,\}.
\]

\medskip
\noindent
\textit{VisionZip} first selects an initial set of tokens to keep using accumulated attention (similar to FastV), and identifies the discarded set $T'_v$. To compensate for information loss from $T'_v$, VisionZip clusters $T_d$ into $k$ groups. The original paper sets $k = \lfloor (b/6.4) \cdot l_v \rfloor$. For each cluster $C_i \subseteq T'_v$, its centroid $c_i$ is computed:
\[
c_i = \frac{1}{|C_i|} \sum_{t \in C_i} t.
\]
The recycled token set is then defined as
\[
R_v = \{\,c_i \mid i=1,2,\dots,k\,\}.
\]
Finally, the tokens in $R_v$ are concatenated with the retained tokens.

\medskip
\noindent
\textit{PruMerge+} also adaptively selects important tokens and merges less important ones to retain critical information. For each visual token \( t_{v,j} \), its importance score is computed via the [CLS] token's attention weights in the vision encoder:
\[
s(t_{v,j}) = A^{[\text{CLS}]}_j,
\]
where \( A^{[\text{CLS}]} \in \mathbb{R}^{l_v} \) denotes the attention weight from the [CLS] token to visual tokens. For architectures without [CLS] tokens, we compute row-wise averages of the attention matrix:
\[
s(t_{v,j}) = \frac{1}{l_v}\sum_{i=1}^{l_v} A_{ij}.
\]

The binary mask \( m \in \{0,1\}^{l_v} \) is determined through interquartile range (IQR) based outlier detection:
\[
m_j = 
\begin{cases} 
1, & \text{if } s(t_{v,j}) \geq \tau, \\ 
0, & \text{otherwise},
\end{cases}
\]
where threshold \( \tau \) is adaptively calculated using the IQR method. To ensure budget compliance, we enforce:
\[
\sum_{j=1}^{l_v} m_j = \lfloor b \cdot l_v \rfloor.
\]
If IQR selection yields fewer than $l'_v = \lfloor b \cdot l_v \rfloor$ tokens (the target budget $b$), PruMerge+ first supplements this set by uniformly sampling from the remaining highest-scoring unselected candidates. All $l'_v$ tokens in the final retained set then undergo feature merging to consolidate information.
Unlike VisionZip, PruMerge+ merges information from pruned tokens into retained ones.

\paragraph{KV Cache Compression}
In the LVLM prefill stage, let $l_v$ and $l_t$ be the number of visual and text tokens, respectively, with total length $l = l_v + l_t$. Let $w$ be the recent window size and $b$ the compression budget fraction. The attention map is $A \in \mathbb{R}^{l \times l}$, where $A_{ij}$ is the attention weight between query $i$ and key $j$.
To compress the KV caches, we introduce:
\[
\mathcal{F}: \mathbb{R}^{l \times l} \,\to\, \{0,1\}^{l},
\]
which produces a binary mask $\mathbf{m}\in\{0,1\}^{l}$ indicating which tokens to retain ($\mathbf{m}_j=1$) or evict ($\mathbf{m}_j=0$). We consider four variants of $\mathcal{F}$ that rank token importance:

\begin{itemize}
    \item \textit{Accumulated Attention $\mathcal{F}_{\mathrm{acc}}$.}
    We sum attention weights along the query tokens dimension:
    \[
    s_j = \sum_{i=1}^{l} A_{ij}.
    \]

    \item \textit{Normalized Attention $\mathcal{F}_{\mathrm{norm}}$.}
    We normalize attention weights for each query $i$, then sum and average:
    \[
    \tilde{s}_j 
    = \textit{Norm}(\sum_{i=1}^{l}A_{ij})
    \]

    \item \textit{Sliding Window Attention $\mathcal{F}_{\mathrm{sw}}$.}
     We compute accumulated attention scores along the query tokens but only over a recent window:
     \[
    \tilde{s}_j 
    = \sum_{i=l-w+1}^{l}A_{ij}.
    \]

    \item \textit{Post-Vision Attention $\mathcal{F}_{\mathrm{pv}}$.}
    Recognizing that in many LVLMs, the textual tokens following the visual tokens are more critical, this variant computes attention scores using only the queries from the text region. Formally, letting the text region be indexed by \(i = l_v+1,\ldots,l\), we define:
    \[
    \tilde{s}_j = \sum_{i=l_v+1}^{l} A_{ij}.
    \]
\end{itemize}

These variants of $\mathcal{F}$ offer different strategies for ranking tokens, allowing us to evaluate how attention-based selection influences KV cache compression in LVLMs. In experiments, H2O and LOOK-M use \textit{Accumulated Attention}, SnapKV and PyramidKV use \textit{Sliding Window Attention} and VL-Cache uses \textit{Post-Vision Attention}.

\paragraph{Model Pruning} aims to identify a binary mask $S \in \{0,1\}^{m \times n}$ so that the pruned weight matrix $\widetilde{W} = W \odot S$ preserves performance while satisfying a desired sparsity level, i.e., $\|S\|_0 = p \cdot (m \times n),$ with $p$ being the fraction of weights to retain. An importance score $s_{ij}$ is computed for each weight $W_{ij}$. The binary mask is then determined by selecting the top-$p$ fraction of weights with the highest scores.

\medskip
\noindent
\textit{Wanda} measures the importance of each weight by the product of its magnitude and the norm of its corresponding input activation:
\[
s_{ij} = |W_{ij}| \cdot \|X_j\|_2,
\]
where $X_j$ is the input activation vector associated with the $j$-th column of $W$. This prioritizes weights coupled with strong activations, preserving contributions critical to model performance even if the weights themselves are small.

\medskip
\noindent
\textit{EcoFLAP} prunes LVLMs layer-wise in a coarse-to-fine manner. The coarse phase computes layer $i$'s importance $S(W_i)$ using an expected zeroth-order gradient approximation:
$||\triangledown_{{W}_{i}}\mathcal{L}(W_i, \mathcal{D})||_2 = \mathbb{E}_{d \sim \mathcal{D}}[ \mathbb{E}_{z \sim N(0, 1)} [ | \frac{\mathcal{L}(W_i + \epsilon z, d) - \mathcal{L}(W_i - \epsilon z, d)}{2\epsilon}| ]]$.
The fine-grained step then uses same score as Wanda locally within each layer to prune weights according to these layer-specific sparsity ratios.

\medskip
\noindent
\textit{SparseGPT} minimizes a layer’s output reconstruction error using a second-order approximation. Its importance metric is derived from an approximate diagonal of the Hessian:
\[
s_{ij}^{\mathrm{SparseGPT}} = [\frac{||W||^2}{\text{diag}(XX^T+\lambda I)^{-1}}]_{ij}
\]

\paragraph{Model Quantization}
Quantization aims to approximate a full-precision weight matrix $W \in \mathbb{R}^{m \times n}$ with a low-bit representation while minimizing performance degradation. In our experiments, we consider two notable approaches of \emph{Post-Training Quantization}: AWQ and GPTQ.

\medskip
\noindent
\textit{AWQ} refines the PTQ process by incorporating input activation statistics to better preserve the layer output. Given the weight matrix $W$ and its corresponding input activations $X$, AWQ selects quantization parameters by solving:
\[
\min_{s,\, z} \left\|WX - \widehat{W}X\right\|_F^2
\]
with $\widehat{W}_{ij} = s\cdot\left(\operatorname{round}\!\left(\frac{W_{ij}}{s}\right) - z\right)$.

\medskip
\noindent
\textit{GPTQ} enhances PTQ by leveraging second-order information to minimize the output error induced by quantization. Starting from the uniform quantization of $W$, GPTQ aims to adjust the quantization parameters so as to minimize the reconstruction error using a Taylor expansion involving an approximate Hessian $H$ (often estimated via the diagonal of $X^\top X$). This Hessian-guided correction allows GPTQ to achieve high quantization fidelity in one shot, even under aggressive bit-width reductions.

\subsection{Implementation Details}
\label{sec:app_implementation_details}
\paragraph{Token Pruning}
To ensure fair comparison of token pruning methods, we standardize average token retention rates across layers. We evaluate rates of 1\%, 5\%, 10\%, 20\%, and 40\%. Higher rates are omitted, as 40\% retention typically preserves performance comparable to the original model.
Additionally, because the metrics of these methods can be architecture-dependent, we adapt each method consistently across all evaluated LVLMs to maintain fair comparisons. 

\paragraph{KV Cache Compression}
For fair comparison of KV cache compression methods, we standardize budget allocation. We test budgets of 1\%, 5\%, 10\%, 20\%, and 40\% of the original cache size. Higher budgets are omitted as 40\% typically yields performance comparable to the uncompressed model. During prefill, 10\% of the currently allocated budget forms a \textit{recent window} to retain the most recent tokens; the remainder is managed by each method’s specific mechanism. In the decoding, to maintain uniformity and fairness across all methods, we do not apply any additional KV cache compression.

\paragraph{Model Pruning}
In our experiments, we focus on pruning the LLM component. We utilize a consistent set of 128 samples from the COCO-Caption\cite{Chen2015MicrosoftCC} as our validation set. We apply unstructured pruning at 20\% and 50\% sparsity levels, as well as semi-structured pruning under 2:4 setting. We do not apply structured pruning because these result in significant performance drops without recovery training~\citep{wang2024cfsp}.

\paragraph{Model Quantization}
We focus exclusively on quantizing the LLM component, aligning with the existing LVLM quantization work. We utilize the same validation dataset as in our pruning experiments. For the LLM's weights, we apply a W4A16g128 quantization scheme, where weights are quantized to 4 bits and activations are kept in FP16, and a group size of 128 is used during quantization. This scheme offers a balanced trade-off between model performance and efficiency. 

\paragraph{Evaluation}
We use lmms-eval~\cite{zhang2024lmmsevalrealitycheckevaluation} framework to perform evaluation. Given the diversity of benchmarks and LVLMs, we select specific LVLMs and benchmarks tailored to each method type, which are detailed in Appendix~\ref{sec:appendix_evaluation_benchmarks}. The batch size is set to 1.
We use 8 $\times$ NVIDIA A800 GPUs. For the efficiency experiments, we run 128 samples with an 8k input token length and 100 token output length on a single NVIDIA A800 GPU.

\begin{figure*}[t!]
\centering
\includegraphics[width=0.8\textwidth]{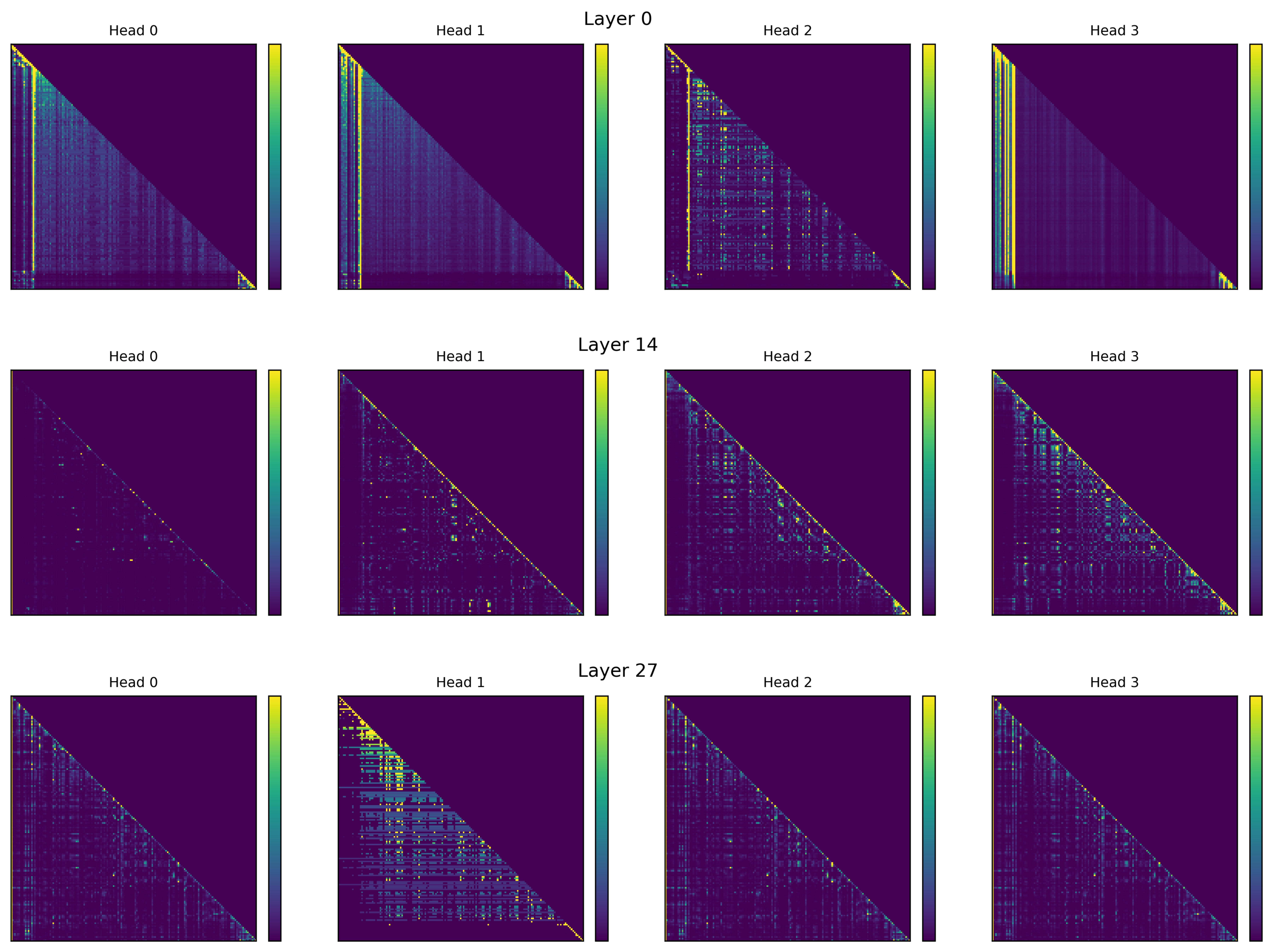}
\caption{Visualizations of attention weight heatmaps from layers 0, 14, and 27 of Qwen2-VL-7B. Different heads within the same layer exhibit notable pattern variations.
}\label{fig:qwen_attention_heatmap}
\end{figure*}

\subsection{Detailed Results of KV Cache Compression}
\label{sec:appendix_main_kvcache}
Table~\ref{tab:kv_cache_results} presents the detailed results of various KV cache compression methods across different models and benchmarks under varying cache budgets.

\begin{table*}[!t] 
\centering
\scriptsize 
\resizebox{0.8\linewidth}{!}{%
\begin{tabular}{c|c|c|c|c|c|c}
\toprule
\textbf{Budgets} & \textbf{Methods} & \textbf{TextVQA} & \textbf{AI2D} & \textbf{MMMU} & \textbf{MME} & \textbf{MathVista} \\
\midrule
\multirow{3}{*}{1\%}
    & VisionZip  & \textbf{44.48} & \textbf{72.11} & 42.56  & \textbf{1704.2} & 39.30 \\
    & iLLaVA & 22.86 & 68.78 & \textbf{43.11}  & 1101.0 & 37.50\\
    & SparseVLM & 42.29 & 71.07 & 43.00  & 1662.5 &\textbf{39.40} \\
\cmidrule(lr){1-7}
\multirow{3}{*}{10\%}
    & VisionZip  & 57.26 & 77.97 & 44.11  & \textbf{1915.0} &  \textbf{45.00} \\
    & iLLaVA & 64.88 & 73.47 &44.66  & 1773.9 & 39.30 \\
    & SparseVLM & \textbf{65.72} & \textbf{79.27} & \textbf{45.22}  & 1841.8 & 44.60 \\
\cmidrule(lr){1-7}
\multirow{3}{*}{40\%}
    & VisionZip  & 68.21 & 83.84 &  46.11  & 1956.8 & 52.60 \\
    & iLLaVA & \textbf{73.26} & 87.62 & \textbf{47.44}  & \textbf{1995.3} & \textbf{56.60} \\
    & SparseVLM & 72.80 & \textbf{88.66} & 45.78  & 1993.8 & 54.10 \\
\cmidrule(lr){1-7}
100\% & Original & 74.79 & 89.96 & 45.44 & 1974.1 & 58.20 \\
\bottomrule
\end{tabular}
}
\caption{Results of iLLaVA and SparseVLM on LLaVA-OneVision-7B model. \textbf{Bold} denotes the best result under the same setting.}
\label{tab:new_token_prune}
\end{table*}

\subsection{More Results of Token Compression}
We additionally evaluate two recent token compression approaches: iLLaVA~\citep{hu2024iLLaVA} and SparseVLM~\citep{zhang2024sparsevlm}, which incorporate \textit{progressive sparsification strategy} that gradually sparsifies tokens across multiple layers, rather than enforcing a target sparsity ratio at a single layer.
We follow the original implementation details and the results against existing methods in our benchmark are shown in Table~\ref{tab:new_token_prune}.

We find that: (i) iLLaVA demonstrates competitive performance at higher token budgets (40\%) thanks to its iterative pruning strategy within the visual encoder. This strategy progressively prunes tokens across layers while preserving critical ones in earlier stages. However, at lower token budgets (1\%), performance declines significantly similar as the phenomenon of FastV, likely due to the visual sink token: relying on text tokens to guide pruning within LLM may inadvertently remove essential visual sink tokens, leading to performance degradation (details are in Section~\ref{sec:sink_token}).
(ii) Regarding SparseVLM, compared to other token pruning methods that also exclusively prune visual tokens in the LLM component (e.g., FastV), it introduces an iterative pruning strategy and a more fine-grained, text-driven selection criterion to guide the pruning process. Consequently, SparseVLM achieves consistently strong performance across various compression budgets and tasks.

\begin{table*}[t!]
\centering
\small
\resizebox{\linewidth}{!}{
\begin{tabular}{c|l|*{12}{c}|c}
\toprule
\raisebox{5ex}{\textbf{Settings}} & \raisebox{5ex}{\textbf{Methods}} & 
\rotatebox{90}{ChartQA} & \rotatebox{90}{TextVQA} & 
\rotatebox{90}{OCRBench} & \rotatebox{90}{AI2D} & \rotatebox{90}{GQA} & 
\rotatebox{90}{MMMU} & \rotatebox{90}{MME} & \rotatebox{90}{RealworldQA} & 
\rotatebox{90}{MMStar} & \rotatebox{90}{MathVista} & \rotatebox{90}{LLaVA-Wilder} & 
\rotatebox{90}{ImageDC} & 
\raisebox{5ex}{\textbf{OP}}\\
\midrule
\multirow{1}{*}{100\%} 
  & Original    &  81.56 & 81.82 & 813 & 91.02 & 62.3 & 50.77  & 2327.78 & 66.53 & 57.11 & 58.3  & 72.7  & 86.35  & 1.000 \\
\midrule
\multirow{3}{*}{20\%} 
  & EcoFLAP    &  81.36 & 81.82 & 805 & 91.13 & 62.30 & 50.00 & 2,288.0 & 66.67 & 56.74 & 57.10 & 52.70 &  87.33  &  0.964\\
  & Wanda      &  75.52 & 80.64 & 754 & 89.67 & 60.86 & 46.11 & 2,090.8 & 65.88 & 53.02 & 52.40 & 50.20 & 84.90 & 0.918 \\
  & SparseGPT  &  71.68 & 77.45 & 722 & 82.32 & 57.84 & 39.56 & 1,871.9 & 59.61 & 45.99 & 40.70 & 44.50 &   81.00 & 0.842 \\

\midrule
\multirow{3}{*}{50\%} 
  & EcoFLAP    &  81.08 & 81.79 & 800 & 91.16 & 62.32 & 50.44 & 2,312.6 & 66.54 & 56.69 & 58.90 & 53.50 &  87.13 & 0.970  \\

  & Wanda      &   77.72 & 80.61 & 784 & 88.63 & 61.20 & 45.00 & 2,129.1 & 64.97 & 52.74 & 53.50 & 51.20 &  85.55 &0.924  \\

  & SparseGPT  & 71.96 & 78.08 & 727 & 82.29 & 58.66 & 38.78 & 1,967.3 & 57.91 & 45.63 & 39.30 & 44.50 &   80.30 &0.845  \\

\midrule
\multirow{3}{*}{2:4} 
  & EcoFLAP    &  81.16 & 82.08 & 809 & 91.03 & 62.32 & 50.78 & 2,317.2 & 67.45 & 56.85 & 59.30 & 54.80 &  88.37 &  0.977\\

  & Wanda      &  73.96 & 80.09 & 790 & 88.44 & 61.54 & 44.89 & 2,106.9 & 65.10 & 53.42 & 53.10 & 52.40 & 86.00 & 0.924 \\

  & SparseGPT  & 61.04 & 76.60 & 721 & 81.19 & 58.24 & 36.89 & 1,675.00 & 59.22 & 42.31 & 40.40 & 48.70 &   85.25 & 	0.832  \\

\bottomrule
\end{tabular}
}
\caption{Results of various parameter compression methods on Qwen2-VL-7B.
}
\label{tab:qwen2vl_compression}
\end{table*}

\subsection{More Results of Parameter Compression}
\label{sec:app_more_params_compression}
Table~\ref{tab:qwen2vl_compression} shows the results of parameter compression methods on Qwen2-VL-7B across various benchmarks. 
To investigate the efficacy of combining token-level and parameter-level compression, we select AWQ and SnapKV as representative methods for token and parameter compression, respectively. The results on Qwen2-VL-7B are shown in Table~\ref{tab:combination}.
Combining AWQ with SnapKV boosts inference speed by 1.65$\times$ over AWQ alone, while maintaining comparable overall performance, particularly on multimodal reasoning tasks (MMMU, MathVista). This highlights the efficiency gains achievable with hybrid compression.
However, for tasks demanding fine-grained visual perception (e.g., TextVQA), integrating token compression like SnapKV results in a marginal performance drop. This underscores that acceleration strategies must be tailored to specific target scenario characteristics.
\begin{table*}[t!]
    \centering
    \small 
    \resizebox{0.8\linewidth}{!}{
    \begin{tabular}{lcccccc}
        \toprule
        \textbf{Methods} & \textbf{Speedup} & \textbf{AI2D} & \textbf{MMMU} & \textbf{MME} & \textbf{TextVQA} & \textbf{MathVista} \\
        \midrule
        AWQ        & 1.0x  & 90.80 & 57.20 & 2320.6 & 50.26 & 66.53 \\
        AWQ+SnapKV & 1.65x & 90.74 & 57.40 & 2235.9 & 48.78 & 66.80 \\
        \bottomrule
        \end{tabular}%
    }
    \caption{Results on Qwen2-VL-7B on combining token and parameter compression.}
    \label{tab:combination}
\end{table*}

\subsection{Architectural Impact on Compression Effectiveness}
We analyze how architectural differences in LVLM backbones (LLaVA-OneVision, Qwen2-VL, InternVL-2.5) affect compression effectiveness. Two primary aspects are considered:

\begin{table*}[t!]
\centering
\small
\resizebox{0.8\linewidth}{!}{
\begin{tabular}{c|ll|*{5}{c}|c}
\toprule
\raisebox{3ex}{\textbf{Settings}} & \raisebox{3ex}{\textbf{Method}} & \raisebox{3ex}{\textbf{Size}} & 
\rotatebox{90}{AI2D} & \rotatebox{90}{MathVista} & 
\rotatebox{90}{MME} & \rotatebox{90}{MMMU} & 
\rotatebox{90}{TextVQA} & 
\raisebox{3ex}{\textbf{OP}}\\
\midrule
\multirow{2}{*}{100\%} 
  & \multirow{2}{*}{Original}  
    & 2B & 83.83 & 46.20 & 1872.0 & 40.85 & 79.70 & 1.0000 \\
  &                             
    & 7B & 91.02 & 58.30 & 2327.7 & 50.77 & 81.82 & 1.0000 \\
\midrule

\multirow{6}{*}{20\%} 
  & \multirow{2}{*}{ECOFLAP}   
    & 2B & 80.99 & 44.00 & 1883.4 & 39.44 & 79.19 & 0.9769 \\
  &                             
    & 7B & 91.13 & 57.10 & 2288.0 & 50.00 & 81.82 & 0.9900 \\ 
  \cmidrule(lr){2-9} 
  & \multirow{2}{*}{Wanda}     
    & 2B & 81.99 & 45.30 & 1884.5 & 39.89 & 79.31 & 0.9874 \\
  &                             
    & 7B & 91.16 & 58.90 & 2312.6 & 50.44 & 81.79 & 1.0000 \\ 
  \cmidrule(lr){2-9} 
  & \multirow{2}{*}{SparseGPT} 
    & 2B & 82.35 & 46.30 & 1852.9 & 40.00 & 79.44 & 0.9901 \\
  &                             
    & 7B & 91.03 & 59.30 & 2317.2 & 50.78 & 82.08 & 1.0030 \\
\midrule

\multirow{6}{*}{50\%} 
  & \multirow{2}{*}{ECOFLAP}   
    & 2B & 71.99 & 33.50 & 1585.2 & 34.44 & 76.39 & 0.8497 \\
  &                             
    & 7B & 89.67 & 52.40 & 2090.8 & 46.11 & 80.64 & 0.9356 \\ 
  \cmidrule(lr){2-9}
  & \multirow{2}{*}{Wanda}     
    & 2B & 72.83 & 34.90 & 1721.8 & 36.44 & 76.26 & 0.8812 \\
  &                             
    & 7B & 88.63 & 53.50 & 2129.1 & 45.00 & 80.61 & 0.9350 \\ 
  \cmidrule(lr){2-9}
  & \multirow{2}{*}{SparseGPT} 
    & 2B & 72.15 & 35.70 & 1535.1 & 33.56 & 75.75 & 0.8472 \\
  &                             
    & 7B & 88.44 & 53.10 & 2106.9 & 44.89 & 80.09 & 0.9309 \\
\midrule

\multirow{6}{*}{2:4} 
  & \multirow{2}{*}{ECOFLAP}   
    & 2B & 44.79 & 28.20 & 1008.7 & 27.11 & 66.12 & 0.6445 \\
  &                             
    & 7B & 82.32 & 40.70 & 1871.9 & 39.56 & 77.45 & 0.8313 \\ 
  \cmidrule(lr){2-9}
  & \multirow{2}{*}{Wanda}     
    & 2B & 47.83 & 27.70 & 1207.2 & 26.89 & 67.01 & 0.6695 \\
  &                             
    & 7B & 82.29 & 39.30 & 1967.3 & 38.78 & 78.08 & 0.8336 \\ 
  \cmidrule(lr){2-9}
  & \multirow{2}{*}{SparseGPT} 
    & 2B & 49.92 & 26.20 & 1027.8 & 29.22 & 66.91 & 0.6624 \\
  &                             
    & 7B & 81.19 & 40.40 & 1675.0 & 36.89 & 76.60 & 0.8000 \\
\midrule

\multirow{2}{*}{W4A16} 
  & \multirow{2}{*}{AWQ}       
    & 2B & 82.90 & 46.50 & 1835.4 & 39.82 & 79.34 & 0.9893 \\
  &                             
    & 7B & 90.80 & 58.30 & 2232.5 & 48.78 & 81.31 & 0.9826 \\
\bottomrule
\end{tabular}%
}
\caption{Results of various parameter compression methods on different model sizes of Qwen2-VL.}
\label{tab:diff_arch_compress}
\end{table*}

\begin{itemize}
    \item \textbf{High-Resolution Image Processing.} LLaVA-OneVision and InternVL-2.5 use \textit{anyres}-like strategies, splitting high-resolution images into many patches, averaging 4180 visual tokens per input on our benchmark. In contrast, Qwen2-VL employs NaViT, a visual encoder with native dynamic-resolution support that merges visual tokens, reducing their average count to 762 (see \citet{dehghani2023patch} for NaViT details).
    This difference in initial visual token count directly impacts \textit{token compression}. For example, LLaVA-OneVision-7B consistently outperforms similarly sized Qwen2-VL-7B in various token compression budgets (Table~\ref{tab:token_length}). This suggests that a larger pool of initial visual tokens offers greater flexibility for compression, as its richer representation can tolerate more aggressive reduction without substantial performance loss.
    \item \textbf{Parameter Size.} To isolate the impact of model scale, we evaluate Qwen2-VL-2B against Qwen2-VL-7B using various parameter compression methods. The results are shown in Table~\ref{tab:diff_arch_compress}.
\end{itemize}

\subsection{Visualization Head Attention Distribution}
\label{sec:appendix_head_attention_distribution}

To illustrate how head-adaptive mechanisms can capture diverse attention patterns, we analyzed attention distributions across heads in several randomly selected layers. Qwen2-VL-7B was chosen for this visualization because its relatively fewer image tokens enhance attention map readability. Indeed, Figure~\ref{fig:qwen_attention_heatmap} shows that different heads within the same layer exhibit distinct attention patterns.

\subsection{Visualizations on VL-Cache Budget Allocation}
\label{sec:additional-figures-vlcache-budget}
\begin{figure*}[ht]
  \centering
  \begin{minipage}[t]{0.32\linewidth}
    \centering
    \includegraphics[width=1\linewidth]{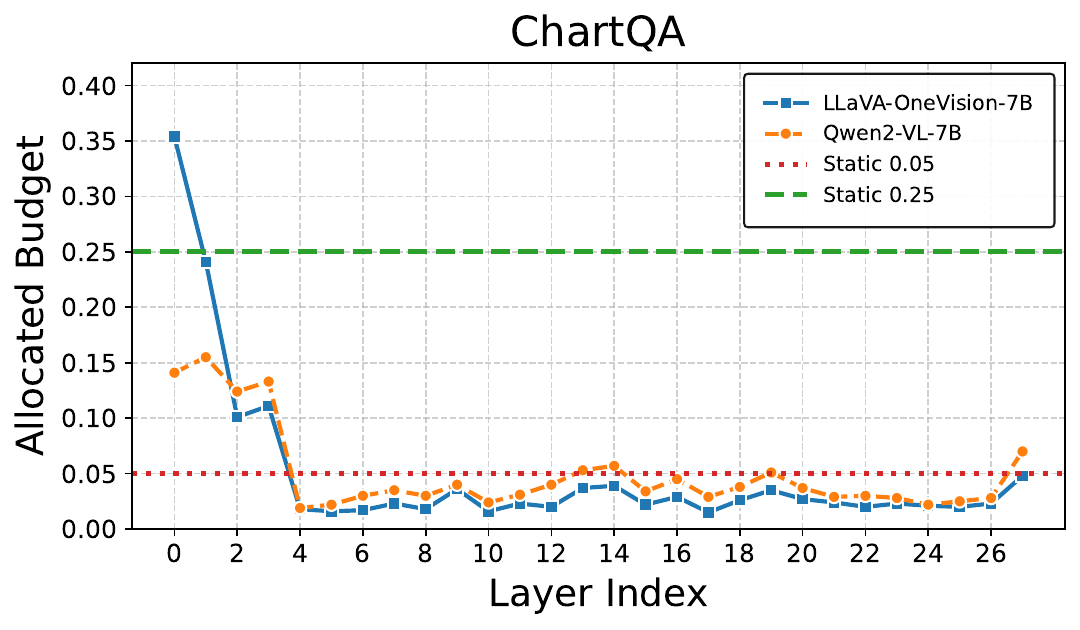}
  \end{minipage}
  \hfill
  \begin{minipage}[t]{0.32\linewidth}
    \centering
    \includegraphics[width=1\linewidth]{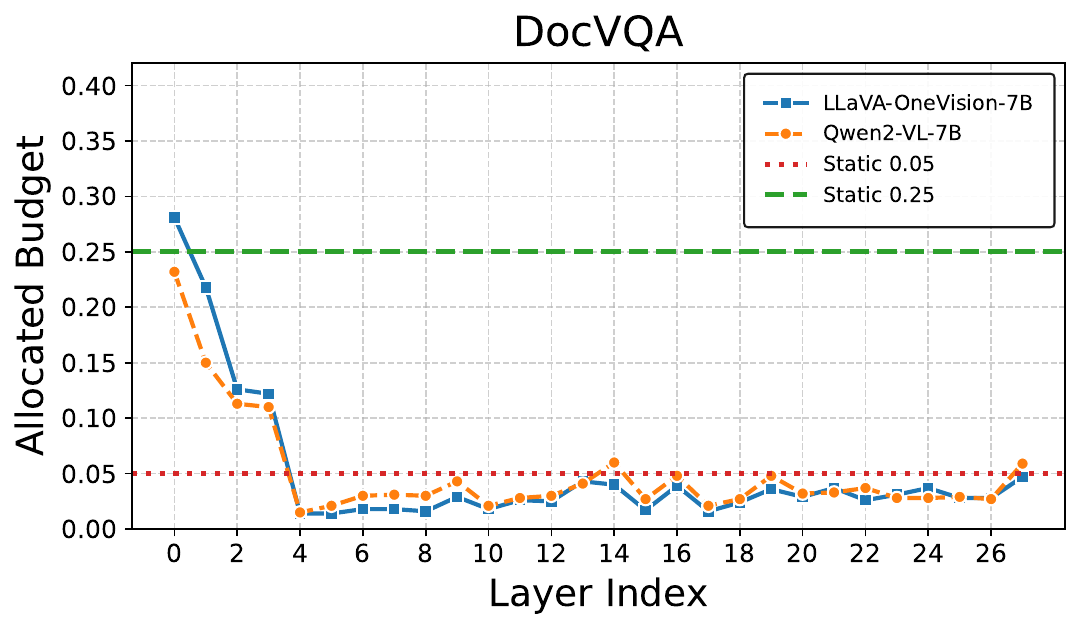}
  \end{minipage}
  \hfill
  \begin{minipage}[t]{0.32\linewidth}
    \centering
    \includegraphics[width=1\linewidth]{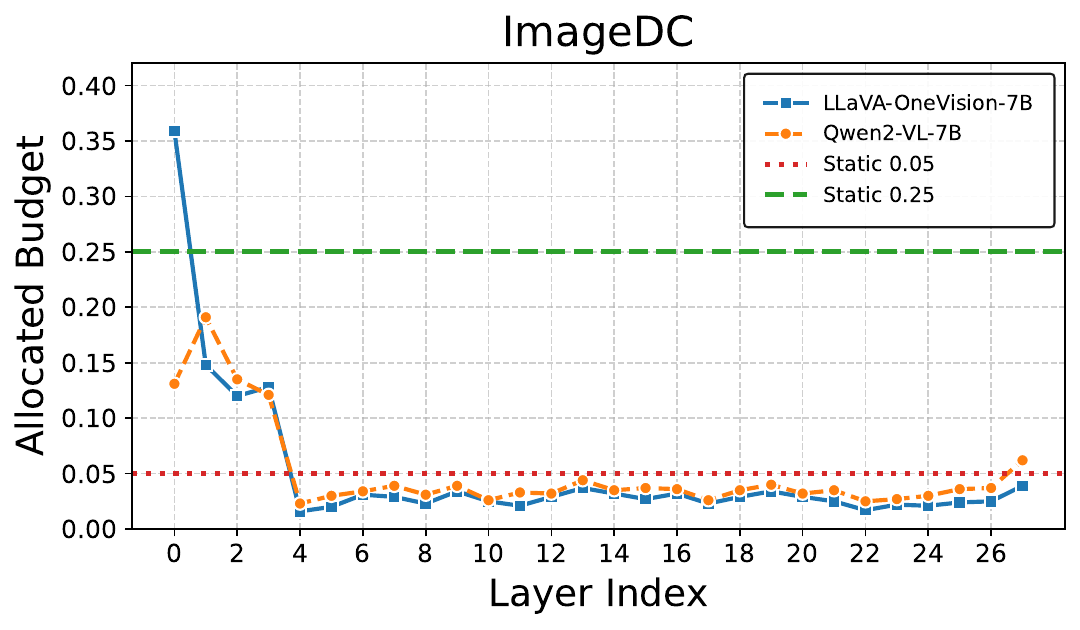}
  \end{minipage}
  \hfill
  \begin{minipage}[t]{0.32\linewidth}
    \centering
    \includegraphics[width=1\linewidth]{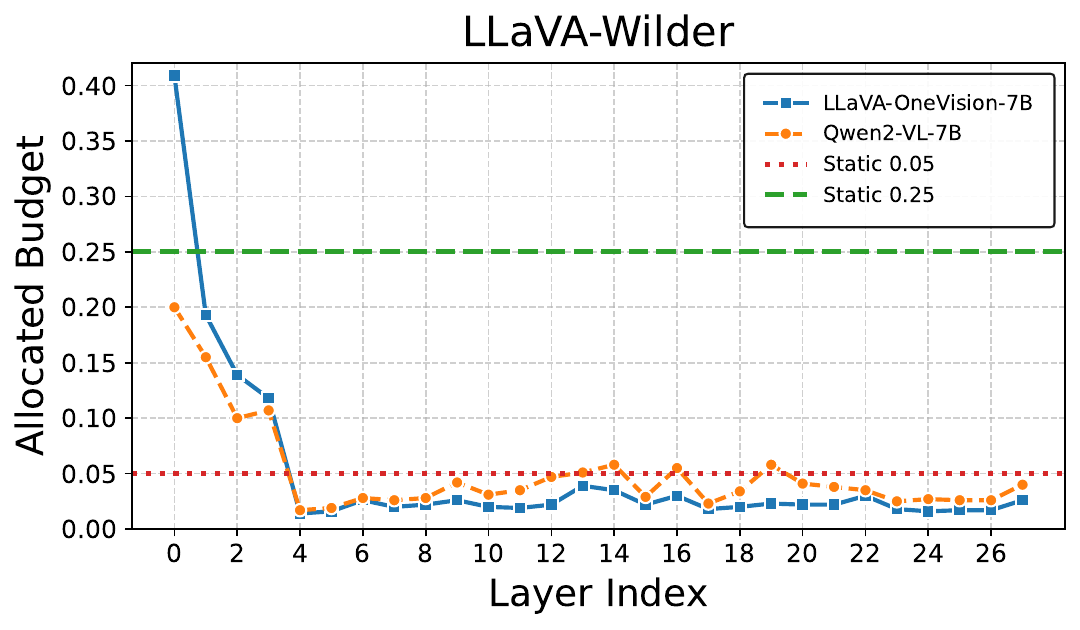}
  \end{minipage}
    \hfill
  \begin{minipage}[t]{0.32\linewidth}
    \centering
    \includegraphics[width=1\linewidth]{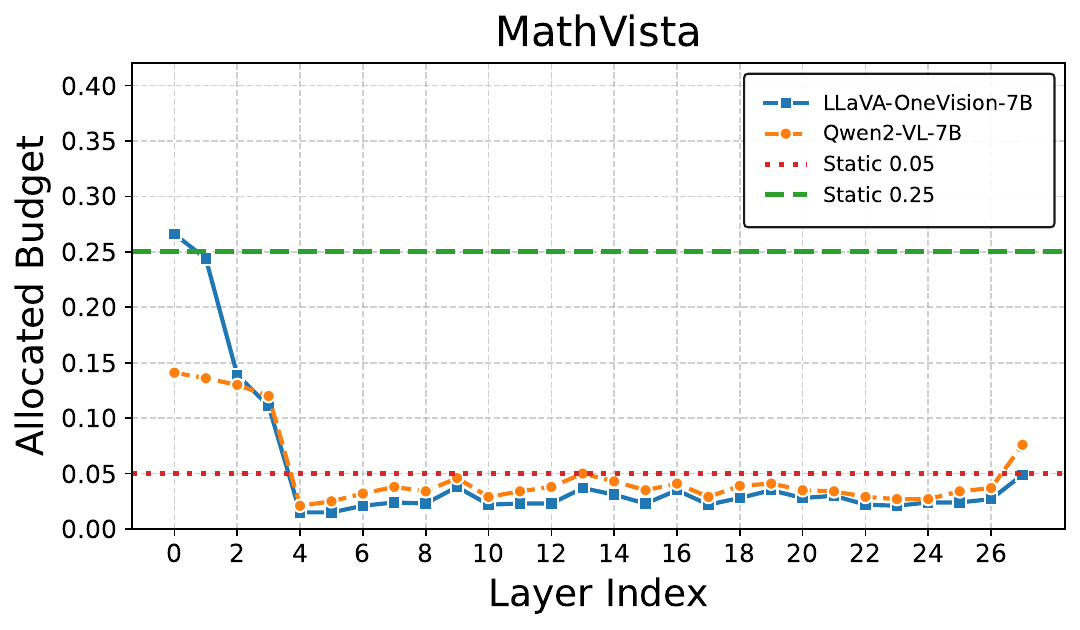}
  \end{minipage}
    \hfill
  \begin{minipage}[t]{0.32\linewidth}
    \centering
    \includegraphics[width=1\linewidth]{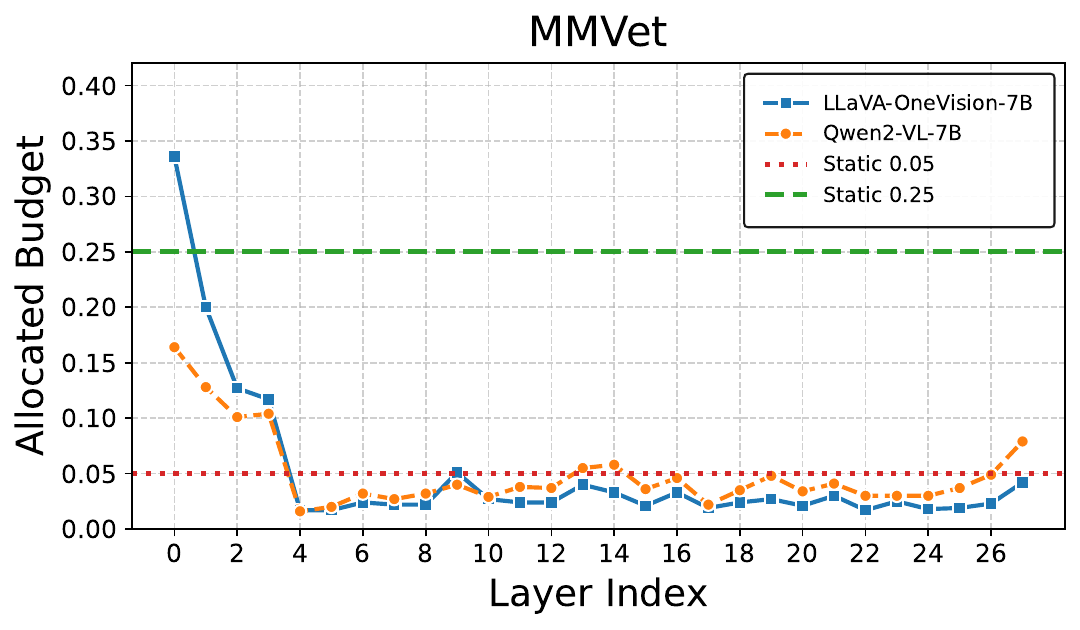}
  \end{minipage}
    \hfill
  \begin{minipage}[t]{0.32\linewidth}
    \centering
    \includegraphics[width=1\linewidth]{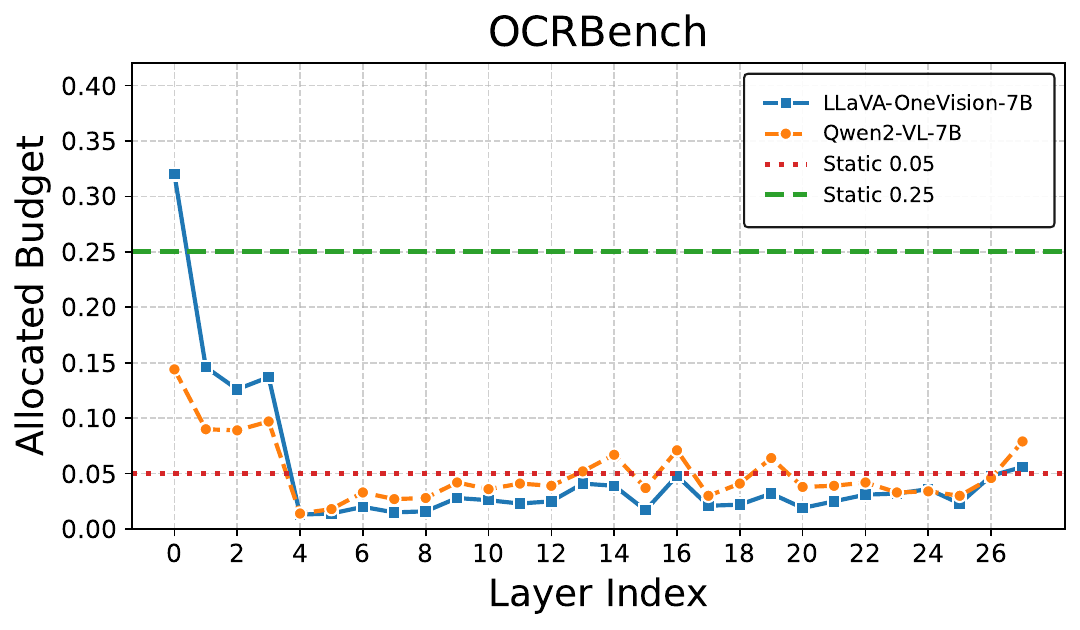}
  \end{minipage}
    \begin{minipage}[t]{0.32\linewidth}
    \centering
    \includegraphics[width=1\linewidth]{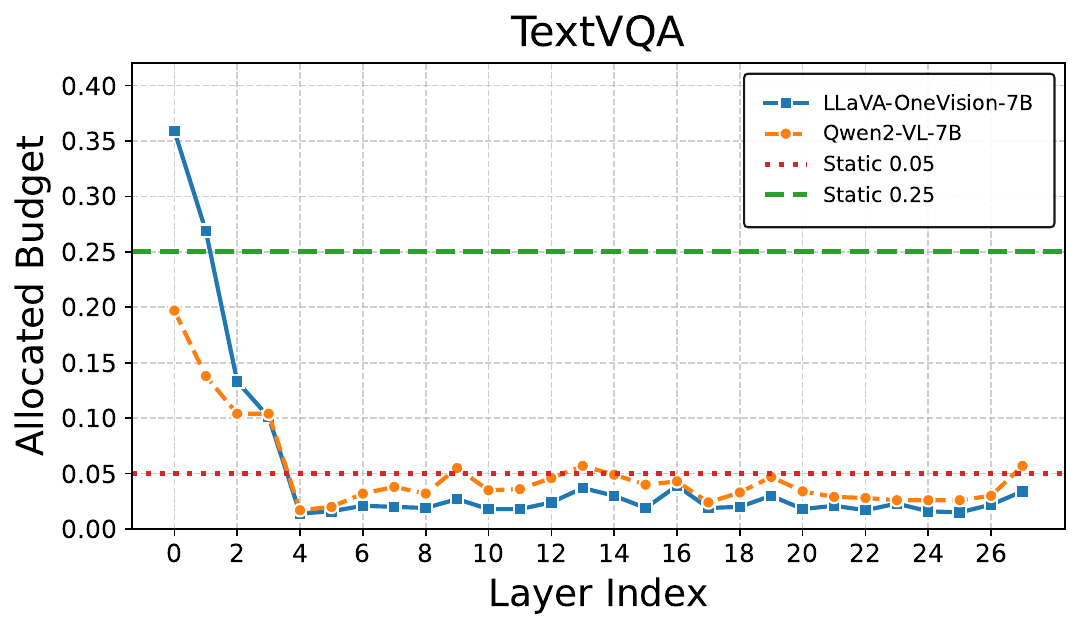}
  \end{minipage}
  
  \caption{Budgets distribution of LLaVA-OV-7B and Qwen2-VL-7B across 8 benchmarks under 5\% budget. Each subplot shows a random sample from the respective benchmark.}
  \label{fig:qwen_ov_budget}
\end{figure*}

\begin{figure*}[ht]
  \centering
  \begin{minipage}[t]{0.32\linewidth}
    \centering
    \includegraphics[width=1\linewidth]{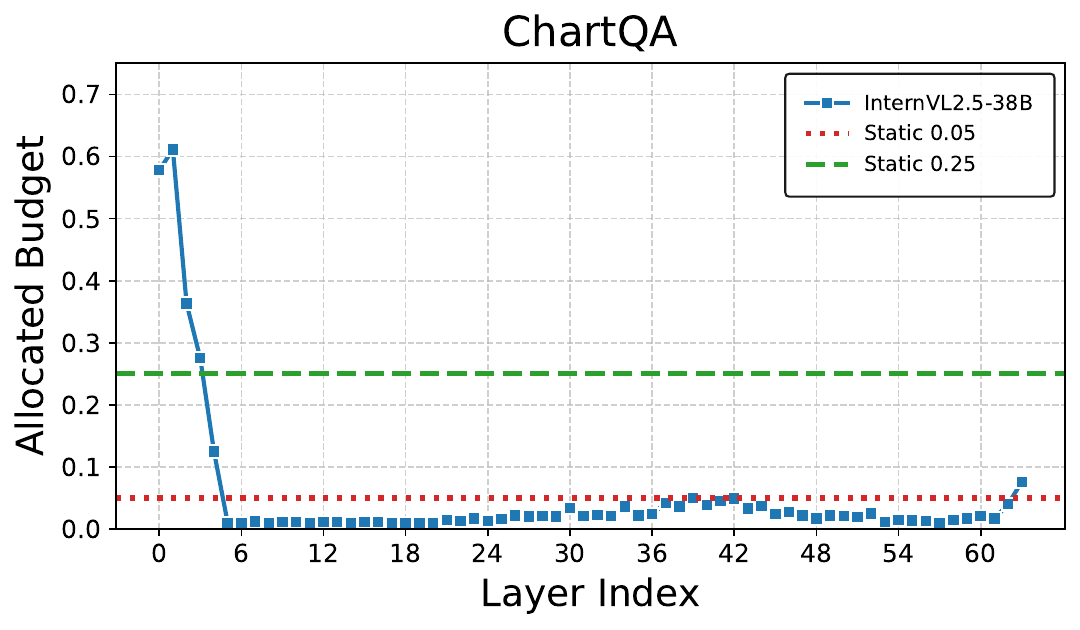}
  \end{minipage}
  \hfill
  \begin{minipage}[t]{0.32\linewidth}
    \centering
    \includegraphics[width=1\linewidth]{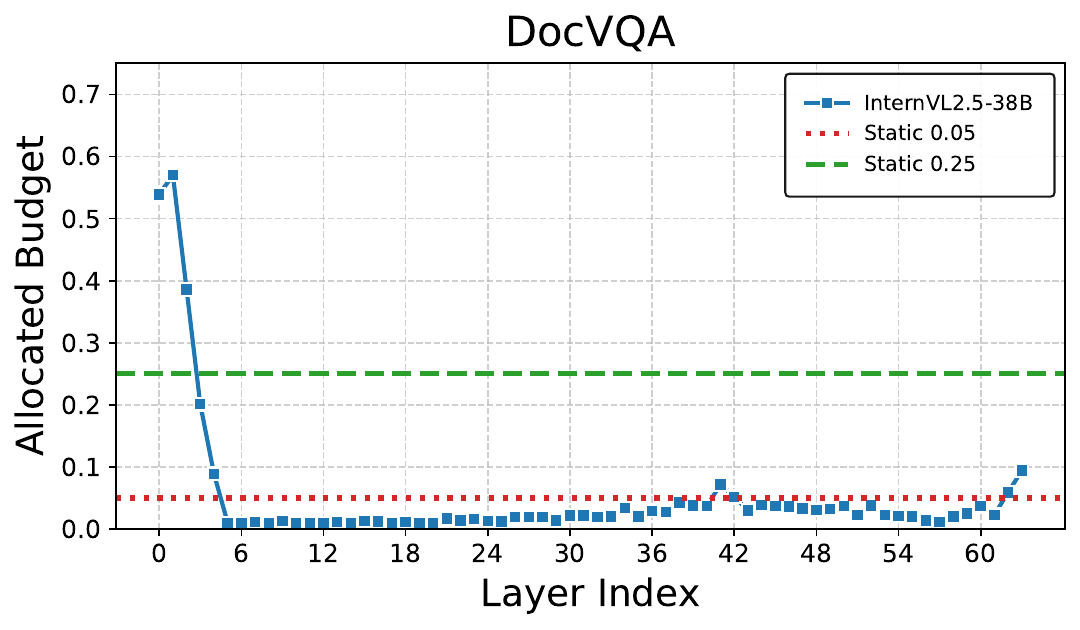}
  \end{minipage}
  \hfill
  \begin{minipage}[t]{0.32\linewidth}
    \centering
    \includegraphics[width=1\linewidth]{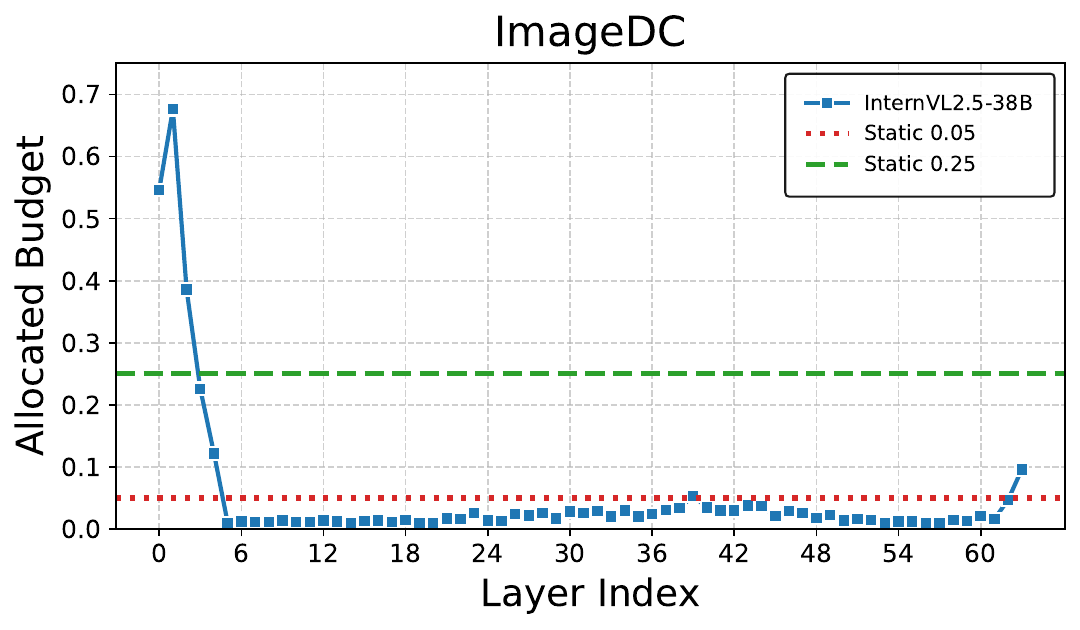}
  \end{minipage}
  \hfill
  \begin{minipage}[t]{0.32\linewidth}
    \centering
    \includegraphics[width=1\linewidth]{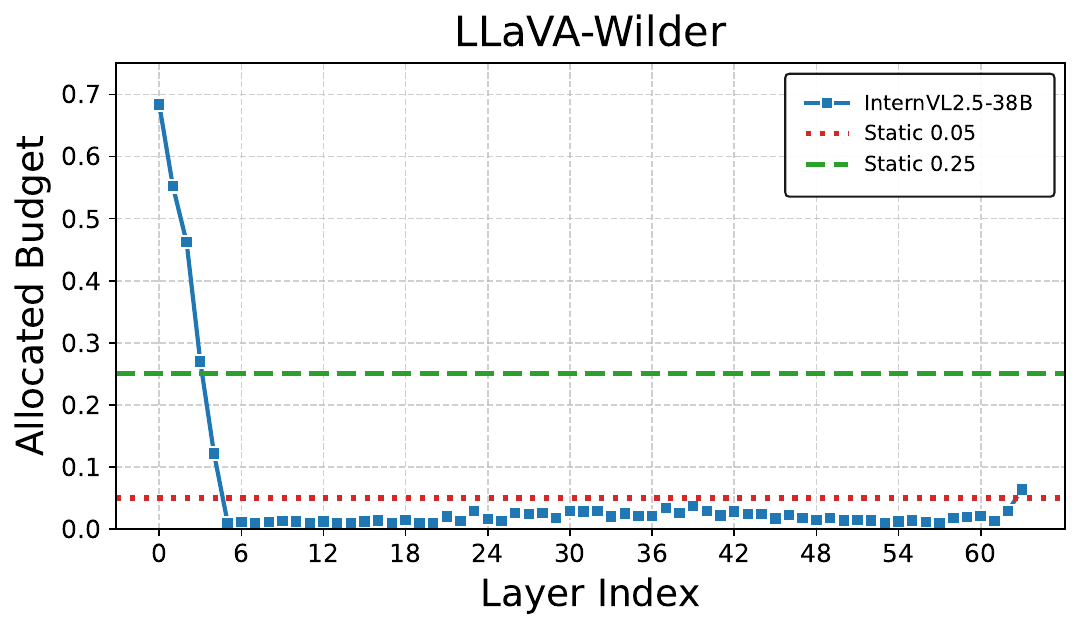}
  \end{minipage}
    \hfill
  \begin{minipage}[t]{0.32\linewidth}
    \centering
    \includegraphics[width=1\linewidth]{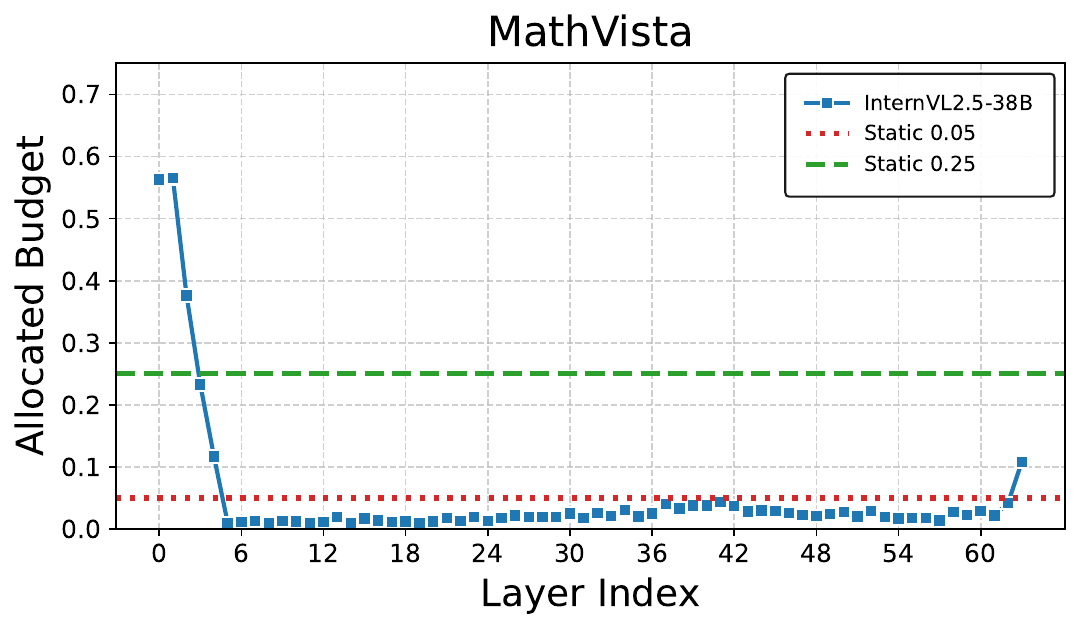}
  \end{minipage}
    \hfill
  \begin{minipage}[t]{0.32\linewidth}
    \centering
    \includegraphics[width=1\linewidth]{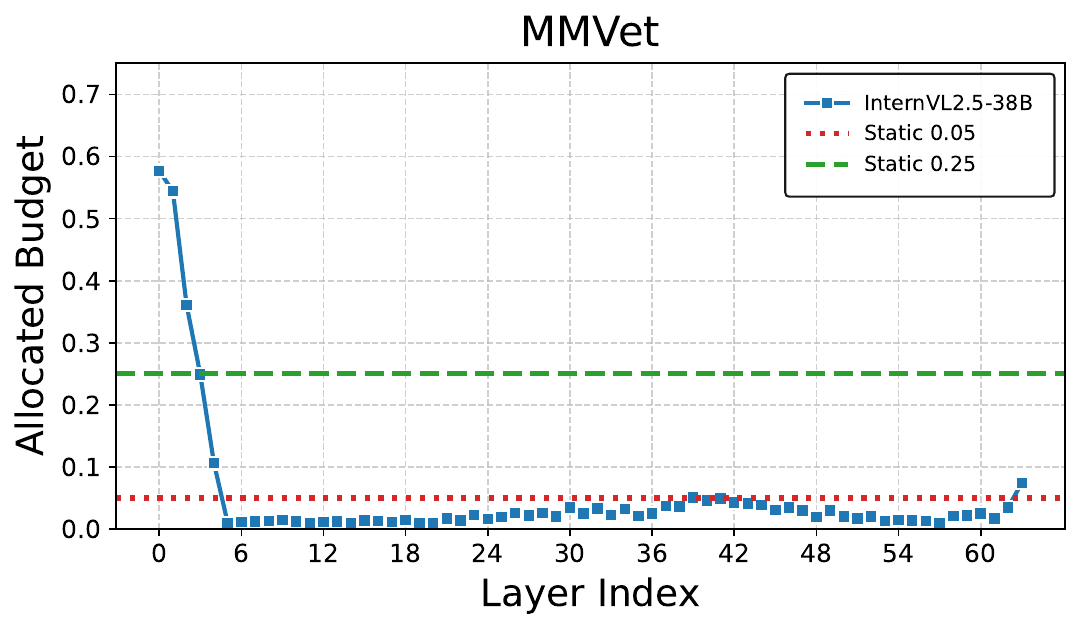}
  \end{minipage}
    \hfill
  \begin{minipage}[t]{0.32\linewidth}
    \centering
    \includegraphics[width=1\linewidth]{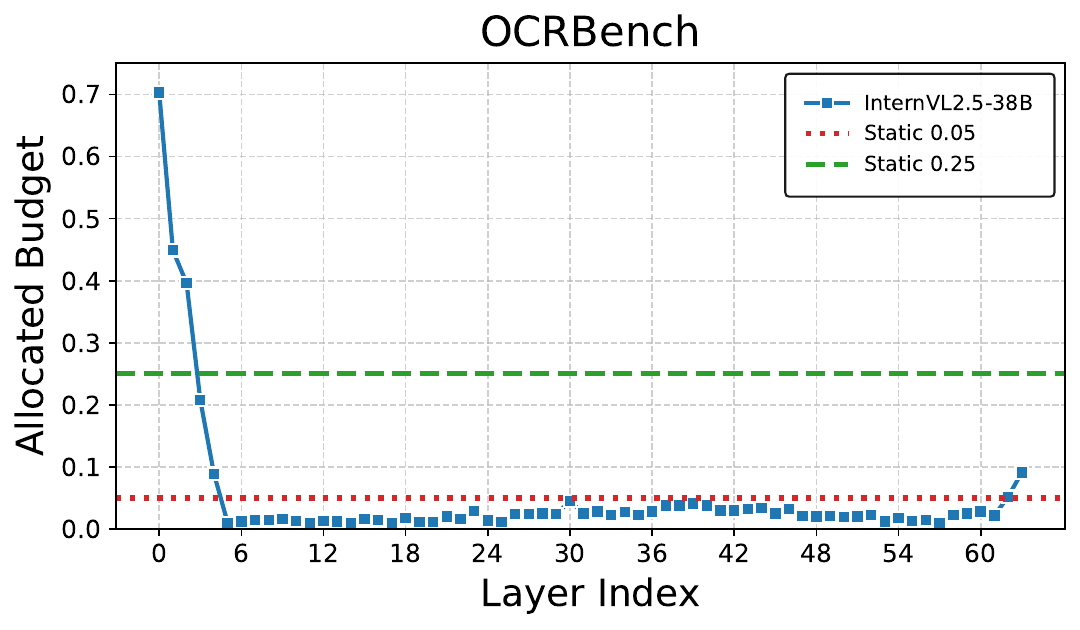}
  \end{minipage}
    \begin{minipage}[t]{0.32\linewidth}
    \centering
    \includegraphics[width=1\linewidth]{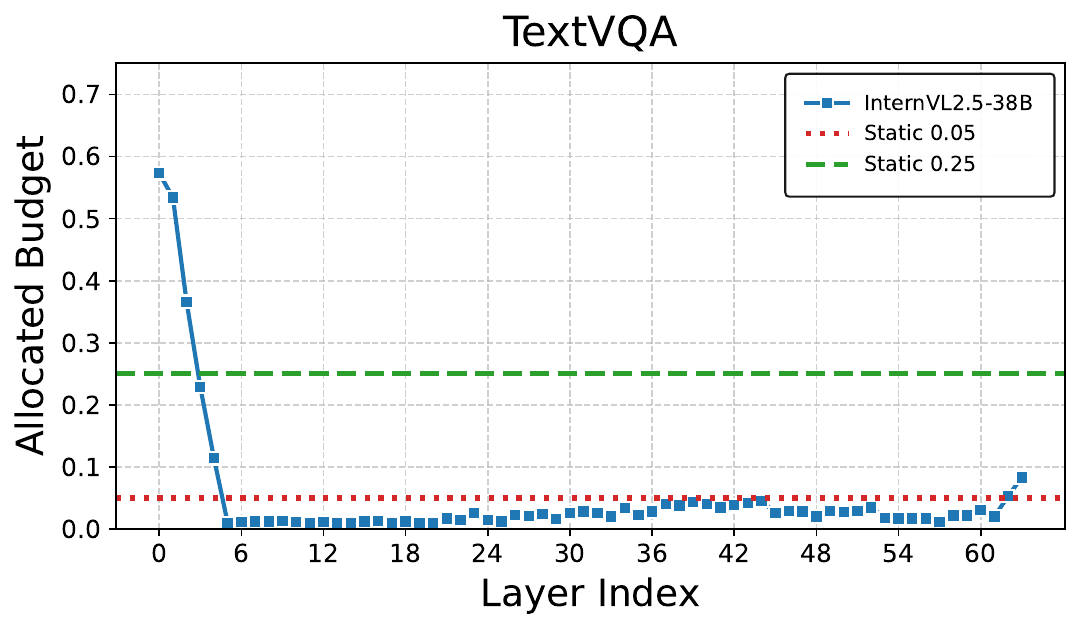}
  \end{minipage}
  
  \caption{Budgets distribution of InternVL2.5-38B across 8 benchmarks under 5\% budget. Each subplot shows a random sample from the respective benchmark.}
  \label{fig:38B_budget}
\end{figure*}

We evaluated VL-Cache on eight benchmarks across three LVLMs, all at a 5\% total budget.
The models consistently exhibited similar, highly skewed layer-wise budget allocations (Figures~\ref{fig:qwen_ov_budget} and~\ref{fig:38B_budget}). True to its design, VL-Cache's layer-adaptive mechanism heavily favored early ``dense'' layers, particularly the first two. Allocation minima were observed at layer~4 for LLaVA-OV-7B and Qwen2-VL-7B, and at layer~5 for InternVL2.5-38B, with subsequent layers typically receiving far below-average budgets.
This consistent front-loading suggests performance could be improved by reallocating budget from these over-resourced early layers to the under-resourced later ones. Effective layer-adaptive strategies therefore demand a more nuanced resource balance, rather than just aggressive front-loading.

\begin{figure*}[t!]
\centering
\includegraphics[width=0.8\textwidth]{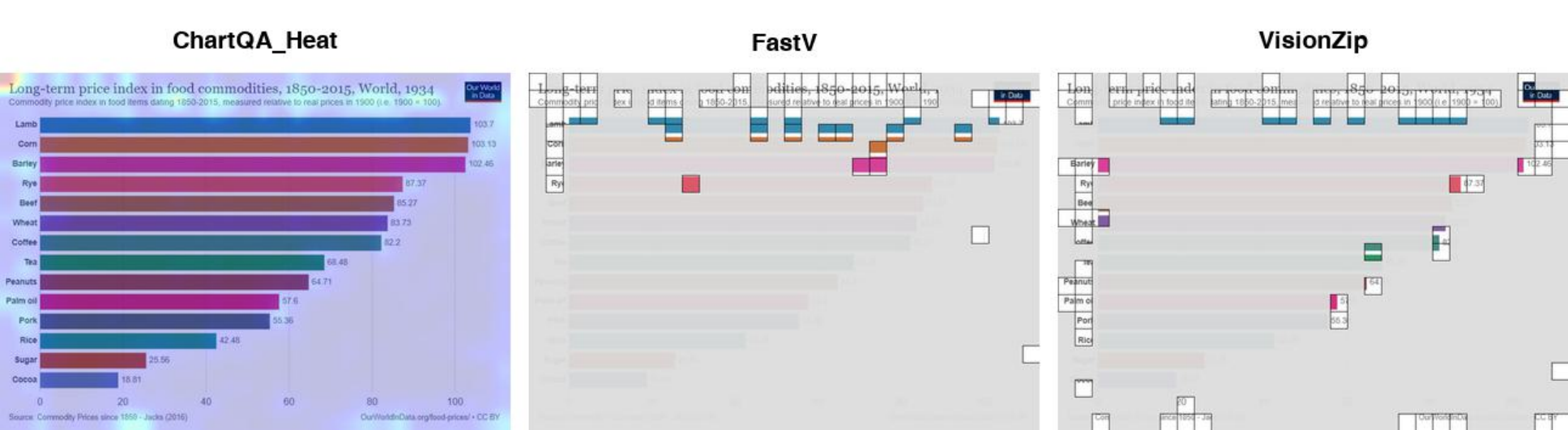}
\caption{
Visualization of token selection strategies on ChartQA with Qwen2-VL-7B under 10\% budget. Left to right: attention heatmap, tokens retained by FastV and VisionZip. VisionZip selects more critical tokens.
}\label{fig:chartqa}
\end{figure*}

\begin{figure*}[ht]
  \centering
  \begin{subfigure}[t]{\linewidth}
    \centering
    \begin{minipage}[t]{0.495\linewidth}
      \centering
      \includegraphics[width=1\linewidth]{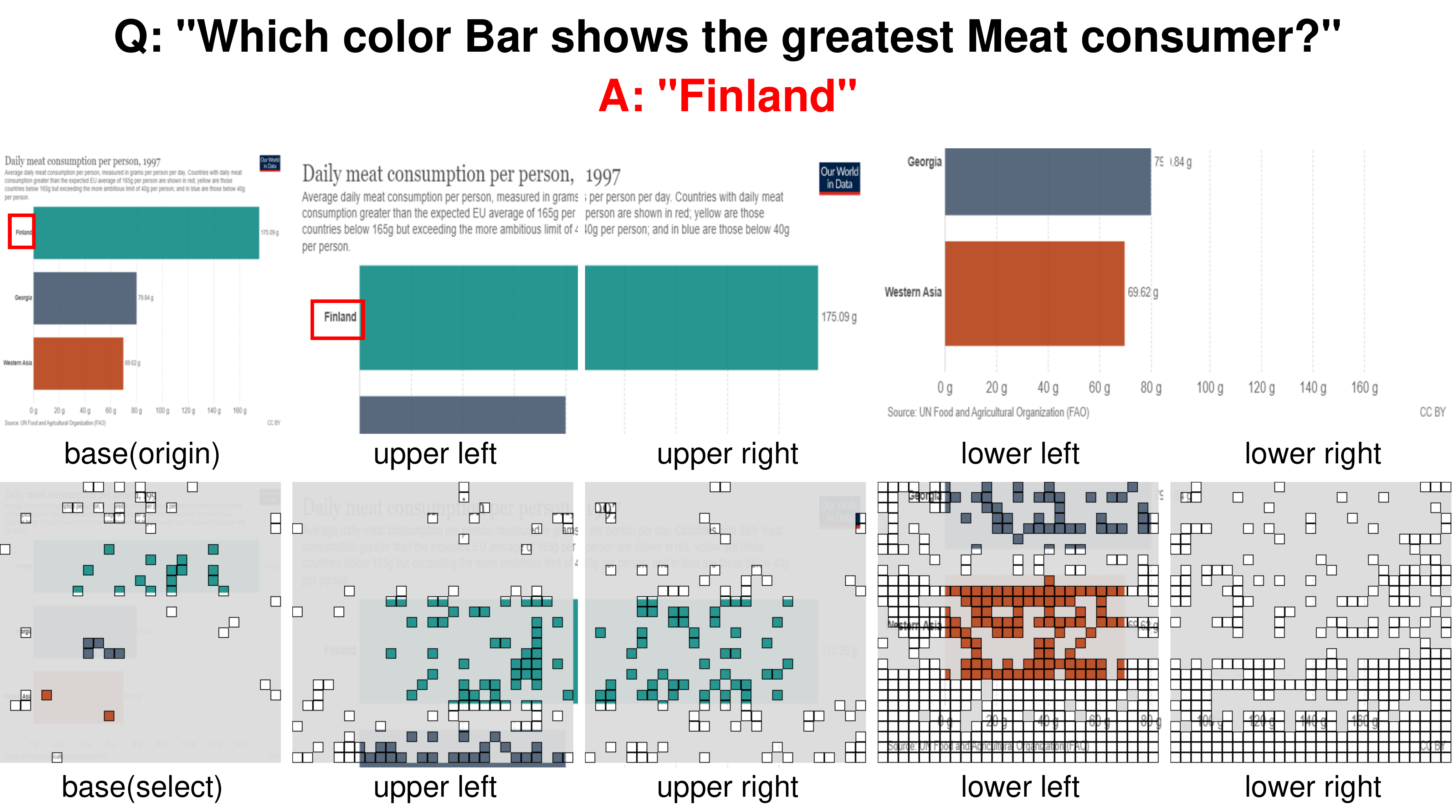}
    \end{minipage}
    \hfill
    \begin{minipage}[t]{0.495\linewidth}
      \centering
      \includegraphics[width=1\linewidth]{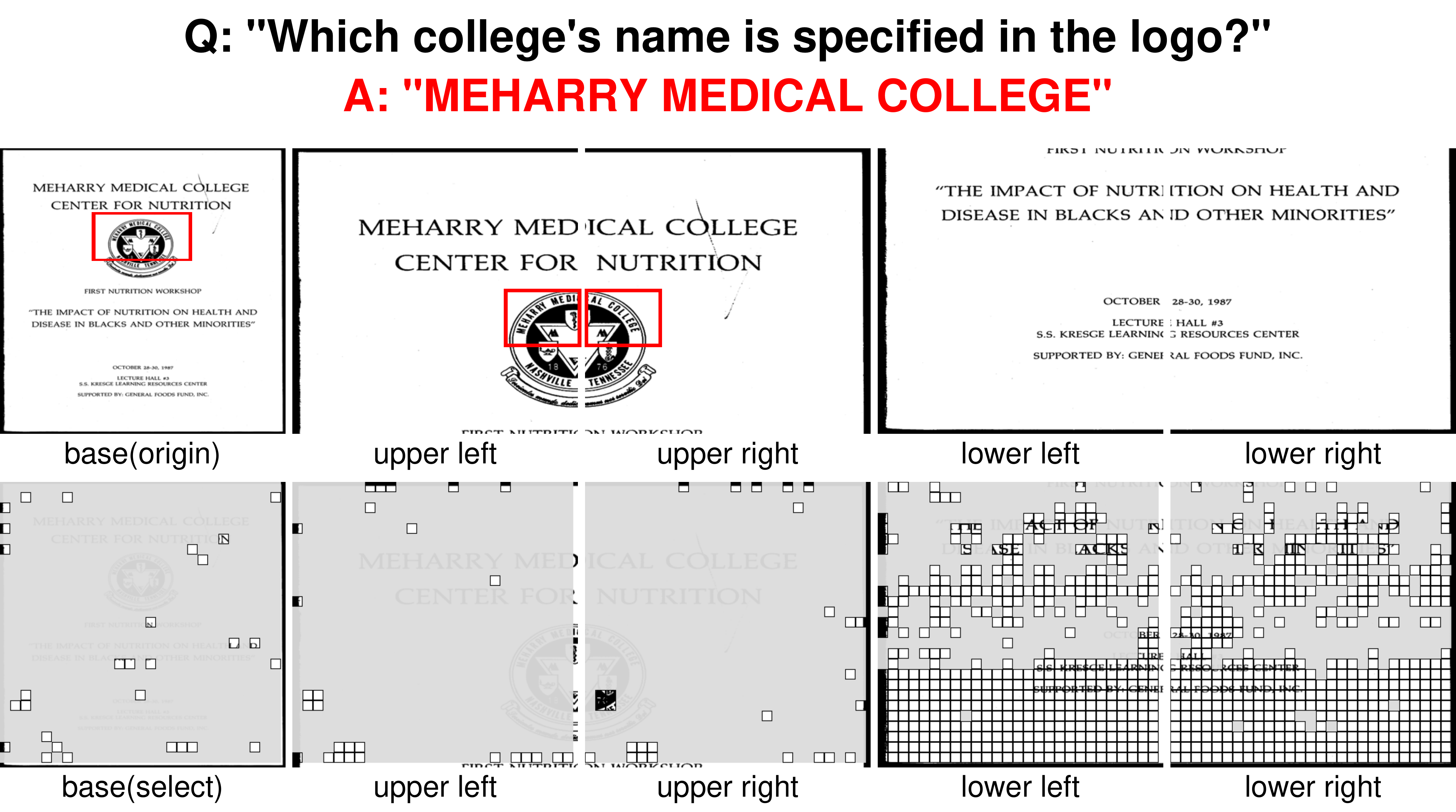}
    \end{minipage}
    \caption{Visualization of Layer 0 visual token selection: representative cases from ChartQA (left) and DocVQA (right)} 
    \label{fig:vlcache_case_row1}
  \end{subfigure}
  
  \vspace{10pt}
  
  \begin{subfigure}[t]{\linewidth}
    \centering
    \begin{minipage}[t]{0.495\linewidth}
      \centering
      \includegraphics[width=1\linewidth]{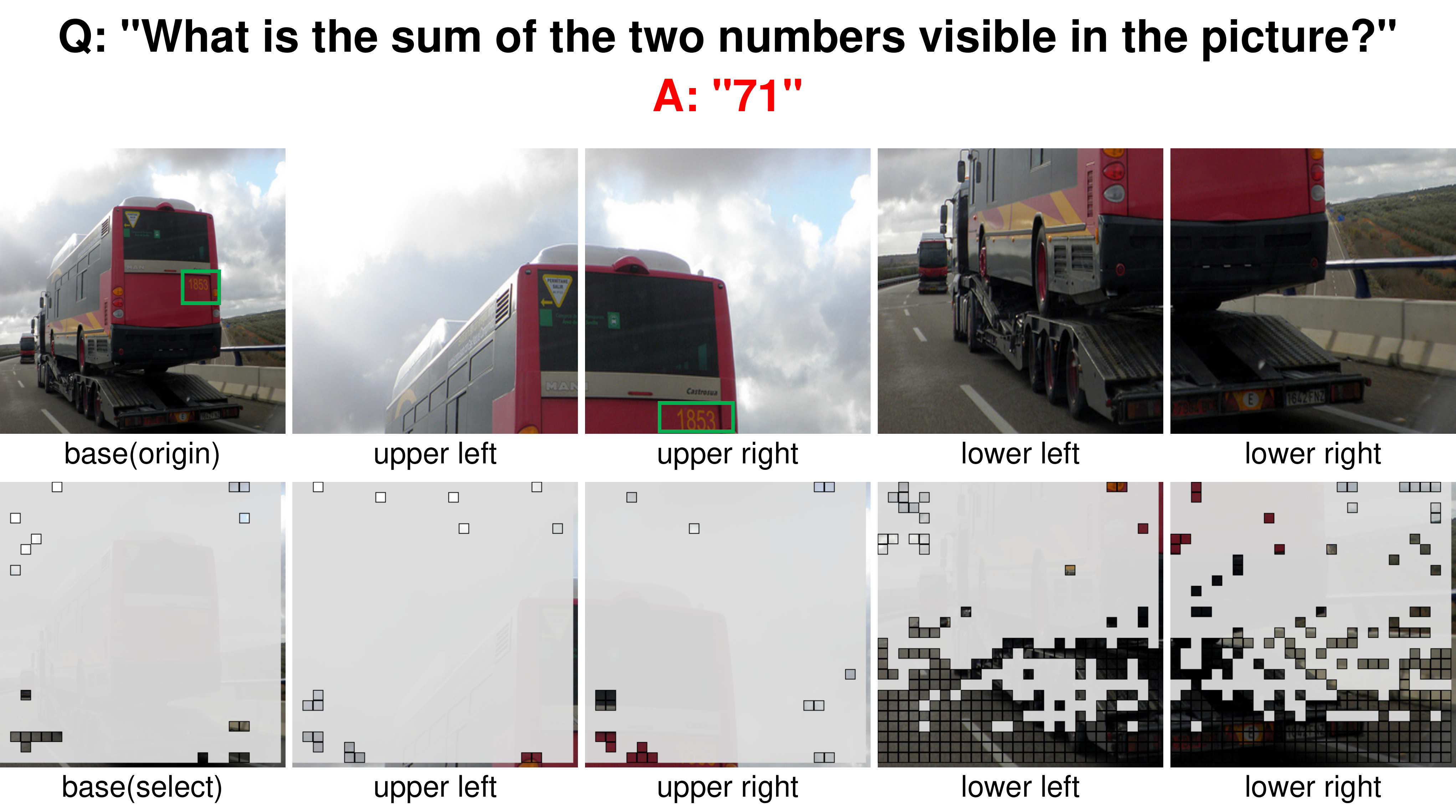}
    \end{minipage}
    \hfill
    \begin{minipage}[t]{0.495\linewidth}
      \centering
      \includegraphics[width=1\linewidth]{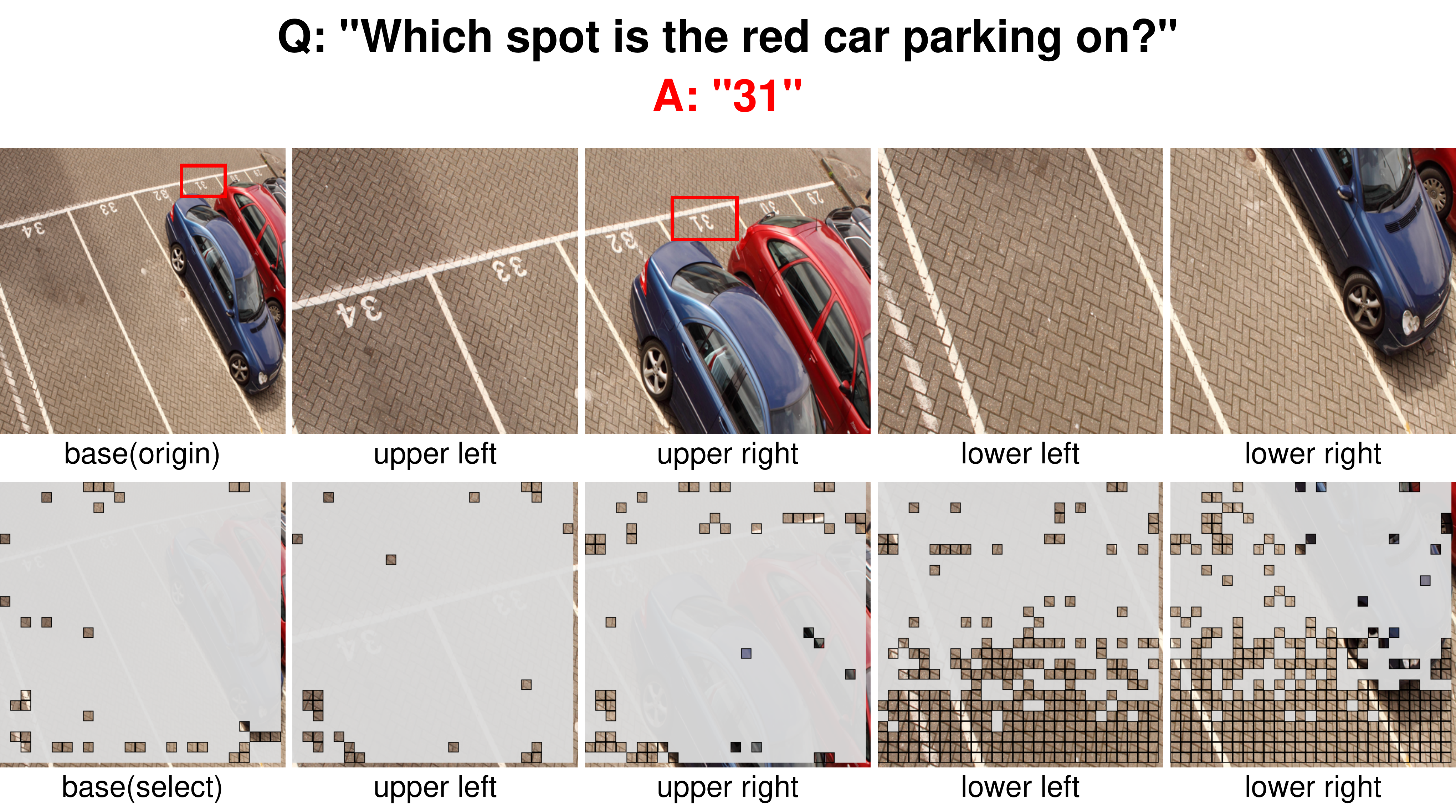}
    \end{minipage}
    \caption{Visualization of Layer 1 visual token selection: representative cases from MathVista (left) and MMVet (right)}
    \label{fig:vlcache_case_row2}
  \end{subfigure}

  \vspace{10pt}

  \begin{subfigure}[t]{\linewidth}
    \centering
    \begin{minipage}[t]{0.495\linewidth}
      \centering
      \includegraphics[width=1\linewidth]{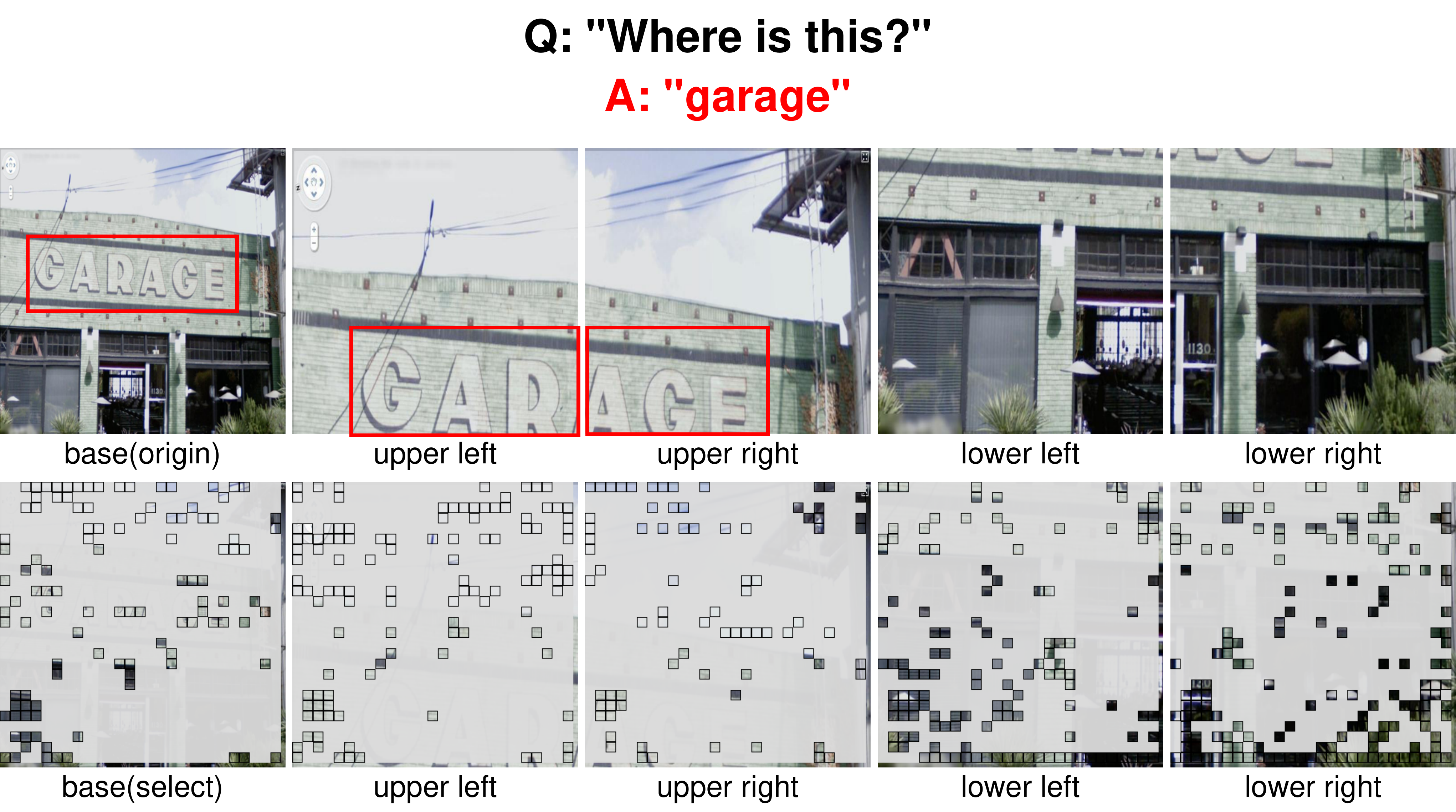}
    \end{minipage}
    \hfill
    \begin{minipage}[t]{0.495\linewidth}
      \centering
      \includegraphics[width=1\linewidth]{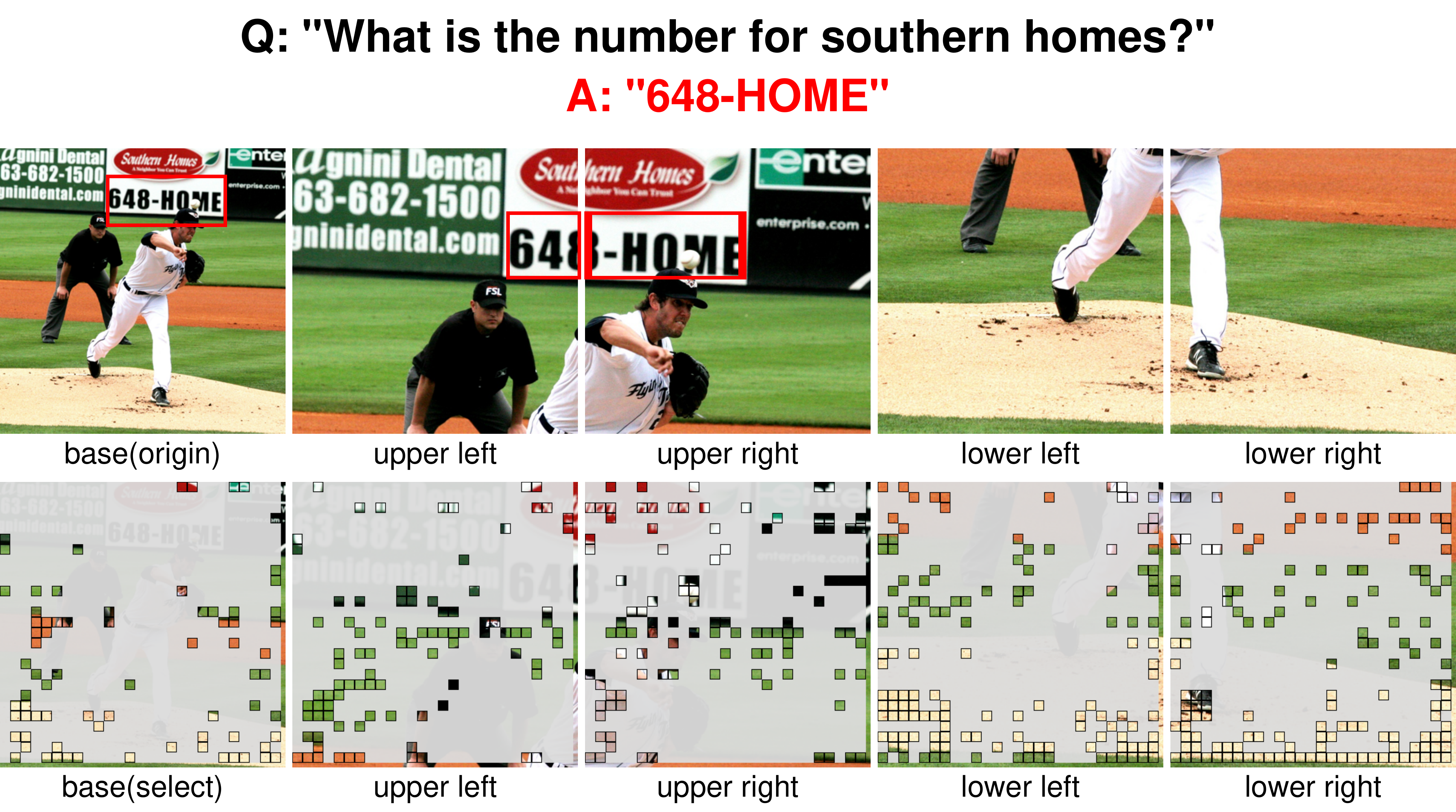}
    \end{minipage}
    \caption{Visualization of Layer 2 visual token selection: representative cases from OCRBench (left) and TextVQA (right)} 
    \label{fig:vlcache_case_row3}
  \end{subfigure}
  
  \caption{Visualization of VL-Cache selected visual tokens in layers 0–2 on LLaVA-OneVision-7B. Highlighted regions with bounding boxes are critical to the answer.} 
  \label{fig:vlcache_case}
\end{figure*}

\subsection{Visualization of VL-Cache Token Selection}
\label{sec:visualization-vlcache-visual-tokens}
We visualize visual tokens selected by VL-Cache at layers 0, 1, and 2 in LLaVA-OV-7B (5\% budget, across 6 benchmarks).
Figure~\ref{fig:vlcache_case} reveals that for the first two layers (0 and 1), selected token distributions are strikingly similar: they predominantly cluster towards the end of the visual token sequence, potentially limiting their utility in deriving final answers. Layer~2, in contrast, displays a more uniform distribution of selected tokens across the entire sequence.
Although VL-Cache's strategy allocates a higher token density to these early layers, these observed patterns suggest such aggressive, front-loaded budgeting may not be optimal.

\subsection{Visualization of Visual Sink Tokens}
\begin{figure*}[ht]
  \centering
  \begin{subfigure}[t]{\linewidth}
    \centering
    \begin{minipage}[t]{0.98\linewidth}
      \centering
      \includegraphics[width=1\linewidth]{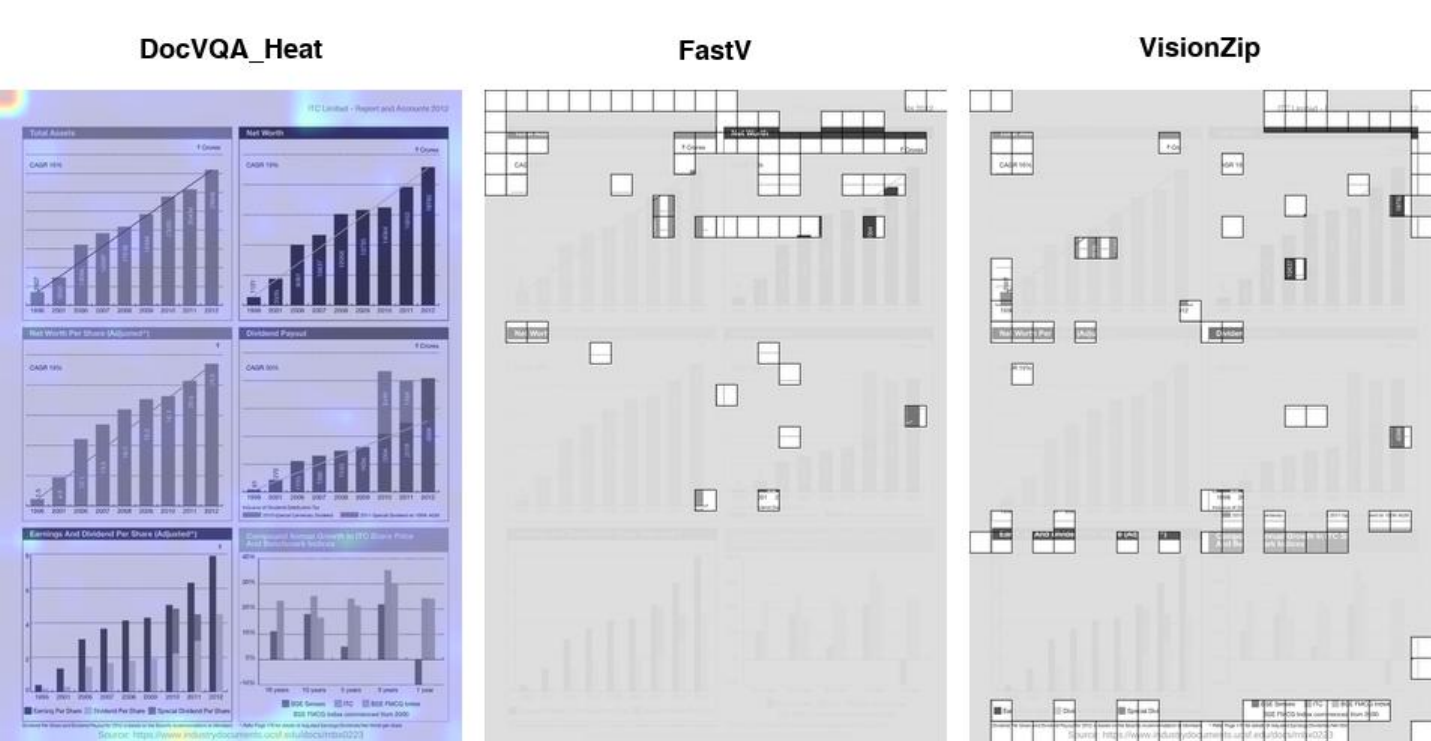}
    \end{minipage}
    \caption{Visualization results of randomly selected cases from DocVQA} 
    \label{fig:docvqa}
  \end{subfigure}
  
  \vspace{10pt}
  
  \begin{subfigure}[t]{\linewidth}
    \centering
    \begin{minipage}[t]{0.99\linewidth}
      \centering
      \includegraphics[width=1\linewidth]{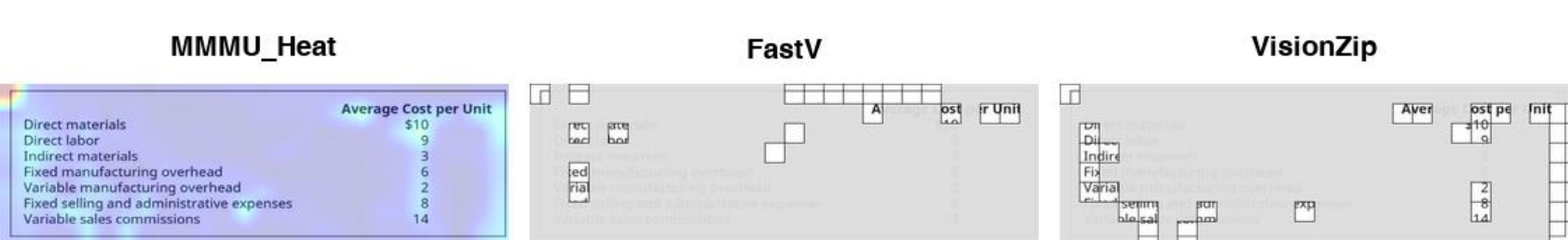}
    \end{minipage}
    \caption{Visualization results of randomly selected cases from MMMU}
    \label{fig:mmmu}
  \end{subfigure}

    \caption{Visualization of token selection strategies on DocVQA and MMMU with Qwen2-VL-7B under 10\% budget. Left to right: attention heatmap, tokens retained by FastV and VisionZip.  VisionZip selects more critical tokens than FastV.} 
    \label{fig:fastv_visionzip}
\end{figure*}

\begin{figure*}[ht]
  \centering
  \begin{subfigure}[t]{\linewidth}
    \centering
    \begin{minipage}[t]{0.98\linewidth}
      \centering
      \includegraphics[width=1\linewidth]{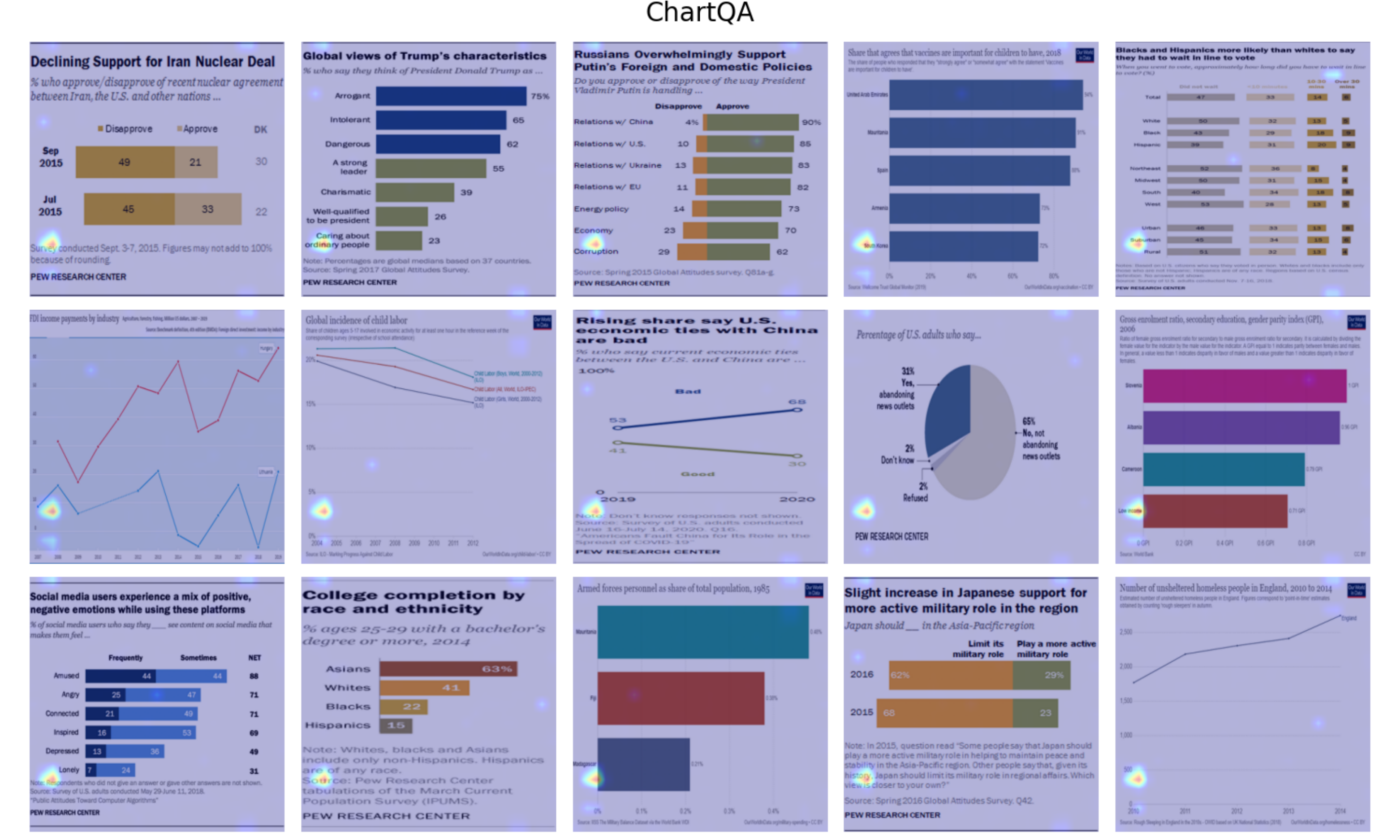}
    \end{minipage}
    \caption{Heatmap visualization results of ChartQA on LLaVA-OneVision-7B} 
    \label{fig:ov_total_chartqa}
  \end{subfigure}
  
  \vspace{10pt}
  
  \begin{subfigure}[t]{\linewidth}
    \centering
    \begin{minipage}[t]{0.99\linewidth}
      \centering
      \includegraphics[width=1\linewidth]{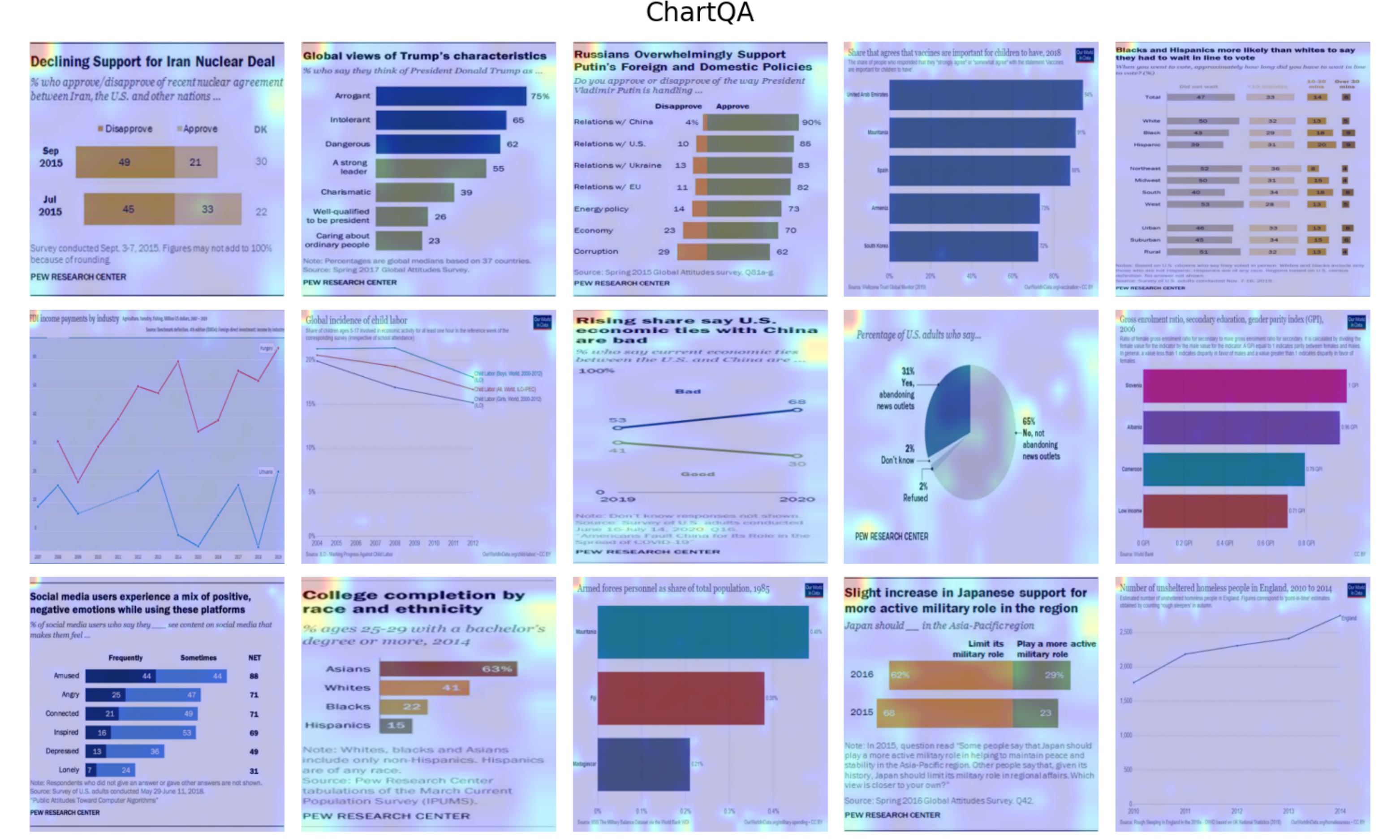}
    \end{minipage}
    \caption{Heatmap visualization results of ChartQA on Qwen2-VL-7B}
    \label{fig:vl_total_chartqa}
  \end{subfigure}

    \caption{Heatmap visualizations of 15 randomly selected examples from the ChartQA benchmark on LLaVA-OneVision-7B and Qwen2-VL-7B.} 
    \label{fig:total_chartqa}
\end{figure*}

To further illustrate the visual token sink phenomenon, we visualize attention heatmaps for 15 randomly selected ChartQA samples using LLaVA-OV-7B and Qwen2-VL-7B.
Figure \ref{fig:total_chartqa} clearly shows a visual sink emerging after the visual encoder's forward pass: image information consistently converges to a limited subset of visual tokens.
Furthermore, Figures~\ref{fig:chartqa} and \ref{fig:fastv_visionzip} compare visual token selection by VisionZip (image-guided) against FastV (text-guided). VisionZip tends to retain these critical sink tokens, whereas FastV often fails to capture many of them, explaining observed performance differences.

\label{sec:appendix_visual_sink}

\subsection{Case of Different Merge Strategies}
\label{sec:appedix_merge_dialog}
Table~\ref{tab:visual-dialog-example} contrasts LOOK-M's outputs under two token merge strategies: cross-modal versus modality-specific. Modality-specific merging yields an accurate image caption; in contrast, cross-modal merging results in an image-irrelevant output.





\begin{table*}[t!]
\centering
\begin{tabularx}{\textwidth}{@{}lX@{}} 
\toprule
\multicolumn{2}{l}{\textbf{Visual input example for image caption}} \\[0.5em]
\midrule

\multicolumn{2}{@{}p{\textwidth}@{}}{
\centering \hspace*{32pt}\includegraphics[width=0.7\textwidth]{figures/bike.pdf}} \\[1em]

\textbf{User} & 
\begin{minipage}[t]{\linewidth}
Please give a brief description of this picture.
\end{minipage} \\[1em]
\midrule

\textbf{Qwen2-VL-7B} &
\begin{minipage}[t]{\linewidth}
The image shows a classic Honda motorcycle parked on a gravel surface. The motorcycle has a black and chrome finish, with a quilted seat and a small windshield.
\end{minipage} \\[1em]
\midrule

\textbf{LOOK-M\textsubscript{origin}} &
\begin{minipage}[t]{\linewidth}
The image shows a close-up of a person's hand holding a small, round object. The object appears to be a small, metallic ball or a similar spherical item. 
\end{minipage} \\[1em]
\midrule

\textbf{LOOK-M\textsubscript{change}} &
\begin{minipage}[t]{\linewidth}
The image shows a motorcycle parked outdoors. The motorcycle has a classic design with a prominent front wheel and a seat that appears to be made of leather.
\end{minipage} \\
\bottomrule
\end{tabularx}
\caption{
Comparative analysis of response quality in Qwen2-VL-7B original model vs. LOOK-M with origin merge and modality-specific merge strategies under extreme compression ratios}
\label{tab:visual-dialog-example}
\end{table*}

\onecolumn
\small
\begin{longtable}{l l l c c c c c c}
\label{tab:kv_cache_results} \\ 
\toprule
\textbf{Benchmarks} & \textbf{Models} & \textbf{Methods} 
& \textbf{1\%} & \textbf{5\%} & \textbf{10\%} 
& \textbf{20\%} & \textbf{40\%} & \textbf{100\%} \\
\midrule
\endfirsthead

\toprule
\textbf{Benchmarks} & \textbf{Models} & \textbf{Methods} 
& \textbf{1\%} & \textbf{5\%} & \textbf{10\%} 
& \textbf{20\%} & \textbf{40\%} & \textbf{100\%} \\
\midrule
\endhead

\multirow{21}{*}{\shortstack{DocVQA\\(Split:test)\\(Metric:ANLS)}}

& \multirow{7}{*}{LLaVA-OneVision-7B}
   & Random        & 31 & 55 & 69 & 79 & 85 & 87 \\
&  & StreamingLLM  & 42 & 62 & 71 & 74 & 79 & 87\\
&  & H2O           & 58 & 78 & 82 & \textbf{86} & \textbf{87} & 87\\
&  & SnapKV        & \textbf{76} & \textbf{82} & 83 & 85 & 86 & 87\\
&  & PyramidKV     & 70 & 76 & 75 & 78 & 81 &  87\\
&  & LOOK-M        & 38 & 74 & 81 & 85 & \textbf{87} & 87\\
&  & VL-Cache      & 69 & \textbf{82} & \textbf{85} & \textbf{86} & \textbf{87} &  87\\

\cmidrule(lr){2-9}

& \multirow{7}{*}{Qwen2-VL-7B}
   & Random        & 13 & 60 & 78 & 89 & \textbf{94} &  95\\
&  & StreamingLLM  & 42 & 56 & 65 & 74 & 84 &  95\\
&  & H2O           & 47 & 67 & 79 & 89 & 93 &  95\\
&  & SnapKV        & \textbf{78} & \textbf{90} & \textbf{92} & 92 & \textbf{94} &  95\\
&  & PyramidKV     & 66 & 83 & 82 & 84 & 87 &  95\\
&  & LOOK-M        & 27 & 58 & 77 & 89 & \textbf{94} &  95\\
&  & VL-Cache      & 52 & 83 & 91 & \textbf{93} & \textbf{94} &  95\\
\cmidrule(lr){2-9}

& \multirow{6}{*}{InternVL2.5-38B}
   & StreamingLLM  & 0 & 46 & 53 & 63 & 80 &  94\\
&  & H2O           & 58 & 76 & 86 & 91 & 93 & 94\\
&  & SnapKV        & 34 & \textbf{91} & \textbf{93} & \textbf{93} & \textbf{94} &  94\\
&  & PyramidKV     & 53 & 90 & 92 & \textbf{93} & 93 &  94\\
&  & LOOK-M        & 12 & 54 & 78 & 90 & 93 &  94\\
&  & VL-Cache      & \textbf{64} & 87 & 91 & \textbf{93} & 93 &  94\\

\midrule

\multirow{21}{*}{\shortstack{ChartQA\\(Split:overall)\\(Metric:Acc)}}

& \multirow{7}{*}{LLaVA-OneVision-7B}
   & Random        & 45.04 & 59.76 & 64.56 & 69.52 & 72.72 &  80.00\\
&  & StreamingLLM  & 58.20 & 65.44 & 69.84 & 73.88 & 75.36 &  80.00\\
&  & H2O           & 67.32 & 75.48 & 76.44 & \textbf{77.48} & \textbf{77.84} &  80.00\\
&  & SnapKV        & \textbf{74.64} & \textbf{76.48} & 76.68 & 77.20 & 77.28 &  80.00\\
&  & PyramidKV     & 72.68 & 76.20 & 76.08 & 76.24 & 76.80 &  80.00\\
&  & LOOK-M        & 34.44 & 73.72 & \textbf{76.72} & 77.40 & 77.72 &  80.00\\
&  & VL-Cache      & 66.64 & 74.12 & 75.96 & 77.08 & 77.56 &  80.00\\

\cmidrule(lr){2-9}

& \multirow{7}{*}{Qwen2-VL-7B}
   & Random        & 13.00 & 45.96 & 66.08 & 72.32 & 79.68 &  81.56\\
&  & StreamingLLM  & 50.28 & 61.52 & 65.04 & 69.60 & 76.32 &  81.56\\
&  & H2O           & 54.36 & 70.12 & 74.56 & 78.60 & 80.88 &  81.56\\
&  & SnapKV        & 55.96 & \textbf{77.32} & \textbf{79.28} & \textbf{80.20} & \textbf{81.32} &  81.56\\
&  & PyramidKV     & \textbf{67.24} & 72.08 & 76.08 & 78.96 & 80.40 &  81.56\\
&  & LOOK-M        & 23.00 & 48.52 & 68.24 & 76.96 & 79.48 &  81.56\\
&  & VL-Cache      & 53.32 & 67.00 & 74.28 & 79.16 & 80.68 &  81.56\\
\cmidrule(lr){2-9}

& \multirow{6}{*}{InternVL2.5-38B}
   & StreamingLLM  & 0.68 & 67.08 & 71.68 & 73.72 & 80.00 &  88.04\\
&  & H2O           & 34.64 & 80.32 & 84.76 & 86.88 & 87.84 &  88.04\\
&  & SnapKV        & 60.60 & 84.80 & 86.64 & 87.28 & 87.92 &  88.04\\
&  & PyramidKV     & \textbf{76.40} & \textbf{85.36} & \textbf{86.68} & 87.24 & 87.52 &  88.04\\
&  & LOOK-M        & 8.80 & 54.48 & 75.48 & 84.04 & 87.16 &  88.04\\
&  & VL-Cache      & 58.60 & 82.48 & 85.88 & \textbf{87.64} & \textbf{88.16} &  88.04\\

\midrule

\multirow{21}{*}{\shortstack{TextVQA\\(Split:val)\\(Metric:Acc)}}

& \multirow{7}{*}{LLaVA-OneVision-7B}
   & Random        & 39.63 & 57.15 & 64.29 & 68.96 & 71.81 & 74.79 \\
&  & StreamingLLM  & 47.01 & 57.29 & 64.49 & 70.25 & 71.41 & 74.79 \\
&  & H2O           & 64.96 & 72.16 & \textbf{73.71} & 74.14 & 74.54 & 74.79 \\
&  & SnapKV        & \textbf{65.62} & \textbf{72.54} & 70.86 & 71.62 & 73.77 & 74.79 \\
&  & PyramidKV     & 58.56 & 68.39 & 66.37 & 66.60 & 69.70 & 74.79 \\
&  & LOOK-M        & 44.74 & 70.29 & 73.07 & \textbf{74.39} & \textbf{74.76} & 74.79 \\
&  & VL-Cache      & 61.85 & 70.73 & 72.92 & 73.46 & 74.16 & 74.79 \\

\cmidrule(lr){2-9}

& \multirow{7}{*}{Qwen2-VL-7B}
   & Random        & 5.21 & 48.04 & 61.58 & 72.78 & 79.78 &  81.82\\
&  & StreamingLLM  & 31.60 & 49.70 & 58.89 & 67.77 & 75.18 &  81.82\\
&  & H2O           & 51.16 & 64.50 & 71.40 & 77.85 & 81.13 &  81.82\\
&  & SnapKV        & 57.38 & \textbf{73.45} & \textbf{78.37} & \textbf{80.83} & 81.19 &  81.82\\
&  & PyramidKV     & \textbf{67.50} & 71.01 & 73.90 & 77.89 & 78.87 &  81.82\\
&  & LOOK-M        & 31.89 & 52.07 & 70.24 & 78.62 & 81.24 &  81.82\\
&  & VL-Cache      & 47.03 & 68.17 & 77.74 & 80.81 & \textbf{81.79} &  81.82\\
\cmidrule(lr){2-9}

& \multirow{6}{*}{InternVL2.5-38B}
   & StreamingLLM  & 0.83 & 51.49 & 53.24 & 58.97 & 70.28 &  82.87\\
&  & H2O           & 63.29 & 76.08 & 79.93 & 81.96 & 82.52 &  82.87\\
&  & SnapKV        & 36.56 & \textbf{80.58} & \textbf{82.15} & \textbf{82.69} & \textbf{82.77} &  82.87\\
&  & PyramidKV     & 56.73 & 79.70 & 80.64 & 81.91 & 82.16 &  82.87\\
&  & LOOK-M        & 12.56 & 61.42 & 76.33 & 81.13 & 82.34 &  82.87\\
&  & VL-Cache      & \textbf{64.40} & 76.54 & 80.35 & 82.29 & 82.71 &  82.87\\

\midrule

\pagebreak

\multirow{21}{*}{\shortstack{OCRBench\\(Split:test)\\(Metric:Acc)}}

& \multirow{7}{*}{LLaVA-OneVision-7B}
   & Random        & 67 & 335 & 455 & 530 & 564 &  595\\
&  & StreamingLLM  & 94 & 222 & 320 & 386 & 502 &  595\\
&  & H2O           & 229 & 407 & \textbf{495} & \textbf{554} & 588 &  595\\
&  & SnapKV        & \textbf{281} & \textbf{415} & 445 & 475 & 553 &  595\\
&  & PyramidKV     & 275 & 337 & 354 & 358 & 426 &  595\\
&  & LOOK-M        & 81 & 375 & 463 & 544 & \textbf{589} &  595\\
&  & VL-Cache      & 231 & 372 & 460 & 505 & 551 &  595\\

\cmidrule(lr){2-9}

& \multirow{7}{*}{Qwen2-VL-7B}
   & Random        & 109 & 192 & 336 & 491 & 658 &  813\\
&  & StreamingLLM  & 116 & 199 & 248 & 349 & 465 &  813\\
&  & H2O           & 153 & 334 & 441 & 570 & 713 &  813\\
&  & SnapKV        & 199 & \textbf{428} & \textbf{542} & \textbf{654} & \textbf{725} &  813\\
&  & PyramidKV     & \textbf{241} & 417 & 535 & 642 & 662 &  813\\
&  & LOOK-M        & 112 & 202 & 299 & 441 & 587 &  813\\
&  & VL-Cache      & 216 & 386 & 505 & 592 & 660 &  813\\
\cmidrule(lr){2-9}

& \multirow{6}{*}{InternVL2.5-38B}
   & StreamingLLM  & 51 & 121 & 162 & 214 & 399 &  802\\
&  & H2O           & 143 & 487 & 656 & 752 & 782 &  802\\
&  & SnapKV        & 359 & \textbf{652} & \textbf{720} & \textbf{777} & \textbf{791} &  802\\
&  & PyramidKV     & \textbf{478} & 637 & 664 & 734 & 765 &  802\\
&  & LOOK-M        & 18 & 183 & 491 & 695 & 771 &  802\\
&  & VL-Cache      & 193 & 487 & 630 & 733 & 778 &  802\\

\midrule

\multirow{21}{*}{\shortstack{MathVista  \\(Split:testmini) \\ (format: COT)\\(Metric: GPT)}}

& \multirow{7}{*}{LLaVA-OneVision-7B}
   & Random        & 41.10 & 44.20 & 47.80 & 53.50 & 55.90 &  58.20\\
&  & StreamingLLM  & 40.40 & 43.10 & 45.20 & 48.50 & 54.40 &  58.20\\
&  & H2O           & 42.40 & 48.90 & 53.70 & 55.40 & \textbf{56.60} &  58.20\\
&  & SnapKV        & \textbf{54.40} & \textbf{55.30} & \textbf{56.30} & \textbf{56.70} & \textbf{56.60} &  58.20\\
&  & PyramidKV     & 53.00 & 54.10 & 54.50 & 56.50 & 56.30 &  58.20\\
&  & LOOK-M        & 44.30 & 53.10 & 53.60 & 54.00 & 55.10 &  58.20\\
&  & VL-Cache      & 49.50 & 53.40 & 55.80 & 55.90 & 56.50 &  58.20\\

\cmidrule(lr){2-9}

& \multirow{7}{*}{Qwen2-VL-7B}
   & Random        & 45.00 & 49.60 & 54.00 & 56.30 & 57.20 &  58.30\\
&  & StreamingLLM  & 47.40 & 54.80 & 55.30 & 56.20 & 56.40 &  58.30\\
&  & H2O           & 49.00 & 55.60 & 55.80 & 57.00 & 57.80 &  58.30\\
&  & SnapKV        & 49.90 & 56.50 & 57.20 & 57.40 & \textbf{58.00} &  58.30\\
&  & PyramidKV     & \textbf{50.30} & \textbf{57.30} & \textbf{57.60} & \textbf{57.50} & \textbf{58.00} &  58.30\\
&  & LOOK-M        & 35.80 & 54.30 & 56.20 & 55.50 & 56.70 &  58.30\\
&  & VL-Cache      & 52.30 & 55.00 & 54.80 & 56.40 & 56.70 &  58.30\\
\cmidrule(lr){2-9}

& \multirow{6}{*}{InternVL2.5-38B}
   & StreamingLLM  & 40.80 & 48.00 & 50.10 & 54.50 & 59.10 &  70.20\\
&  & H2O           & 50.40 & 60.90 & 64.60 & \textbf{68.60} & 70.50 &  70.20\\
&  & SnapKV        & 53.70 & 64.30 & \textbf{68.40} & 68.40 & 70.00 &  70.20\\
&  & PyramidKV     & \textbf{56.10} & \textbf{66.10} & 66.20 & 68.00 & 69.60 &  70.20\\
&  & LOOK-M        & 38.30 & 56.50 & 61.60 & 66.70 & 69.40 &  70.20\\
&  & VL-Cache      & 53.90 & 60.70 & 65.00 & 67.60 & \textbf{70.60} &  70.20\\

\midrule

\multirow{21}{*}{\shortstack{MMVet  \\(Split:test) \\ (Metric: GPT)}}

& \multirow{7}{*}{LLaVA-OneVision-7B}
   & Random        & 19.60 & 34.10 & 40.80 & 46.50 & 47.70 &  55.00\\
&  & StreamingLLM  & 24.60 & 39.80 & 46.30 & 48.10 & 50.00 &  55.00\\
&  & H2O           & \textbf{33.60} & 46.10 & \textbf{51.60} & \textbf{53.20} & \textbf{54.10} &  55.00\\
&  & SnapKV        & 32.90 & \textbf{47.90} & 50.20 & 52.20 & 53.30 &  55.00\\
&  & PyramidKV     & \textbf{33.60} & 43.60 & 47.80 & 47.20 & 51.60 &  55.00\\
&  & LOOK-M        & 20.60 & 45.20 & 49.10 & 52.20 & 53.60 &  55.00\\
&  & VL-Cache      & 30.00 & 42.50 & 49.30 & 50.20 & 51.90 &  55.00\\

\cmidrule(lr){2-9}

& \multirow{7}{*}{Qwen2-VL-7B}
   & Random        & 4.10 & 10.90 & 22.30 & 33.90 & 52.20 &  65.40\\
&  & StreamingLLM  & 1.10 & 16.30 & 28.00 & 38.60 & 44.60 &  65.40\\
&  & H2O           & 8.30 & 27.30 & 39.20 & 53.70 & 56.00 &  65.40\\
&  & SnapKV        & \textbf{12.10} & \textbf{33.50} & \textbf{43.50} & \textbf{55.10} & \textbf{56.50} &  65.40\\
&  & PyramidKV     & 11.20 & 31.20 & 39.30 & 46.70 & 51.40 &  65.40\\
&  & LOOK-M        & 1.40 & 17.80 & 30.40 & 46.30 & 55.30 &  65.40\\
&  & VL-Cache      & 7.00 & 25.80 & 38.10 & 47.40 & 53.30 &  65.40\\
\cmidrule(lr){2-9}

& \multirow{6}{*}{InternVL2.5-38B}
   & StreamingLLM  & 0.00 & 21.20 & 32.20 & 37.70 & 53.70 &  69.40\\
&  & H2O           & 15.90 & 47.50 & 58.80 & \textbf{68.00} & 70.70 &  69.40\\
&  & SnapKV        & 23.00 & \textbf{51.40} & \textbf{63.60} & 67.10 & 69.20 &  69.40\\
&  & PyramidKV     & \textbf{33.30} & 51.10 & 57.70 & 64.00 & 67.10 &  69.40\\
&  & LOOK-M        & 2.80 & 29.20 & 50.50 & 57.80 & \textbf{71.10} &  69.40\\
&  & VL-Cache      & 16.50 & 35.50 & 48.40 & 57.50 & 67.00 &  69.40\\

\midrule
\pagebreak

\multirow{21}{*}{\shortstack{LLaVA-Wilder  \\(Split:test) \\ (Metric: GPT)}}

& \multirow{7}{*}{LLaVA-OneVision-7B}
   & Random        & 34.10 & 51.60 & 58.20 & 63.60 & 67.70 &  71.40\\
&  & StreamingLLM  & 48.00 & 65.70 & 69.20 & \textbf{71.20} & 69.40 &  71.40\\
&  & H2O           & 62.50 & \textbf{69.20} & \textbf{70.00} & 71.00 & 70.60 &  71.40\\
&  & SnapKV        & \textbf{63.70} & 66.20 & 69.40 & 71.00 & \textbf{70.90} &  71.40\\
&  & PyramidKV     & 62.30 & 67.20 & 66.80 & 67.50 & 69.00 &  71.40\\
&  & LOOK-M        & 43.40 & 65.50 & 65.70 & 66.80 & 68.40 &  71.40\\
&  & VL-Cache      & 62.40 & 67.10 & 69.10 & 69.50 & \textbf{70.90} &  71.40\\

\cmidrule(lr){2-9}

& \multirow{7}{*}{Qwen2-VL-7B}
   & Random        & 12.50 & 21.00 & 31.50 & 42.30 & 59.90 &  72.70\\
&  & StreamingLLM  & 19.50 & 39.60 & 53.40 & 61.00 & 69.20 &  72.70\\
&  & H2O           & 37.20 & 57.30 & 62.00 & 65.60 & \bf{72.50} &  72.70\\
&  & SnapKV        & 27.70 & 58.10 & \bf{64.50} & 66.60 & 69.50 &  72.70\\
&  & PyramidKV     & 36.30 & \bf{58.50} & 64.10 & 65.00 & 68.70 &  72.70\\
&  & LOOK-M        & 19.10 & 52.30 & 60.50 & 67.80 & 69.80 &  72.70\\
&  & VL-Cache      & \bf{39.80} & 57.90 & 63.40 & \bf{68.00} & 69.30 &  72.70\\
\cmidrule(lr){2-9}

& \multirow{6}{*}{InternVL2.5-38B}
   & StreamingLLM  & 21.20 & 45.80 & 58.70 & 66.60 & 71.00 & 74.80 \\
&  & H2O           & 51.90 & 69.10 & \bf{72.60} & \bf{73.40} & 74.10 & 74.80\\
&  & SnapKV        & 52.70 & 69.10 & 70.90 & 72.60 & \bf{74.40} & 74.80 \\
&  & PyramidKV     & \bf{56.20} & \bf{69.90} & 70.30 & 71.80 & 72.30 & 74.80 \\
&  & LOOK-M        & 16.80 & 59.20 & 67.10 & 71.80 & 73.60 & 74.80 \\
&  & VL-Cache      & 55.20 & 66.80 & 69.80 & 72.80 & 72.00 & 74.80 \\

\midrule

\multirow{21}{*}{\shortstack{ImageDC  \\(Split:DC100\_EN) \\(Metric: GPT)}}

& \multirow{7}{*}{LLaVA-OneVision-7B}
   & Random        & 27.35 & 54.35 & 73.80 & 83.35 & 86.65 &  87.25\\
&  & StreamingLLM  & 38.40 & 77.95 & 80.50 & 85.25 & 86.50 &  87.25\\
&  & H2O           & \bf{77.05} & \bf{85.30} & \bf{86.00} & \bf{86.75} & 86.53 &  87.25\\
&  & SnapKV        & 28.00 & 77.00 & 80.15 & 83.25 & 86.25 &  87.25\\
&  & PyramidKV     & 27.75 & 69.55 & 75.45 & 80.40 & 84.00 &  87.25\\
&  & LOOK-M        & 24.55 & 83.95 & 85.99 & 86.50 & \bf{86.70} &  87.25\\
&  & VL-Cache      & 21.00 & 51.80 & 70.95 & 80.00 & 84.95 &  87.25\\

\cmidrule(lr){2-9}

& \multirow{7}{*}{Qwen2-VL-7B}
   & Random        & 15.25 & 38.30 & 56.60 & 69.50 & 82.10 &  86.35\\
&  & StreamingLLM  & 16.70 & 31.35 & 76.95 & 82.15 & 85.10 &  86.35\\
&  & H2O           & \bf{29.55} & \bf{82.25} & \bf{83.85} & \bf{85.80} & \bf{85.80} &  86.35\\
&  & SnapKV        & 17.30 & 29.85 & 64.05 & 77.95 & 83.75 &  86.35\\
&  & PyramidKV     & 16.90 & 31.05 & 56.80 & 69.90 & 78.55 &  86.35\\
&  & LOOK-M        & 24.05 & 23.65 & 29.15 & 82.10 & 85.55 &  86.35\\
&  & VL-Cache      & 15.05 & 17.80 & 41.80 & 70.35 & 82.30 &  86.35\\
\cmidrule(lr){2-9}

& \multirow{6}{*}{InternVL2.5-38B}
   & StreamingLLM  & 13.25 & 48.10 & 51.40 & 73.30 & 83.50 &  86.45\\
&  & H2O           & \bf{59.60} & \bf{83.55} & \bf{85.05} & \bf{85.85} & \bf{86.65} &  86.45\\
&  & SnapKV        & 37.60 & 75.05 & 80.60 & 85.50 & 85.25 &  86.45\\
&  & PyramidKV     & 46.75 & 77.65 & 81.95 & 84.45 & 86.10 &  86.45\\
&  & LOOK-M        & 19.35 & 62.00 & 70.50 & 76.50 & 86.40 &  86.45\\
&  & VL-Cache      & 24.90 & 56.55 & 71.40 & 82.95 & 85.50 &  86.45\\

\bottomrule
\caption{Main results of various KV cache compression methods on different models and tasks.\textbf{Bold} denotes the best result under the same setting.} 
\end{longtable}

\end{document}